\documentclass[Afour,sageh,times]{sagej}

\usepackage{bbding}
\usepackage{moreverb,url}
\usepackage{url}
\usepackage{tablefootnote}
\usepackage[linktoc=all,colorlinks,bookmarksopen,bookmarksnumbered,citecolor=blue,urlcolor=blue]{hyperref}
\usepackage{balance}
\usepackage[table]{xcolor}
\usepackage{graphicx}
\usepackage{amsfonts}
\usepackage{fancyhdr}
\usepackage{comment}
\usepackage{times}
\usepackage{amsmath}
\usepackage{changepage}
\usepackage{amssymb}
\usepackage{enumerate}
\usepackage{algorithmic}
\usepackage{bm}
\usepackage{calligra}
\usepackage{multirow}
\usepackage{tabularx}
\usepackage{booktabs}
\usepackage{mathtools}
\usepackage{arydshln}
\usepackage{latexsym} 
\usepackage{amssymb}
\usepackage{color}
\usepackage[ruled,vlined]{algorithm}
\usepackage{ulem}
\usepackage{float}
\usepackage{tabstackengine}
\usepackage{siunitx}
\usepackage{colortbl}
\usepackage{hhline}
\usepackage{bookmark}
\usepackage{dirtree}
\usepackage{subfigure}
\usepackage{orcidlink}
\usepackage{wasysym}
\usepackage{enumitem}
\usepackage{overpic}
\usepackage{pdflscape}
\usepackage{afterpage}

\newcommand\BibTeX{{\rmfamily B\kern-.05em \textsc{i\kern-.025em b}\kern-.08em
T\kern-.1667em\lower.7ex\hbox{E}\kern-.125emX}}

\newcommand{\bt}{\textcolor[rgb]{0,0,1}}

\def\volumeyear{2024}

\begin{document}

\runninghead{Wei et al.}

% \title{FusionPortableV2:  A Multi-Sensor Multi-Scale Dataset for Evaluation of Localization and Mapping Accuracy (and Beyond) on Diverse Platforms}

\title{FusionPortableV2: A Unified Multi-Sensor Dataset for Generalized SLAM Across Diverse Platforms and Scalable Environments}

% \author{
%   Hexiang Wei$^{1*}$ $^{\orcidlink{0000-0000-0000-0000}}$, Jianhao Jiao\affilnum{2*}$^{\orcidlink{0000-0000-0000-0000}}$, Xiangcheng Hu, Jingwen Yu, Xupeng Xie, Jin Wu,
%   Yilong Zhu, Yuxuan Liu, Lujia Wang, Ming Liu
% }
% Lujia Wang\affilnum{1},

\author{Hexiang Wei\affilnum{1}$^{\bm{*}}$, Jianhao Jiao\affilnum{2}$^{\bm{*}}$, Xiangcheng Hu\affilnum{1}, Jingwen Yu\affilnum{1,3}, Xupeng Xie\affilnum{1}, Jin Wu\affilnum{1}, Yilong Zhu\affilnum{1}, Yuxuan Liu\affilnum{1}, Lujia Wang\affilnum{4}, Ming Liu\affilnum{4}}

\affiliation{\affilnum{1}Department of Electronic and Computer Engineering, The Hong Kong University of Science and Technology, Hong Kong SAR, China.\\
    \affilnum{2}Robot Perception and Learning Lab, Department of Computer Science, University College London, UK.\\
    %       {\tt\small hweiak@connect.ust.hk}.
    \affilnum{3}Shenzhen Key Laboratory of Robotics and Computer Vision, Southern University of Science and Technology, China.\\
    \affilnum{4}The Hong Kong University of Science and Technology (Guangzhou), Guangzhou, China.\\
    $^{\bm{*}}$Hexiang Wei and Jianhao Jiao Contributed Equally.
    %       {\tt\small ucacjji@ucl.ac.uk} \\
    %    \textit{Corresponding Author: Jianhao Jiao} \\
}

\corrauth{Jianhao Jiao, Department of Computer Science, University College London, Gower Street, WC1E 6BT, London, UK.}
\email{ucacjji@ucl.ac.uk, jiaojh1994@gmail.com}

\begin{abstract}
    Simultaneous Localization and Mapping (SLAM) technology has been widely applied in various robotic scenarios, from rescue operations to autonomous driving.
    However, the generalization of SLAM algorithms remains a significant challenge, as current datasets often lack scalability in terms of platforms and environments.
    To address this limitation, we present FusionPortableV2, a multi-sensor SLAM dataset featuring sensor diversity, varied motion patterns, and a wide range of environmental scenarios.
    Our dataset comprises $27$ sequences, spanning over $2.5$ hours and collected from four distinct platforms: a handheld suite, a legged robot, a unmanned ground vehicle (UGV), and a vehicle. These sequences cover diverse settings, including buildings, campuses, and urban areas, with a total length of $38.7km$. Additionally, the dataset includes ground-truth (GT) trajectories and RGB point cloud maps covering approximately $0.3km^2$.
    To validate the utility of our dataset in advancing SLAM research, we assess several state-of-the-art (SOTA) SLAM algorithms. Furthermore, we demonstrate the dataset's broad application beyond traditional SLAM tasks by investigating its potential for monocular depth estimation. The completae dataset, including sensor data, GT, and calibration details, is accessible at \href{https://fusionportable.github.io/dataset/fusionportable\_v2}{https://fusionportable.github.io/dataset/fusionportable\_v2}.
\end{abstract}

\keywords{SLAM Dataset, Sensor Fusion, Mapping, Mobile Robots}

\maketitle
\renewcommand{\thefootnote}{\arabic{footnote}}

\section{Introduction}
\label{sec:introduction}
\runninghead{Wei \textit{et~al.}}
\subsection{Motivation}
\label{sec:motivation}
Real-world robotic datasets are vital for algorithm development.
They provide diverse environments and challenging real-world sequences for training and evaluation of systems.
They reduce costs and workforce requirements, such as system integration, calibration, and field operations \citep{nguyen2022ntu}, promoting broader participation in robotic research and fostering novel algorithm development.
As robotics research transitions from traditional handcrafted methods to data-driven and hybrid approaches \citep{brohan2022rt,shah2023vint}, the importance of these datasets continues to grow.
Building upon this trend, this paper contributes to the exploration of SLAM dataset's potential by introducing a diverse multi-sensor dataset.
Our goal aims to develop a dataset to address the generalization challenges in robotic perception and navigation across various environments and operational conditions.

As outlined in Table \ref{tab:related_work_dataset}, recent SLAM datasets exhibit two key trends:
(1) they encompass a variety of large-scale real-world environments (e.g., urban roads \citep{burnett2023boreas}, subterranean \citep{reinke2022locus}, and forests \citep{knights2023wild}), and
(2) they incorporate heterogeneous sensors to enhance perceptual awareness.
For example, cameras capture dense and high-resolution 2D images containing texture and pattern information of surroundings.
However, cameras are vulnerable to adverse illumination conditions (\textit{e.g.}, darkness and glare) due to their passive nature in measuring.
In contrast, range sensors such as LiDARs and Radars provide sparse but accurate structural information by exploiting their respective light sources.
Integrating cameras with range sensors often yields more reliable results across a variety of perception tasks compared to relying on a single sensor type.
Therefore, the complementary strengthen offered by various sensors drives the exploration of novel sensor fusion algorithms \citep{lin2021r3live}, enabling robots to autonomously navigate through diverse environments with increased accuracy and robustness.

Despite the advancements in SLAM, a gap persists between the diversity of real-world scenarios and available datasets, especially concerning variations in environments, sensor modalities, and platforms executing diverse motions.
This disparity affects the further development and evaluation of SLAM algorithms, potentially limiting their generalization capability and robustness in diverse real-world environments.
Drawing on the recent success in the generalized manipulation and navigation models such as RT-2 \citep{brohan2023rt} and GNM \citep{shah2023gnm}, it is our conviction that datasets featuring high motion and environmental diversity are crucial for the development of a versatile and generalized SLAM system.

\subsection{Contributions}
\label{sec:contribution}
This paper aims to provide diverse, high-quality data using a consistent hardware setup, facilitating the development and benchmarking of robust and generalized SLAM systems.
To achieve this objective, we present a comprehensive multi-sensor dataset, along with a detailed description of our data collection methodologies and complete benchmarking tools for evaluation.
Building upon our previous FusionPortable dataset \citep{jiao2022fusionportable}, we introduce \textbf{FusionPortableV2}, a significant upgrade that expands the dataset in terms of data modalities, scenarios, and motion capabilities. This paper presents two main contributions:

\begin{enumerate}[leftmargin=0.45cm]
    \item \textbf{Diversity}: We have improved the FusionPortable dataset by expanding the range of platforms to include high-speed vehicles and to cover over $12$ types of environments (\textit{e.g.,} campuses, underground areas, parking lots, highways, etc.). These enhancements not only increase the dataset's diversity and complexity but also substantially improve the accuracy of ground truth data and the integration of raw kinematic data collected from ground robots. Our collection includes $27$ sequences, spanning $2.5$ hours and covering a total distance of $38.7km$. We also provide valuable ground-truth trajectories and maps encompassing a campus area of approximately $0.3km^{2}$ to serve as benchmarks for evaluating various navigation algorithms beyond SLAM.

    \item \textbf{Versatile}: We demonstrate the dataset's value for training models and evaluating algorithms through experiments on SLAM and monocular depth estimation. These experiments highlight the challenges and the need for cross-platform adaptability in diverse environments. With its rich variety of platforms and settings, the dataset can be extended to benchmarks for tasks such as cross-view and cross-modality localization \citep{shi2023boosting}, environmental mapping \citep{kerbl20233d,hu2024ms}, and navigation tasks with the high-quality map as simulation.
\end{enumerate}
During the platform development and data collection process, we addressed numerous technical challenges and meticulously documented the issues encountered and their solutions. This guidance should be a valuable resource for future researchers in the field. To promote collaborative advancements, we have publicly released all data and implementation details. We believe this will open up a wide range of research opportunities in field robotics and aid in the development of versatile, resilient robotic systems.

\subsection{Organization}
The remainder of this paper is structured in the following manner:
Section \ref{sec:related_works} discusses related works most of on SLAM datasets and summarizes key contributions of this paper.
Section \ref{sec:system_overview} outlines the hardware setup and sensor details.
Section \ref{sec:calibration} covers sensor calibration procedures.
Section \ref{sec:dataset_description} describes the dataset, including platform characteristics and scenarios.
Section \ref{sec:data_post_process} introduces details post-processing steps on raw sensor measurements and GT data.
Section \ref{sec:experiment} presents the methodologies used for evaluating localization, mapping, and monocular depth estimation. Known issues of this dataset are also discussed.
Finally, Section \ref{sec:conclusion} concludes the paper and suggests directions for future research.

\section{Related Works}
\label{sec:related_works}
\runninghead{Wei \textit{et~al.}}

\begin{table*}[h]
  \centering
  \caption{Comparing SLAM Datasets: Highlights differences in platforms, sensors, and ground-truth methods, focusing on FusionPortableV2's attributes.
    It classifies environment scales into small ($<100m^2$), medium ($<500m^2$), and large, acknowledging this as a broad estimation. Abbreviations include: UGV (Unmanned ground vehicle), MoCap (Motion capture), LT (Laser tracker), O (Out-of-the-box kinematic-inertial odometry),S (Subterranean), and V (Vegetated) areas.
    The symbol \CIRCLE\ and \Circle\ indicate whether dataset satisfies the option or not. \RIGHTcircle\ indicates that the dataset misses robot kinematic data.}
  \renewcommand\arraystretch{0.89}
  \renewcommand\tabcolsep{1.5pt}
  \scriptsize
  \begin{tabular}{clcccccccccccccc}
    \toprule[0.03cm]
    \multirow{3}{*}{\rotatebox{90}{\textbf{}}}                                & \multirow{3}{*}{\textbf{Dataset}}            & \multicolumn{5}{c}{\textbf{Sensors Modality
    }}                                                                        & \multicolumn{4}{c}{\textbf{Mobile Platform}} & \multicolumn{3}{c}{\textbf{Environment Scale}} & \multirow{3}{*}{\textbf{GT Pose}} & \multirow{3}{*}{\textbf{GT Map}}                                                                                                                                                                                  \\
    \cmidrule(lr){3-7} \cmidrule(lr){8-11} \cmidrule(lr){12-14}
                                                                              &                                              & \textbf{IMU}                                   & \begin{tabular}[c]{@{}c@{}}\textbf{Frame} \\ \textbf{Camera}\end{tabular}         & \begin{tabular}[c]{@{}c@{}}\textbf{Event} \\ \textbf{Camera}\end{tabular}        & \textbf{LiDAR} & \textbf{GPS} & \textbf{Handhold} & \textbf{UGV}    & \textbf{Legged} & \textbf{Vehicle} & \textbf{Small} & \textbf{Medium} & \textbf{Large}                   \\
    \toprule[0.03cm]

    \multirow{13}{*}{\rotatebox{90}{\normalsize \textbf{Portable Suite}}}     & \begin{tabular}[c]{@{}l@{}}UZH-Event\\\citep{mueggler2017event} \end{tabular}
                                                                              & \CIRCLE                                      & \Circle                                        & \CIRCLE                           & \Circle                          & \Circle        & \CIRCLE      & \Circle           & \Circle         & \Circle         & \CIRCLE          & \CIRCLE        & \Circle         & MoCap                  & /       \\
    %& \CheckmarkBold & \XSolidBrush
                                                                              & \begin{tabular}[c]{@{}l@{}}PennCOSYVIO\\\citep{pfrommer2017penncosyvio} \end{tabular}
                                                                              & \CIRCLE                                      & \CIRCLE                                        & \Circle                           & \Circle                          & \Circle        & \CIRCLE      & \Circle           & \Circle         & \Circle         & \CIRCLE          & \CIRCLE        & \Circle         & Apriltag               & /       \\

                                                                              & \begin{tabular}[c]{@{}l@{}}TUM VI\\\citep{schubert2018tum} \end{tabular}
                                                                              & \CIRCLE                                      & \CIRCLE                                        & \Circle                           & \Circle                          & \Circle        & \CIRCLE      & \Circle           & \Circle         & \Circle         & \CIRCLE          & \CIRCLE        & \CIRCLE         & MoCap                  & /       \\

                                                                              & \begin{tabular}[c]{@{}l@{}}UMA-VI\\\citep{zuniga2020vi} \end{tabular}
                                                                              & \CIRCLE                                      & \CIRCLE                                        & \Circle                           & \Circle                          & \Circle        & \CIRCLE      & \Circle           & \Circle         & \Circle         & \CIRCLE          & \CIRCLE        & \CIRCLE         & SfM                    & /       \\

                                                                              & \begin{tabular}[c]{@{}l@{}}Newer College\\\citep{ramezani2020newer} \end{tabular}
                                                                              & \CIRCLE                                      & \CIRCLE                                        & \Circle                           & \CIRCLE                          & \Circle        & \CIRCLE      & \Circle           & \Circle         & \Circle         & \CIRCLE          & \CIRCLE        & \Circle         & ICP                    & Scanner \\

                                                                              & \begin{tabular}[c]{@{}l@{}}Hilti-Oxford\\\citep{zhang2022hilti} \end{tabular}
                                                                              & \CIRCLE                                      & \CIRCLE                                        & \Circle                           & \CIRCLE                          & \Circle        & \CIRCLE      & \Circle           & \Circle         & \Circle         & \CIRCLE          & \Circle        & \Circle         & ICP                    & Scanner \\

                                                                              & \begin{tabular}[c]{@{}l@{}}VECTor\\\citep{gao2022vector} \end{tabular}
                                                                              & \CIRCLE                                      & \CIRCLE                                        & \CIRCLE                           & \CIRCLE                          & \Circle        & \CIRCLE      & \Circle           & \Circle         & \Circle         & \CIRCLE          & \Circle        & \Circle         & MoCap/\ ICP            & Scanner \\
    \midrule[0.01cm]

    \multirow{13}{*}{\rotatebox{90}{\normalsize \textbf{Autonomous Driving}}} & \begin{tabular}[c]{@{}l@{}}MIT DARPA\\\citep{huang2010high} \end{tabular}
                                                                              & \CIRCLE                                      & \CIRCLE                                        & \Circle                           & \CIRCLE                          & \CIRCLE        & \Circle      & \Circle           & \Circle         & \CIRCLE         & \Circle          & \Circle        & \CIRCLE         & D-GNSS                 & /       \\

                                                                              & \begin{tabular}[c]{@{}l@{}}KITTI\\\citep{geiger2013vision} \end{tabular}
                                                                              & \CIRCLE                                      & \CIRCLE                                        & \Circle                           & \CIRCLE                          & \CIRCLE        & \Circle      & \Circle           & \Circle         & \CIRCLE         & \Circle          & \Circle        & \CIRCLE         & RTK-GNSS               & /       \\

                                                                              & \begin{tabular}[c]{@{}l@{}}Oxford RobotCar\\\citep{maddern20171} \end{tabular}
                                                                              & \CIRCLE                                      & \CIRCLE                                        & \Circle                           & \CIRCLE                          & \CIRCLE        & \Circle      & \Circle           & \Circle         & \CIRCLE         & \Circle          & \Circle        & \CIRCLE         & D-GNSS                 & /       \\

                                                                              & \begin{tabular}[c]{@{}l@{}}KAIST-Complex Urban\\\citep{jeong2019complex} \end{tabular}
                                                                              & \CIRCLE                                      & \CIRCLE                                        & \Circle                           & \CIRCLE                          & \CIRCLE        & \Circle      & \Circle           & \Circle         & \CIRCLE         & \Circle          & \Circle        & \CIRCLE         & RTK-GNSS               & SLAM    \\

                                                                              & \begin{tabular}[c]{@{}l@{}}Ford Multi-AV\\\citep{agarwal2020ford} \end{tabular}
                                                                              & \CIRCLE                                      & \CIRCLE                                        & \Circle                           & \CIRCLE                          & \CIRCLE        & \Circle      & \Circle           & \Circle         & \CIRCLE         & \Circle          & \Circle        & \CIRCLE         & D-GNSS                 & SLAM    \\

                                                                              & \begin{tabular}[c]{@{}l@{}}DSEC\\\citep{gehrig2021dsec} \end{tabular}
                                                                              & \Circle                                      & \CIRCLE                                        & \CIRCLE                           & \CIRCLE                          & \CIRCLE        & \Circle      & \Circle           & \Circle         & \CIRCLE         & \Circle          & \Circle        & \CIRCLE         & RTK-GNSS               & SLAM    \\

                                                                              & \begin{tabular}[c]{@{}l@{}}Boreas\\\citep{burnett2023boreas} \end{tabular}
                                                                              & \CIRCLE                                      & \CIRCLE                                        & \Circle                           & \CIRCLE                          & \CIRCLE        & \Circle      & \Circle           & \Circle         & \CIRCLE         & \Circle          & \Circle        & \CIRCLE         & RTX-GNSS               & /       \\

    \midrule[0.01cm]

    % \midrule[0.01cm]

    \multirow{13}{*}{\rotatebox{90}{\normalsize \textbf{More Diversity}}}     & \begin{tabular}[c]{@{}l@{}}NCLT\\\citep{carlevaris2016university} \end{tabular}
                                                                              & \CIRCLE                                      & \CIRCLE                                        & \Circle                           & \CIRCLE                          & \CIRCLE        & \Circle      & \CIRCLE           & \Circle         & \Circle         & \CIRCLE          & \CIRCLE        & \Circle         & RTK-GNSS/\ SLAM        & /       \\

                                                                              & \begin{tabular}[c]{@{}l@{}}MVSEC\\\citep{zhu2018multivehicle} \end{tabular}
                                                                              & \CIRCLE                                      & \CIRCLE                                        & \CIRCLE                           & \CIRCLE                          & \CIRCLE        & \CIRCLE      & \Circle           & \Circle         & \CIRCLE         & \CIRCLE          & \Circle        & \CIRCLE         & GNSS/\ MoCap/\ SLAM    & SLAM    \\

                                                                              & \begin{tabular}[c]{@{}l@{}}Rosario\\\citep{pire2019rosario} \end{tabular}
                                                                              & \CIRCLE                                      & \CIRCLE                                        & \Circle                           & \Circle                          & \CIRCLE        & \Circle      & \CIRCLE           & \Circle         & \Circle         & \Circle          & \CIRCLE(V)     & \Circle         & RTK-GNSS               & /       \\

                                                                              & \begin{tabular}[c]{@{}l@{}}M2DGR\\\citep{yin2021m2dgr} \end{tabular}
                                                                              & \CIRCLE                                      & \CIRCLE                                        & \CIRCLE                           & \CIRCLE                          & \CIRCLE        & \Circle      & \RIGHTcircle      & \Circle         & \Circle         & \CIRCLE          & \CIRCLE        & \Circle         & RTK-GNSS/\ MoCap/\ LT  & /       \\

                                                                              & \begin{tabular}[c]{@{}l@{}}Nebula\\\citep{reinke2022locus} \end{tabular}
                                                                              & \CIRCLE                                      & \Circle                                        & \Circle                           & \CIRCLE                          & \CIRCLE        & \Circle      & \CIRCLE           & \RIGHTcircle(O) & \Circle         & \Circle          & \CIRCLE(S)     & \Circle         & SLAM                   & Scanner \\

                                                                              & \begin{tabular}[c]{@{}l@{}}FusionPortable\\\citep{jiao2022fusionportable} \end{tabular}
                                                                              & \CIRCLE                                      & \CIRCLE                                        & \CIRCLE                           & \CIRCLE                          & \CIRCLE        & \CIRCLE      & \RIGHTcircle      & \RIGHTcircle    & \Circle         & \CIRCLE          & \CIRCLE        & \Circle         & MoCap/\ RTK-GNSS/\ NDT & Scanner \\

                                                                              & \begin{tabular}[c]{@{}l@{}}M3ED\\\citep{chaney2023m3ed} \end{tabular}
                                                                              & \CIRCLE                                      & \CIRCLE                                        & \CIRCLE                           & \CIRCLE                          & \CIRCLE        & \Circle      & \Circle           & \RIGHTcircle    & \CIRCLE         & \CIRCLE          & \CIRCLE(V)     & \CIRCLE         & RTK-GNSS/\ SLAM        & SLAM    \\

    \midrule[0.03cm]
                                                                              & \textbf{Ours (FusionPortableV2)}
                                                                              & \CIRCLE                                      & \CIRCLE                                        & \CIRCLE                           & \CIRCLE                          & \CIRCLE        & \CIRCLE      & \CIRCLE           & \CIRCLE         & \CIRCLE         & \CIRCLE          & \CIRCLE        & \CIRCLE         & RTK-GNSS/\ LT          & Scanner \\
    \bottomrule[0.03cm]
  \end{tabular}
  \label{tab:related_work_dataset}
  \vspace{-0.2cm}
\end{table*}

In the last decade, the availability of high-quality datasets has significantly accelerated the development of SOTA SLAM algorithms by reducing the time and cost associated with data acquisition and algorithm evaluation. The rapid progress in sensor and robotics technology has led to the widespread adoption of multiple sensors across various robotic platforms. This evolution has set new benchmarks and hastened the enhancement of SOTA algorithms, spanning both handcrafted and data-driven methods such as VINS-Mono \citep{qin2018vins}, FAST-LIO2 \citep{xu2021fast}, VILENS \citep{wisth2022vilens}, DROID-SLAM \citep{teed2021droid}, and Gaussian Splatting SLAM \citep{matsuki2024gaussian}.

Recent advancements in the SLAM field have also extended to areas such as place recognition and collaborative SLAM. Although these areas are not the primary focus of our work, they contribute significantly to the broader SLAM research community.
Datasets such as Wild-Places \citep{knights2023wild} and HeLiPR \citep{jung2023helipr} are specifically designed for LiDAR-based place recognition, emphasizing large-scale, long-term localization under varying appearance conditions \citep{yin2024general}.
Collaborative SLAM, which enables information sharing and cooperative mapping among multiple robots, has gained attention with datasets such as Kimera-Multi \citep{tian2023resilient}, GrAco \citep{zhu2023graco}, and S3E \citep{feng2022s3e}. These datasets provide multi-sensor data (e.g., cameras and LiDARs) and benchmarking tools, which are useful for both collaborative and single-robot SLAM evaluation. However, as they focus primarily on multi-robot scenarios with identical platforms, their utility for exploring cross-platform variability and single-robot SLAM challenges is limited.

In contrast, our FusionPortableV2 dataset is primarily designed for short-term odometry and SLAM, with a focus on accurately tracking the robot's pose and constructing consistent maps within a single session or over a shorter time period.
The dataset features diverse platforms, sensor configurations, and environments, making it better suited for studying SLAM generalization across different conditions compared to most related works, as detailed in Table \ref{tab:related_work_dataset}.

\subsection{Specific-Platform Datasets}
\label{sec:specfic-platform}

Early SLAM datasets predominantly focused on visual-inertial fusion, targeting specific platforms and environments. This focus was largely due to the ubiquity and convenience of visual and inertial sensors which are cheap and lightweight.
They cover sequences which were captured by different platforms ranging from handheld devices \citep{pfrommer2017penncosyvio, schubert2018tum, zuniga2020vi}, drones \citep{burri2016euroc,majdik2017zurich,delmerico2019we,li2024mars}, unmanned ground vehicles \citep{pire2019rosario}, and aquatic vehicles \citep{miller2018visual}, respectively.
Notably, the UZH-FPV dataset \citep{delmerico2019we} stands out for its integration of event cameras and the inclusion of rapid trajectories from aggressive drone flights.

Concurrently, in the automotive industry, urban environment datasets introduce specific challenges including adverse lighting, weather conditions, and larger scales.
Long-range sensors such as LiDARs and Radars are preferred for their capabilities, even though they were initially bulky and costly.
The KITTI dataset \citep{geiger2013vision} sets a benchmark in autonomous driving with its rich urban sensor data collection.
Further developments in driving-related datasets have expanded across dimensions of duration \citep{maddern20171}, urban complexity \citep{jeong2019complex}, and weather adversity \citep{agarwal2020ford}.
The DSEC dataset \citep{gehrig2021dsec}, akin to UZH-FPV, leverages stereo event cameras for extensive driving scenes.
Moreover, Radars are essential for outdoor perception, offering advantages in range, velocity measurement via the Doppler effect and weather resilience.
Related datasets such as Boreas \citep{burnett2023boreas}, Oxford Radar RoboCar \citep{barnes2020oxford}, and OORD \citep{gadd2024oord} collected data under conditions such as fog, rain, and snow.

The trend towards multi-sensor fusion spurred the creation of diverse and complex datasets.
Datasets such as NCLT \citep{carlevaris2016university}, M2DGR \citep{yin2021m2dgr}, NTU-VIRAL \citep{nguyen2022ntu}, ALITA \citep{yin2022alita}, and FusionPortable \citep{jiao2022fusionportable} also pose challenges for SLAM, given their diverse environmental appearance and structure.
These datasets feature a variety of environments, including dense vegetation, open spaces, and complex buildings with multiple levels and detailed layouts.
The changing lighting, seasonal foliage variations, and movement of pedestrians and vehicles add complexity to campus environments. As highlighted in NCLT dataset, these factors are crucial for life-long SLAM challenges.
However, these datasets, collected via specific platforms such as unmanned ground vehicles (UGVs) and drones, fall short in showcasing diverse motion patterns, especially aggressive maneuvers.

Recent advancements in sensor technology have enabled the development of portable multi-sensor suites capable of collecting high-quality, multi-modal data in diverse environments (e.g., multi-floor buildings). Notable examples include the Newer College \citep{ramezani2020newer} and Hilti-Oxford \citep{zhang2022hilti} datasets. These datasets also offer dense, high-quality 3D global maps, enabling the generation of high-rate $6$-DoF reference poses and the evaluation of mapping algorithms.
The RELLIS-3D dataset \citep{jiang2021rellis}, though it primarily focuses on semantic scene understanding in off-road environments, also provides diverse sensor data that can be valuable for SLAM research.

\subsection{Cross-Platform Datasets}
\label{sec:cross-platform}
As SLAM research progressed from single-platform or collaborative applications, there has been a growing interest in cross-platform generalization.
Building upon the specific-platform datasets discussed in Section \ref{sec:specfic-platform}, several datasets explore the generalization of algorithms across different platforms and scales, aiming to integrate motion characteristics from varied platforms with minimal parameter tuning for diverse scenarios.
The MVSEC dataset \citep{zhu2018multivehicle} collected multi-sensor data with diverse platforms, excluding UGV sequences.
Conversely, the Nebula dataset \citep{reinke2022locus}, developed during the DARPA Subterranean Challenge, includes field environments with both wheeled and legged robots, providing precise maps and trajectories. However, it lacks urban data and primarily focuses on LiDAR-based perception.
The M3ED dataset \citep{chaney2023m3ed}, although closely aligned with our objectives, lacks indoor data and platform-specific kinematic measurements, underscoring the unique contribution of our dataset.

Despite these advancements, there remains an absence of datasets that comprehensively cover a wide range of platforms, environments, and sensor modalities within a single, unified framework.

%%%%%%%%%%%%%%%%%%%%%%%%%%%%%%%%%%%%%%%%%%%
\begin{table*}[t]
  \centering
  \caption{The sensors used in this dataset and their corresponding specifications.
    The detailed definition of ROS message type and naming of coordinate frames of each sensor are provided in the dataset website.}
  \renewcommand\arraystretch{0.9}
  \renewcommand\tabcolsep{4pt}
  \scriptsize
  % \footnotesize
  % \resizebox{\linewidth}{!}{
  \begin{tabular}{llllr}
    \toprule
    Sensor                                      & Characteristics                                     & ROS Topic                                       & ROS Message Type                    & Rate (Hz) \\
    \midrule[0.03cm]

    \multirow{3}{*}{3D LiDAR}                   &
    \begin{tabular}[l]{@{}l@{}}
      Ouster OS$1$-$128$, $45^{\circ}$vert.$\times$ $360^{\circ}$horiz. FOV \\
    \end{tabular}
                                                & \texttt{/os\_cloud\_node/points}                    & \text{sensor\_msgs/PointCloud2}                 & $10$                                            \\
                                                &
    IMU: ICM$20948$, $9$-axis MEMS              & \texttt{/os\_cloud\_node/imu}                       & \text{sensor\_msgs/Imu}                         & $100$                                           \\
                                                &
    Range, near-ir, reflectivity, signal images & \texttt{/os\_image\_node/(range,nearir,...)\_image} & \text{sensor\_msgs/Image}                       & $10$                                            \\
    \midrule[0.01cm]

    \multirow{1}{*}{Frame Camera}               &
    \begin{tabular}[l]{@{}l@{}}
      Stereo FILR BFS-U$3$-$31$S$4$C, global shutter        \\
      $66.5^{\circ}$vert. $\times$ $82.9^{\circ}$horiz. FOV \\
      $1024\times 768$ resolution                           \\
    \end{tabular}
                                                & \texttt{/stereo/frame\_(left,right)/image\_raw}     & sensor\_msgs/CompressedImage                    & $20$                                            \\
    \midrule[0.01cm]

    \multirow{3}{*}{Event Camera}               &
    \begin{tabular}[l]{@{}l@{}}
      Stereo DAVIS$346$, $67^{\circ}$vert., $\times$ $83^{\circ}$horiz. FOV \\
      $346\times 240$ resolution                                            \\
    \end{tabular}
                                                & \texttt{/stereo/davis\_(left,right)/events}         & \text{dvs\_msgs/EventArray}                     & $30$                                            \\
                                                & Images that capture color data                      & \texttt{/stereo/davis\_(left,right)/image\_raw} & \text{sensor\_msgs/CompressedImage} & $20$      \\
                                                & IMU: MPU$6150$, $6$-axis MEMS                       & \texttt{/stereo/davis\_(left,right)/imu}        & \text{sensor\_msgs/Imu}             & $1000$    \\
    \midrule[0.01cm]

    \multirow{1}{*}{IMU}                        & STIM$300$, $6$-axis MEMS
    % , bias instability $0.3^{\circ}/h$, Allan variance $25^{\circ}C$ 
                                                & \texttt{/stim300/imu}                               & sensor\_msgs/Imu                                & $200$                                           \\
    \midrule[0.01cm]

    \multirow{3}{*}{INS}                        &
    \multirow{3}{*}{\begin{tabular}[l]{@{}l@{}}
        $3$DM-GQ$7$-GNSS/INS \\
        Dual-antenna, RTK-enabled INS
      \end{tabular}}
    % $4$ concurrent GNSS, L$1$/L$2$/L$5$ RTK 
                                                & \texttt{/3dm\_ins/nav/odom}                         & nav\_msgs/Odometry                              & $10$                                            \\
                                                &                                                     & \texttt{/3dm\_ins/gnss\_(left,right)/fix}       & \text{sensor\_msgs/NavStatFix}      & $10$      \\
                                                &                                                     & \texttt{/3dm\_ins/imu}                          & \text{sensor\_msgs/Imu}             & $200$     \\
    \midrule[0.01cm]

    \multirow{1}{*}{Wheel Encoder}              & Omron E6B2-CWZ6C, $1000$P/R
    % , Incremental measurement 
                                                & \texttt{/mini\_hercules/encoder}                    & \text{sensor\_msgs/Joinstate}                   & $100$                                           \\
    \midrule[0.01cm]

    \multirow{3}{*}{Legged Sensor}
                                                & Built-in joint encoders and contact sensors         & \texttt{/unitree/joint\_state}                  & \text{sensor\_msgs/JointState}      & $50$      \\
                                                & Built-in IMU                                        & \texttt{/unitree/imu}                           & \text{sensor\_msgs/Imu}             & $50$      \\
                                                & Out-of-the-box kinematic-ineratial odometry         & \texttt{/unitree/body\_odom}                    & \text{nav\_msgs/Odometry}           & $50$      \\
    \bottomrule[0.03cm]
  \end{tabular}
  \label{tab:sensor_list}
  \vspace{-0.4cm}
\end{table*}

\section{System Overview}
\label{sec:system_overview}
\runninghead{Wei \textit{et~al.}}

%%%%%%%%%%%%%%%%%%%%%%%%%%%%%%%%%%%%%%%%%%
\begin{figure}[]
    \centering
    \includegraphics[width=0.44\textwidth]{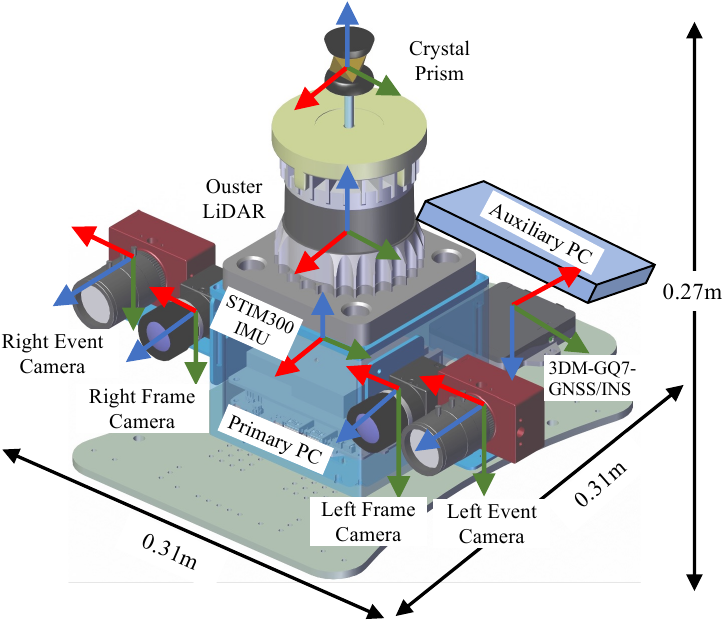}
    \caption{CAD model of the sensor rig where axes are marked:
        \textcolor{red}{red}: $X$, \textcolor{green}{green}: $Y$, \textcolor{blue}{blue}: $Z$.
        It visualizes the position of each component of the handheld multi-senosr suite.}
    \label{fig:device_layout}
    \vspace{-0.4cm}
\end{figure}
%%%%%%%%%%%%%%%%%%%%%%%%%%%%%%%%%%%%%%%%%%

This section presents our developed multi-sensor suite, designed for integration with various mobile platforms through plug-and-play functionality. All sensors are securely mounted on a aluminum alloy frame, facilitating a unified installation.
Additionally, we detail the devices employed for collecting GT trajectories and maps.

\subsection{Suite Setup and Synchronization}
\label{sec:suite_setup_and_sync}
The Multi-Sensor Suite (MSS) integrates exteroceptive and proprioceptive sensors, including a 3D Ouster LiDAR, stereo frame and event cameras, and IMUs, as depicted in its CAD model in Fig. \ref{fig:device_layout}.
We use two PCs that are synchronized via a Network Time Protocol (NTP) server for data collection.
The primary PC processes data from the frame cameras and IMU, while the auxiliary PC handles additional data types. Both PCs are equipped with a 1TB SSD, 64GB of DDR4 memory, and an Intel i7 processor, running Ubuntu with a real-time kernel patch and employing the Robot Operating System (ROS) for data collection.
This distributed architecture reduces the number of ROS nodes running on each individual PC, thereby alleviating the risk of queuing problems during data collection.
After the mission, separate data collected by these two PCs are merged and postprocessed offline.
The subsequent sections will elaborate on the synchronization approach and the features of each sensor.

\begin{figure}[t]
    \centering
    \subfigure[Illustration of data flow and synchronization]{
        \includegraphics[width=0.42\textwidth]{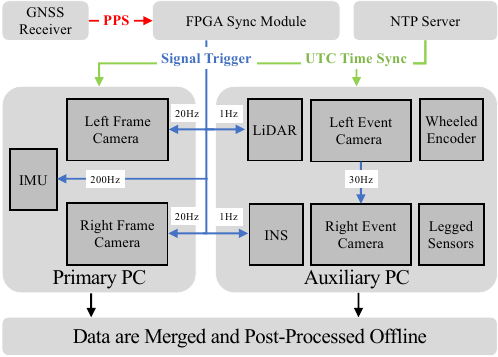}
        \label{fig:block_diagram1}
    }
    \subfigure[Synchronization reduces delay in $t_{delay}$ compared with $t_{offset}$]{
        \includegraphics[width=0.49\textwidth]{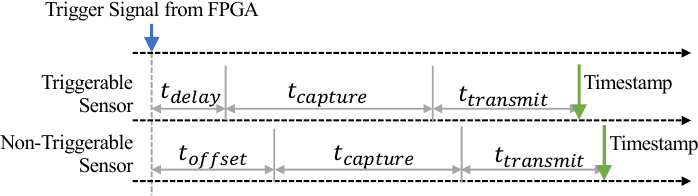}
        \label{fig:device_timing}
    }
    \caption{(a) Illustration of data collection which shows the data flow and synchronization processes. The \textcolor{red}{red} arrow indicate PPS signals for synchronization, \textcolor{green}{green} arrows show UTC time synchronization, and \textcolor{blue}{blue} arrows represent sensor triggering signals, and \textcolor{black}{black} arrows depict the flow of raw data. (b) The timing diagram for triggerable (our case) and non-triggerable sensors, illustrating the unknown time offset caused by the delay in starting data capture, the duration of data capture, and the time required for data transmission from the sensor to the PC. Our synchronization solution can reduce the time delay (i.e., $t_{offset} - t_{delay}$) but cannot address other factors, which require online time calibration algorithms.}
    \label{fig:block_diagram}
\end{figure}

\subsubsection{\textbf{Synchronization:}}
\label{sec:synchronization}
The synchronization process is illustrated in Fig. \ref{fig:block_diagram}.
Generally, the field-programmable gate array (FPGA) board synchronizes with the pulse-per-second (PPS) signal from the external GNSS receiver, producing higher frequency trigger signals for the IMU, stereo frame cameras, and LiDAR clock alignment.
In GPS-denied environments, it utilizes its internal low drift oscillator for synchronization, achieving a time accuracy below $1ms$ between multiple trigger signals.
To synchronize LiDAR and camera data, we phase-lock\endnote{\url{https://static.ouster.dev/sensor-docs/image_route1/image_route2/time_sync/time-sync.html}} the LiDAR's rotation so its forward-facing direction aligns with the camera's capture timing, accounting for the continuous nature of LiDAR's spinning data.
We use the internal (Master-Slave mode) mechanism to synchronize stereo event cameras, where the left event camera is assigned as the master to send trigger signals to the right camera.

Fig. \ref{fig:device_timing} illustrates our synchronization scheme. The FPGA sends trigger signals to connected sensors, which starts capturing data after a small delay (commonly $t_{delay}<1ms$).
$t_{capture}$ is defined as the time for capturing data such as camera exposure or LiDAR rotation.
It is followed by a transmission process that sends data to the host with time $t_{transmit}$.
Non-triggerable sensors such as wheeled encoders use their internal clocks for timestamping, resulting in a time offset, $t_{offset}$, relative to the FPGA's trigger signal.
Our scheme cannot recover $t_{delay}$ and $t_{capture}$.

\begin{figure*}[t]
	\centering
	\subfigure[Top View of the UGV.]{\label{fig:ugv_layout1}\centering\includegraphics[width=.255\linewidth]{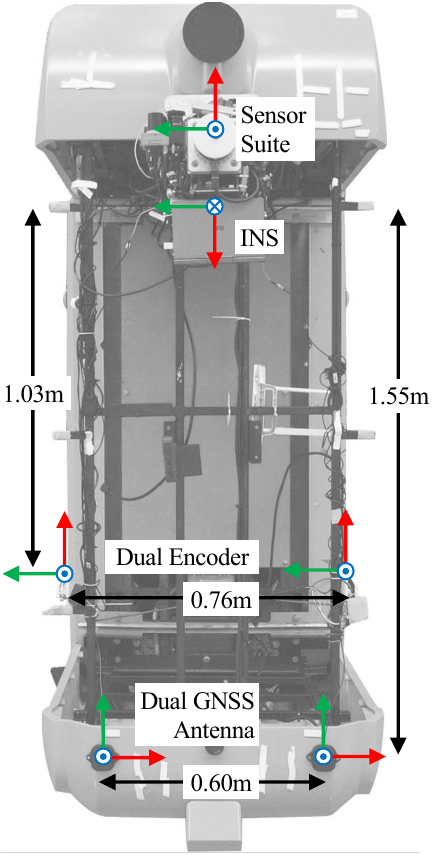}}
	\subfigure[Side View of the UGV.]{\label{fig:ugv_layout2}\centering\includegraphics[width=.635\linewidth]{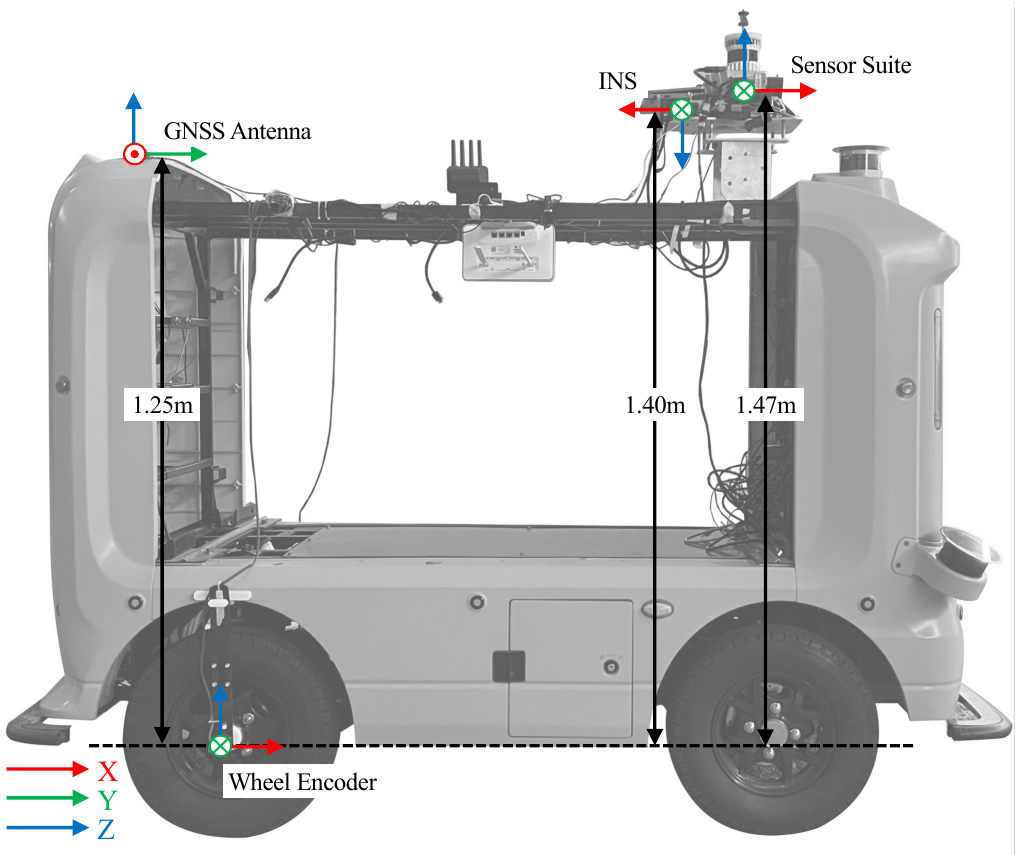}}
	\subfigure[Legged Robot.]{\label{fig:legged_layout}\centering\includegraphics[width=.415\linewidth]{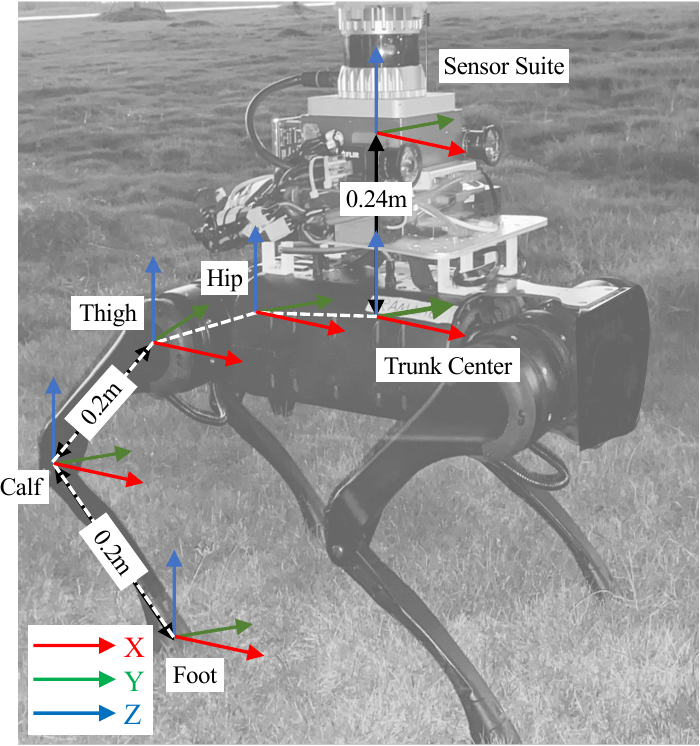}}\hspace{0.3cm}
	\subfigure[High-Speed Vehicle.]{\label{fig:vehicle_layout}\centering\includegraphics[width=.49\linewidth]{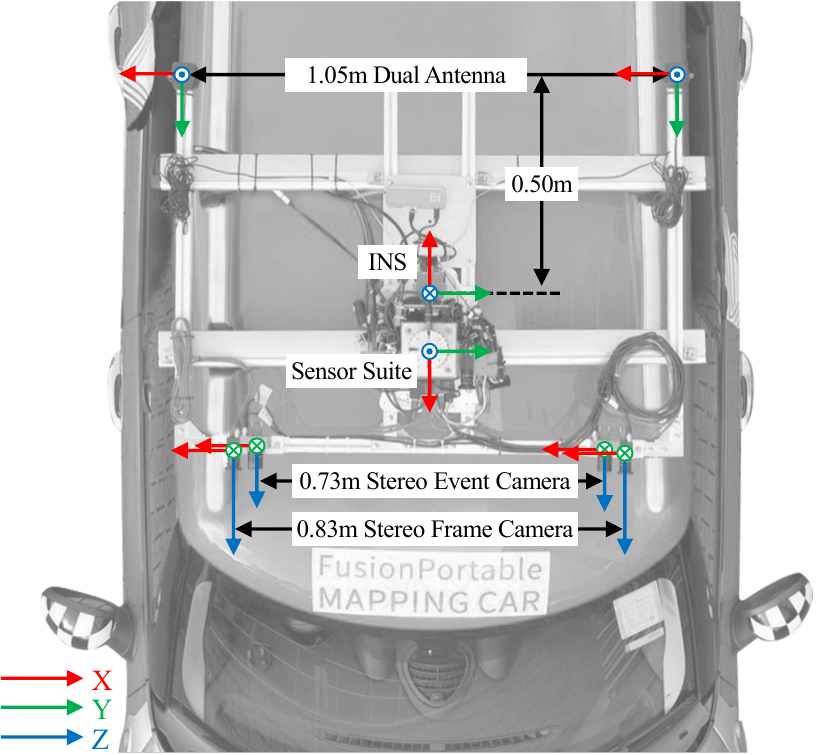}}
	\caption{Layouts of the platform-specific sensor setup, including different coordinate systems and their relative translation. More detailed and accurate dimensional data are provided in our calibration files.}
	\label{fig:layout}
	\vspace{-0.3cm}
\end{figure*}

\subsubsection{\textbf{3D LiDAR:}}
Our LiDAR choose the OS1-$128$ Gen$5$ LiDAR that operates at $10Hz$.
It features a built-in IMU capturing gyroscope, acceleration, and magnetometer data
and generates four types of images to facilitate the usage of image-based algorithms: \textbf{range}, \textbf{near-ir}, \textbf{reflectivity}, and \textbf{signal} image.
Each image measures different properties of the surroundings:
(1) range images display the distance of the point from the sensor origin, calculated using the time of flight of the laser pulse;
(2) near-ir images capture the strength of sunlight at the $865nm$ light wavelength collected, also expressed in the number of photons detected that were not produced by the sensor's laser pulse;
(3) reflectivity images display the reflectivity of the surface or object that was detected by the sensor; and
(4) signal images show the strength of the light returned from the given point, which is influenced by various factors including distance, atmospheric conditions, and objects' reflectivity.
The timestamp of the LiDAR data represents the end of the scanning period.

\subsubsection{\textbf{Stereo Frame Cameras:}}
Our setup includes two FLIR BFS-U$3$-$31$S$4$C global-shutter color cameras for stereo imaging, synchronizing to output images at $1024\times 768$ pixels and $20Hz$.
The exposure time $\tau$ for both cameras is manually set as a fixed value.
Image timestamps are adjusted by subtracting $0.5\tau$ to properly align image features with IMU measurements since an image is the result of integrating light over the exposure interval \citep{furgale2013unified,rehder2016general}.
To prevent abrupt changes in color space, the white balance settings are also fixed.
Additionally, metadata like exposure time and gain are included for image analysis.

\subsubsection{\textbf{Stereo Event Cameras:}}
Two event cameras, which are known for their high temporal resolution, extensive dynamic range, and energy efficiency, are used for data collection.
These cameras output events data, $346\times 260$ frame images, and high-rate IMU measurements.
Frame images cannot be synchronized, resulting in a $10$-$20ms$ delay.
Infrared filters are used to lessen LiDAR light interference.
Exposure times are set fixedly, whereas outdoor settings use auto-exposure to maintain image quality under varying light conditions.

\begin{figure*}[t]
  \centering
  \includegraphics[width=0.18\linewidth]{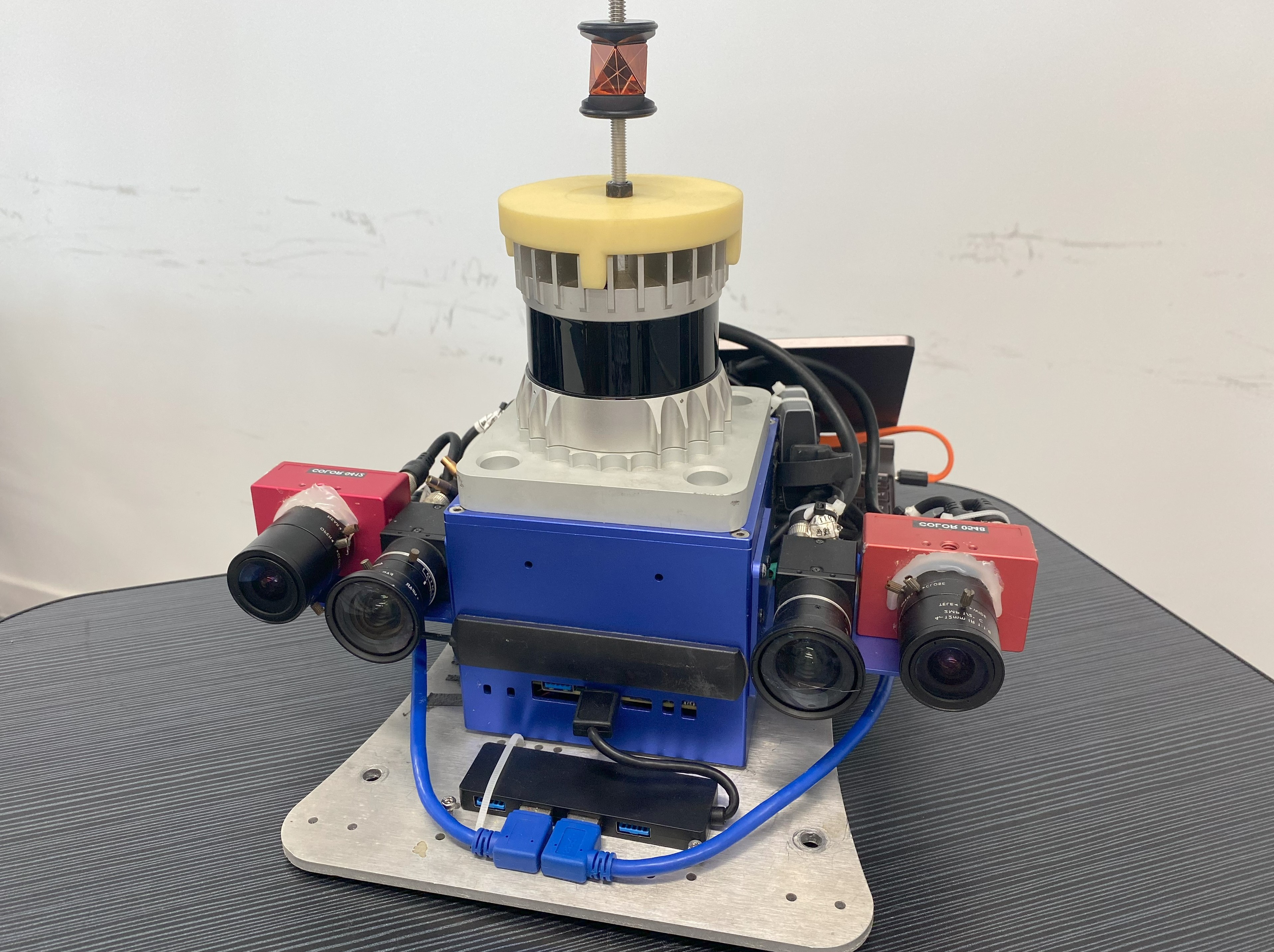}\hfill
  \includegraphics[width=0.18\linewidth]{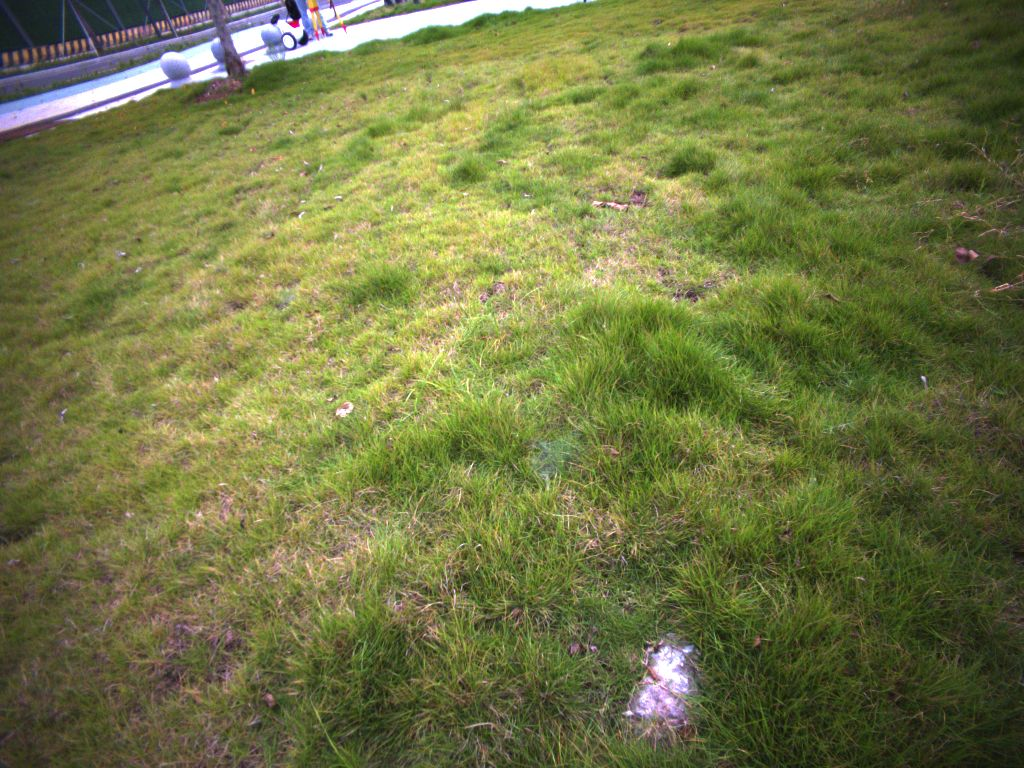}\hfill
  \includegraphics[width=0.18\linewidth]{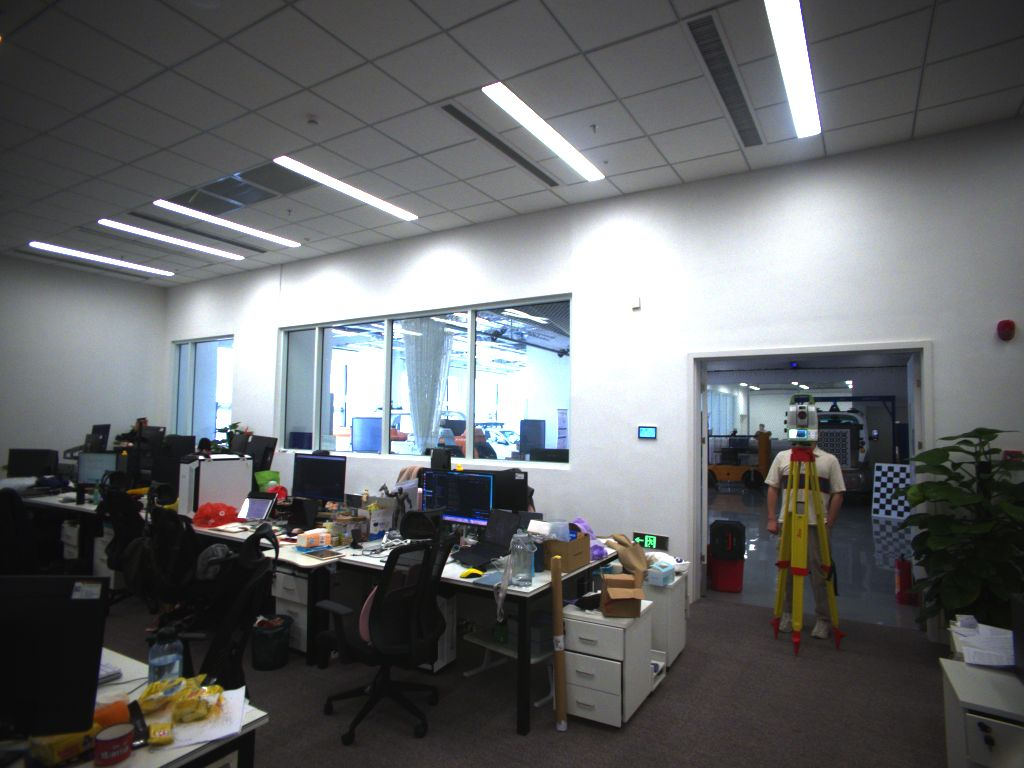}\hfill
  \includegraphics[width=0.18\linewidth]{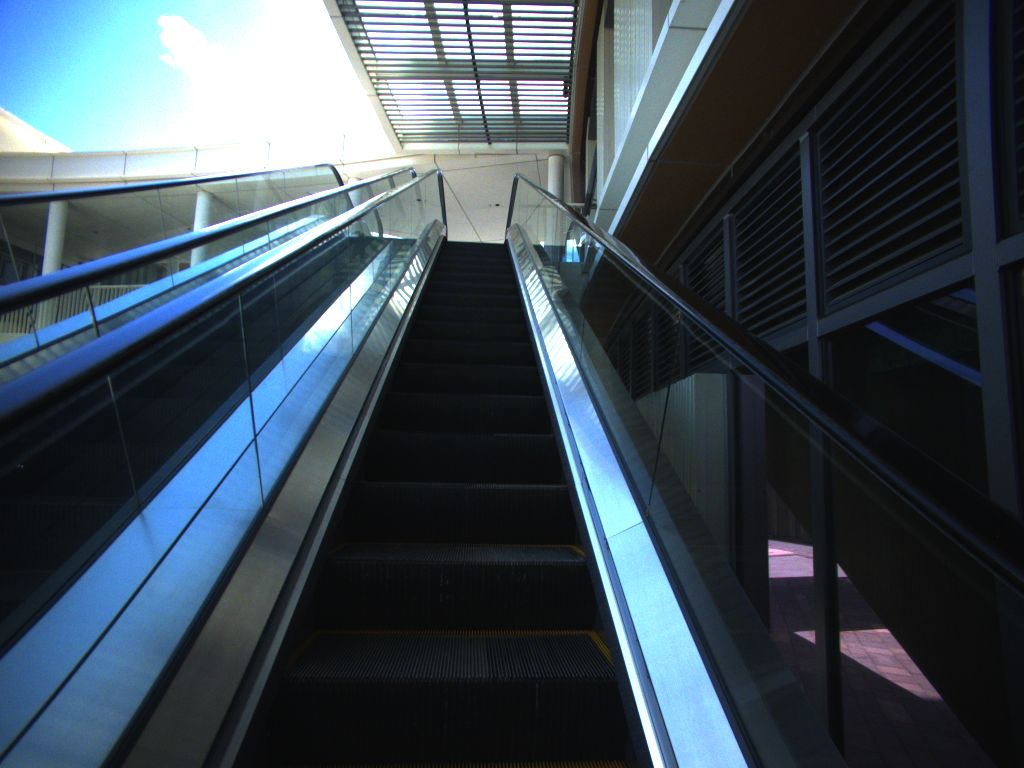}\hfill
  \includegraphics[width=0.18\linewidth]{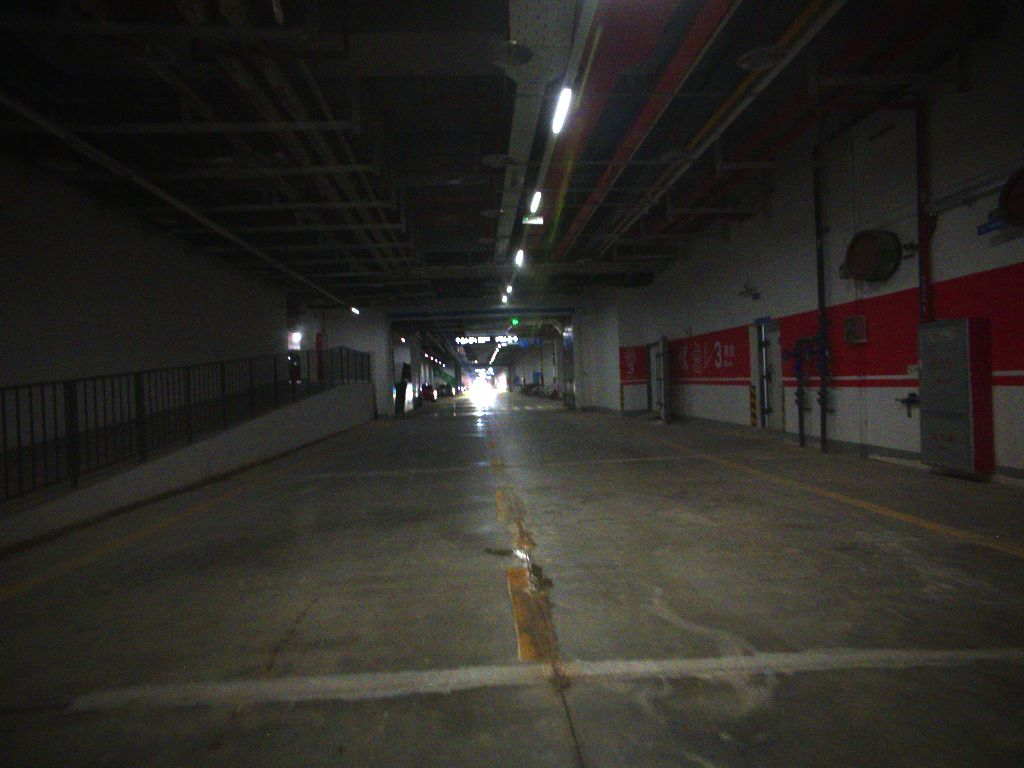}\\
  \centering{(a) The handheld multi-sensor rig and scene images covering the grassland, lab, escalator, and underground tunnel.}\vspace{0.1cm}\\
  \includegraphics[width=0.18\linewidth]{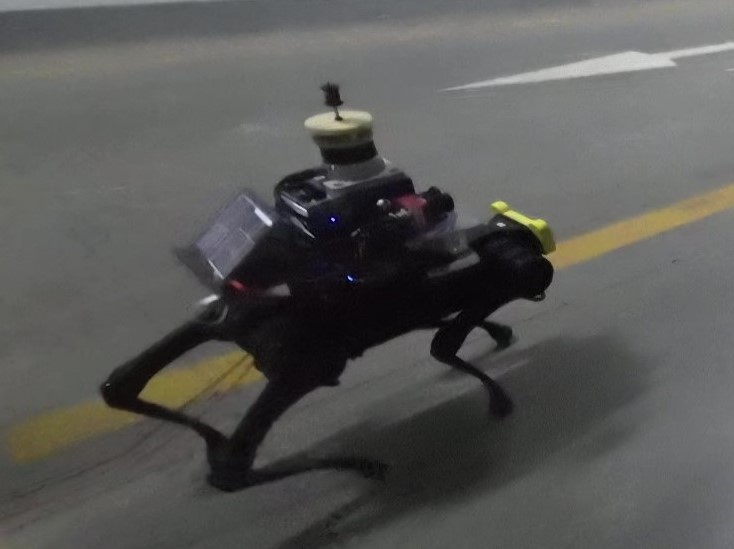}\hfill
  \includegraphics[width=0.18\linewidth]{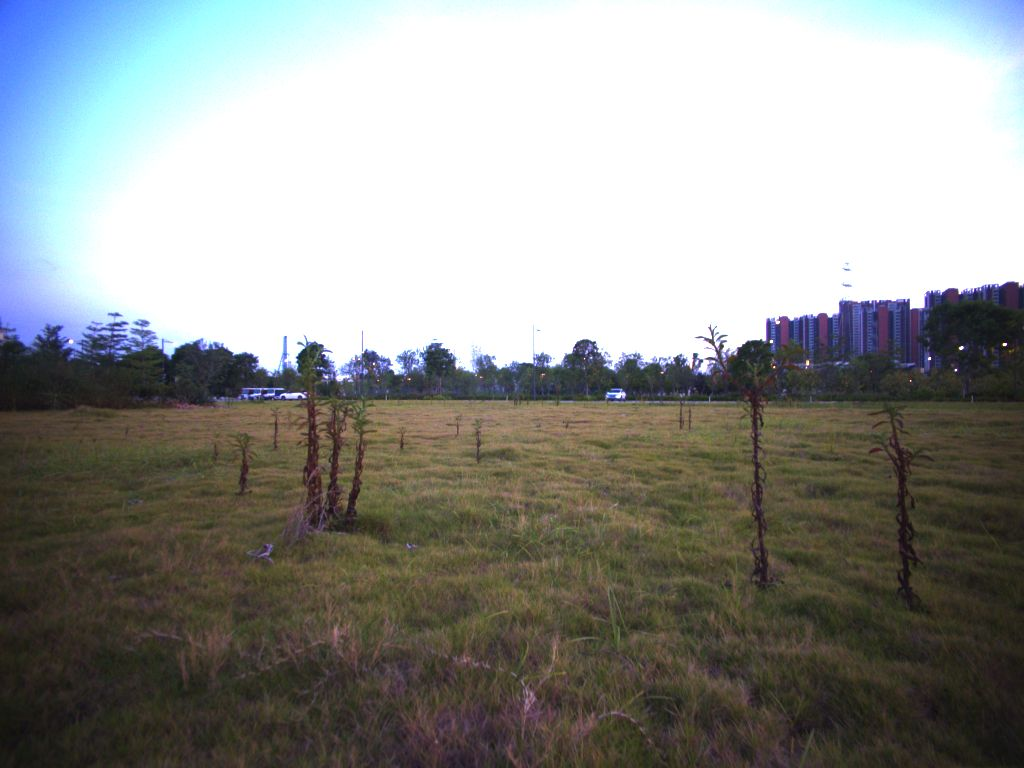}\hfill
  \includegraphics[width=0.18\linewidth]{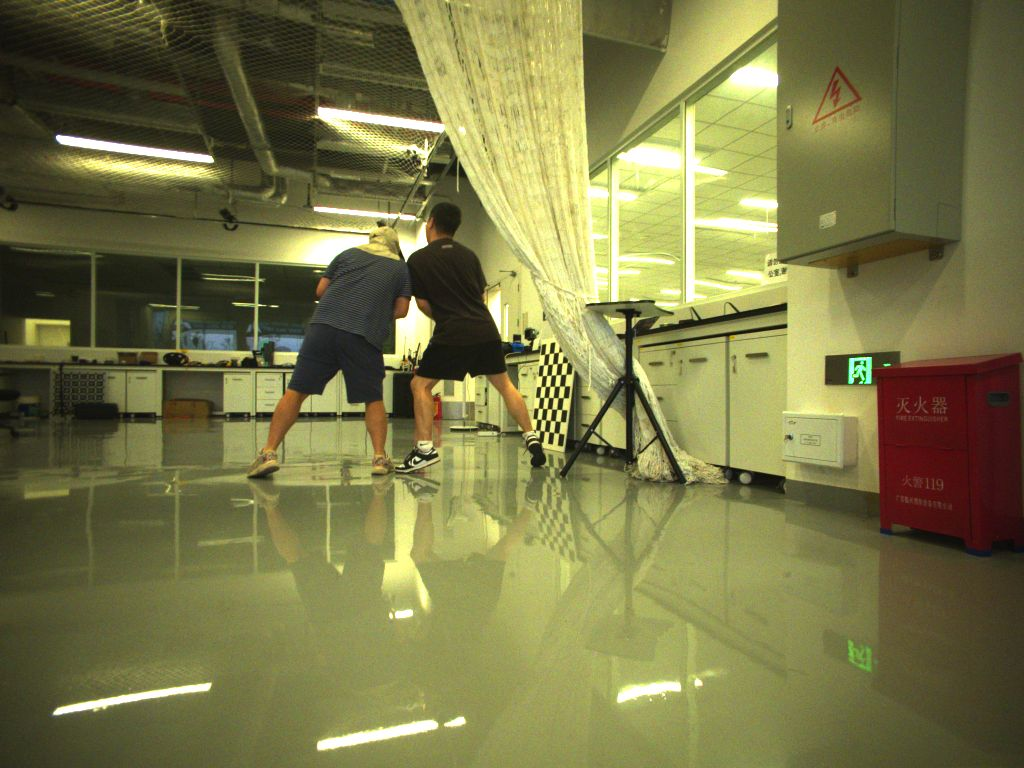}\hfill
  \includegraphics[width=0.18\linewidth]{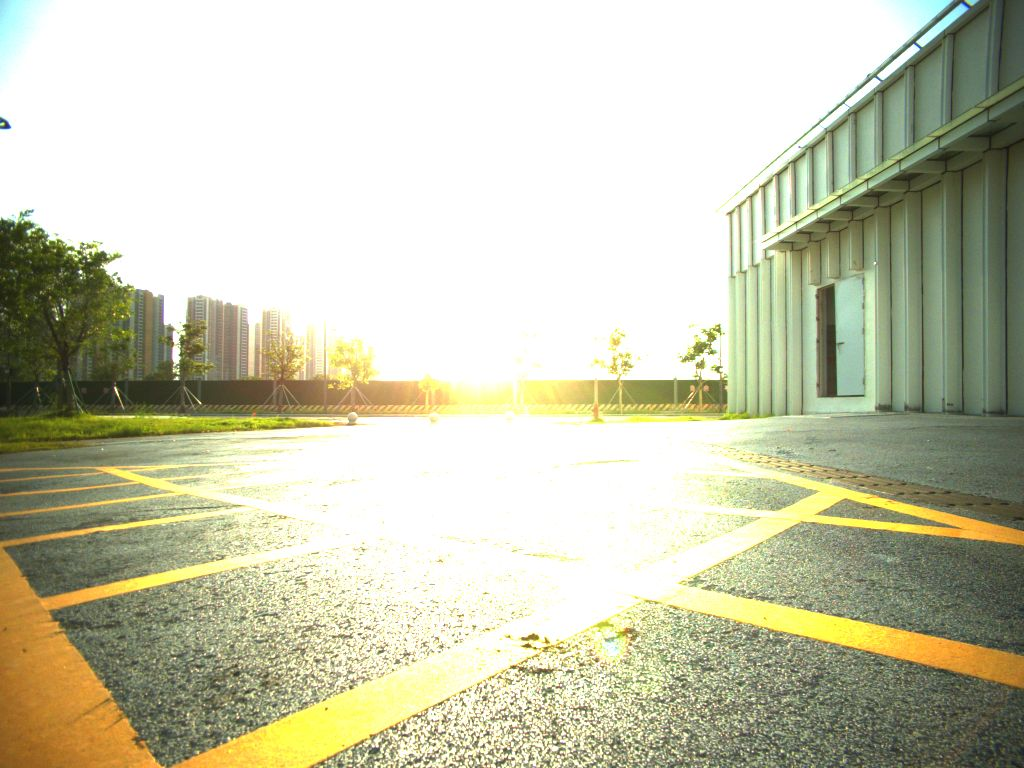}\hfill
  \includegraphics[width=0.18\linewidth]{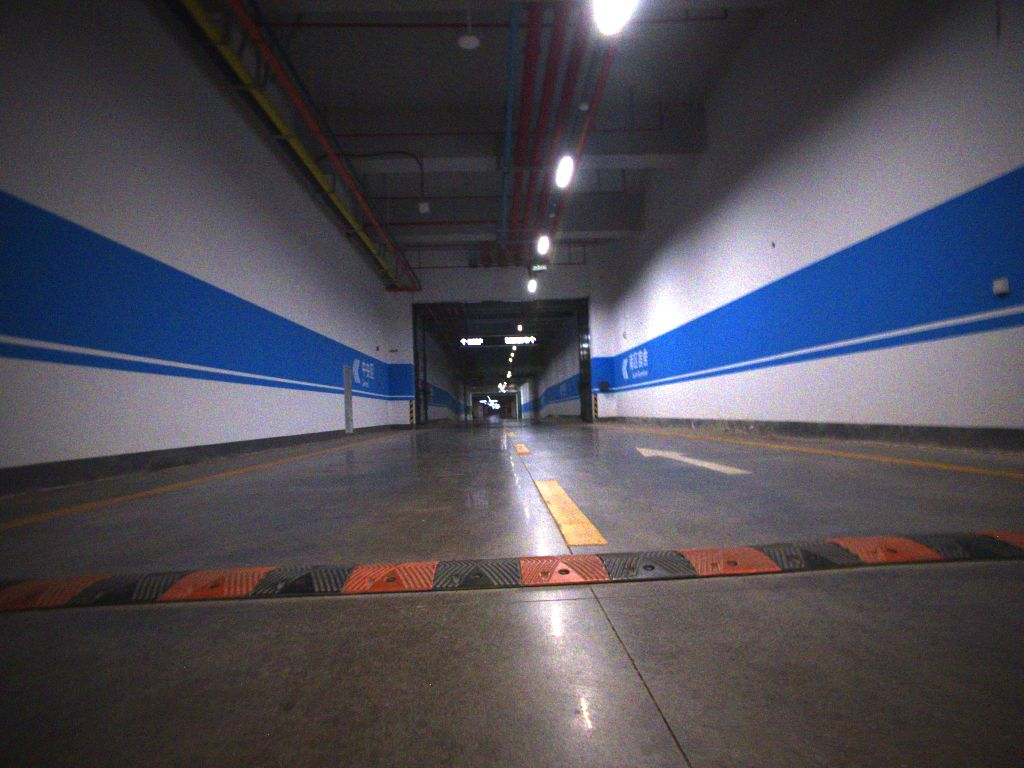}\\
  \centering{(b) The legged robot and scene images covering the grassland, lab, campus, and underground tunnel.}\vspace{0.1cm}\\
  \includegraphics[width=0.18\linewidth]{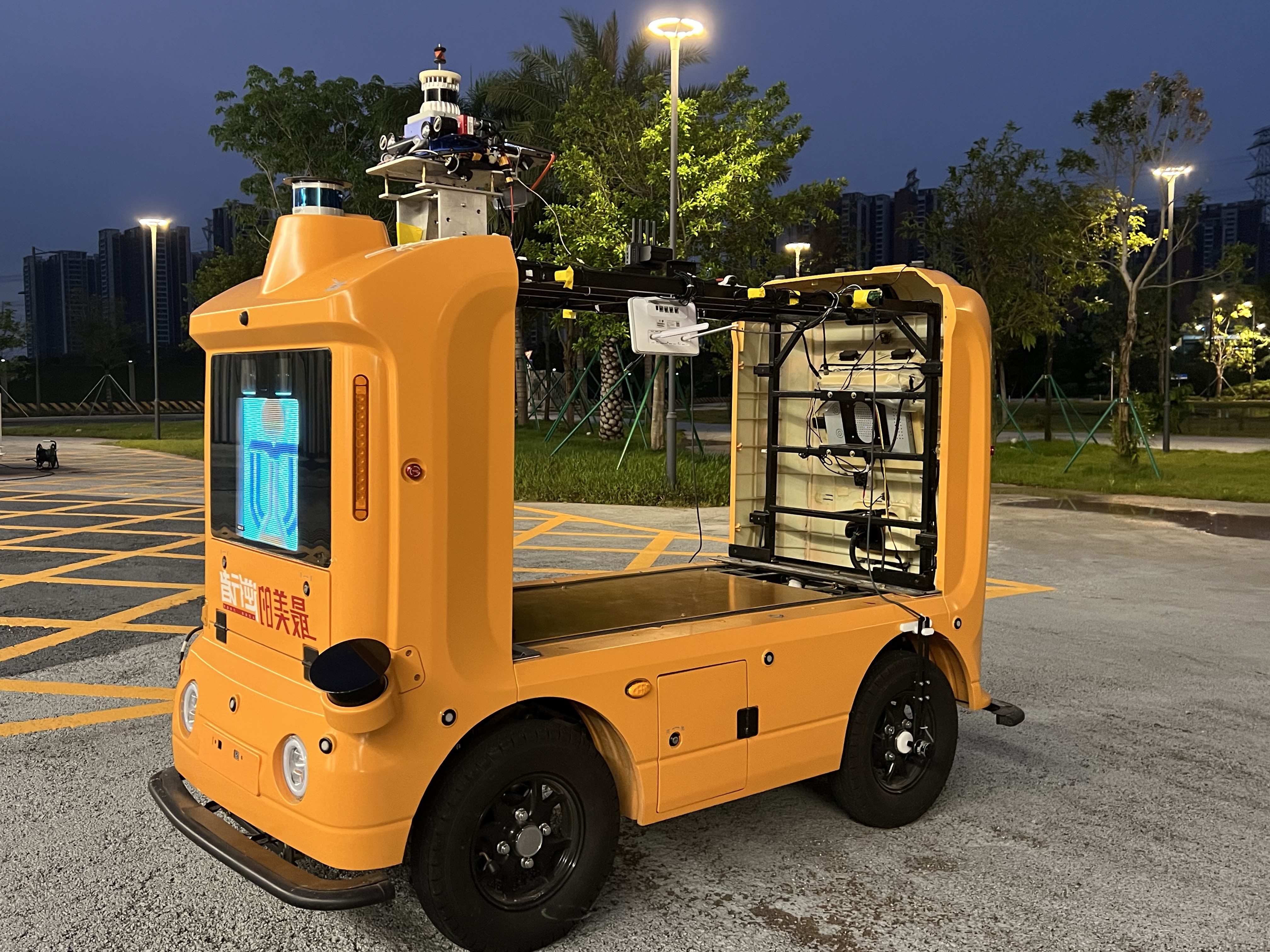}\hfill
  \includegraphics[width=0.18\linewidth]{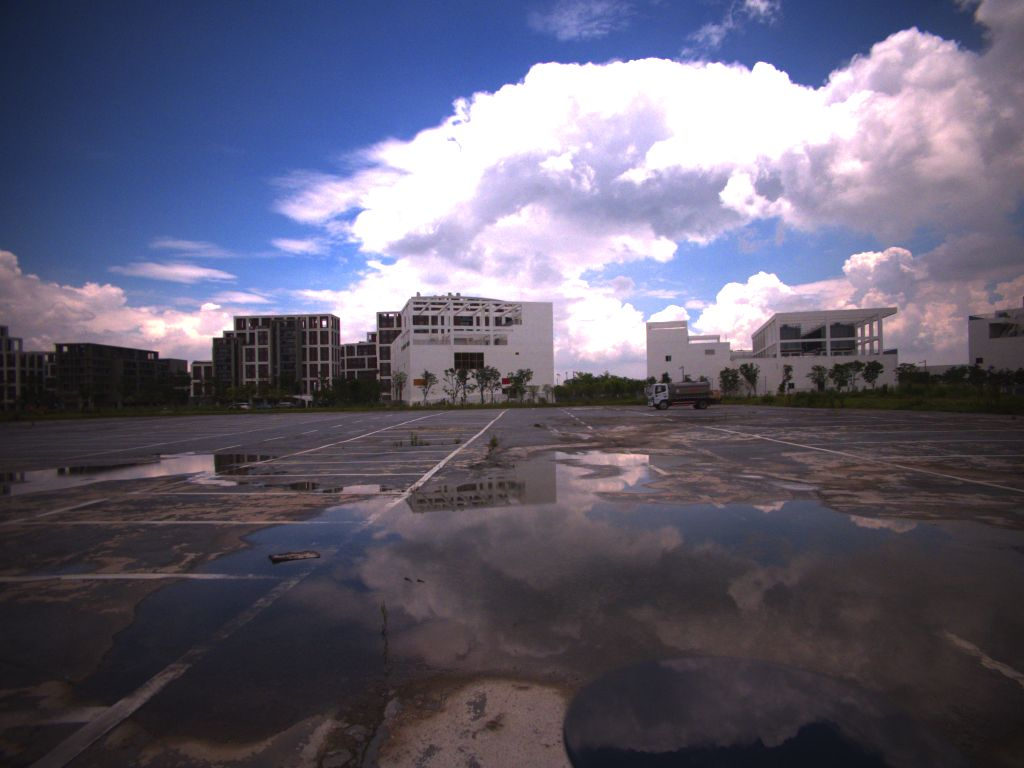}\hfill
  \includegraphics[width=0.18\linewidth]{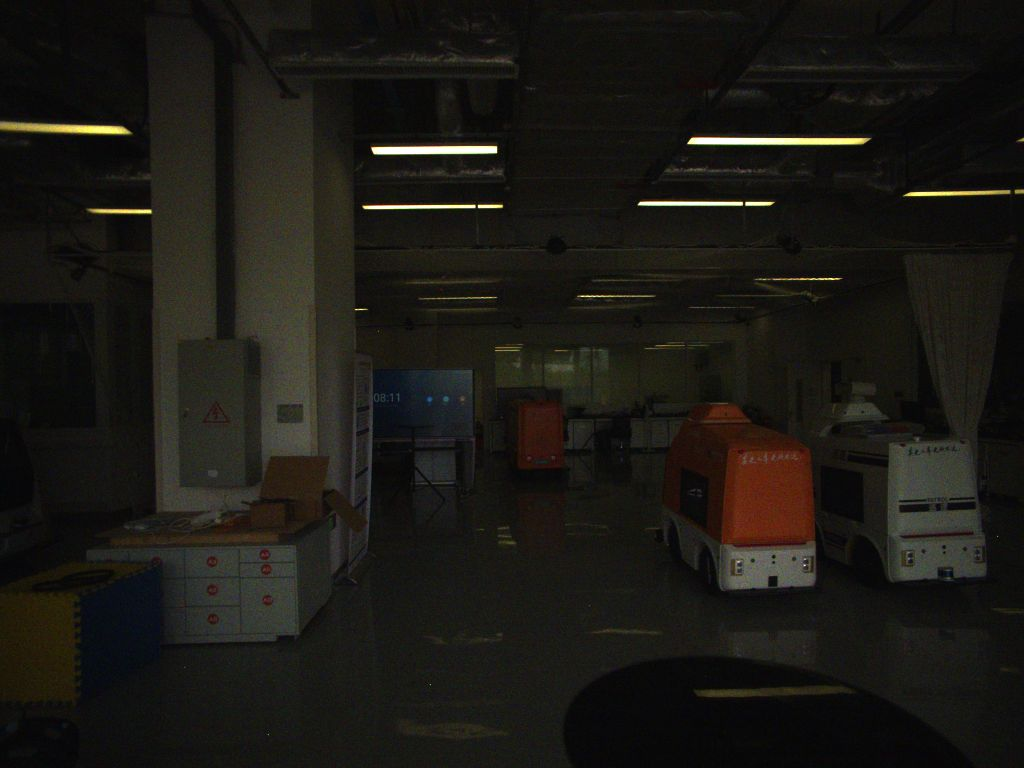}\hfill
  \includegraphics[width=0.18\linewidth]{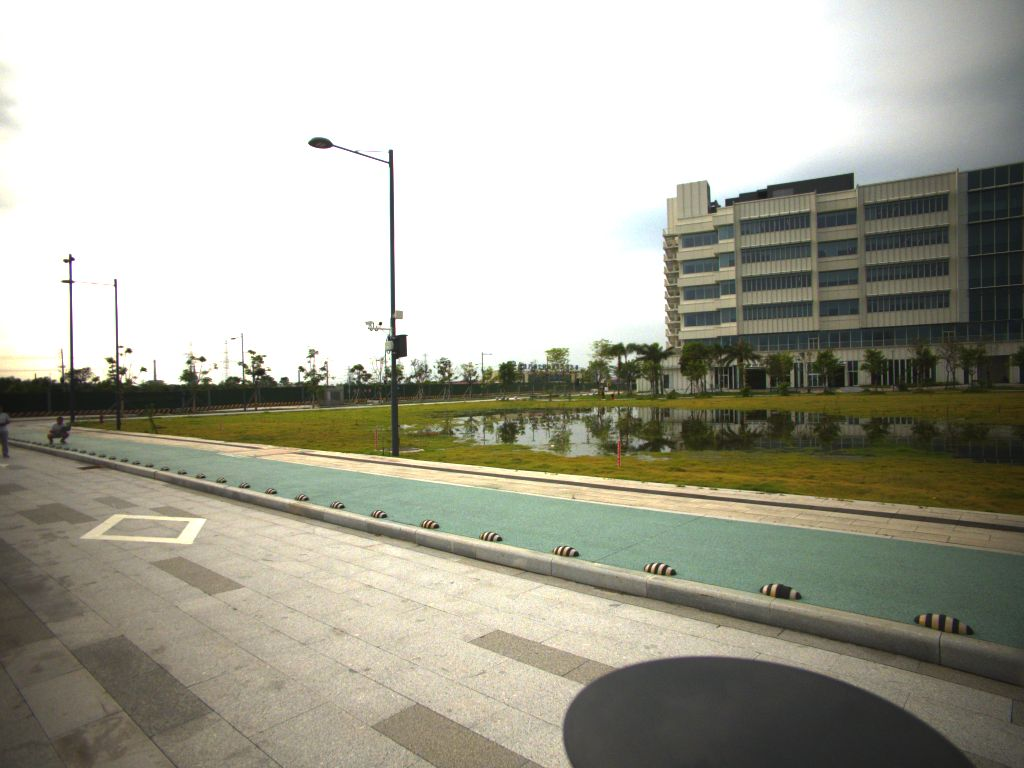}\hfill
  \includegraphics[width=0.18\linewidth]{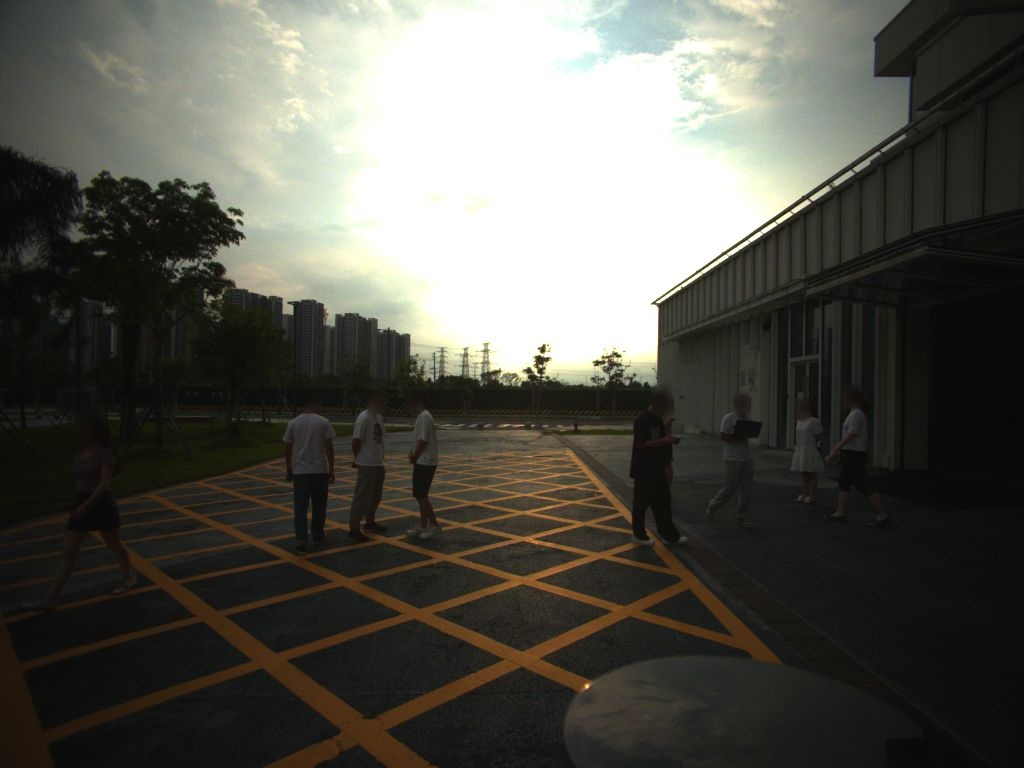}\\
  \centering{(c) The UGV and scene images covering the outdoor parking lot, garage,
    and campus.}\vspace{0.1cm}\\
  \includegraphics[width=0.18\linewidth]{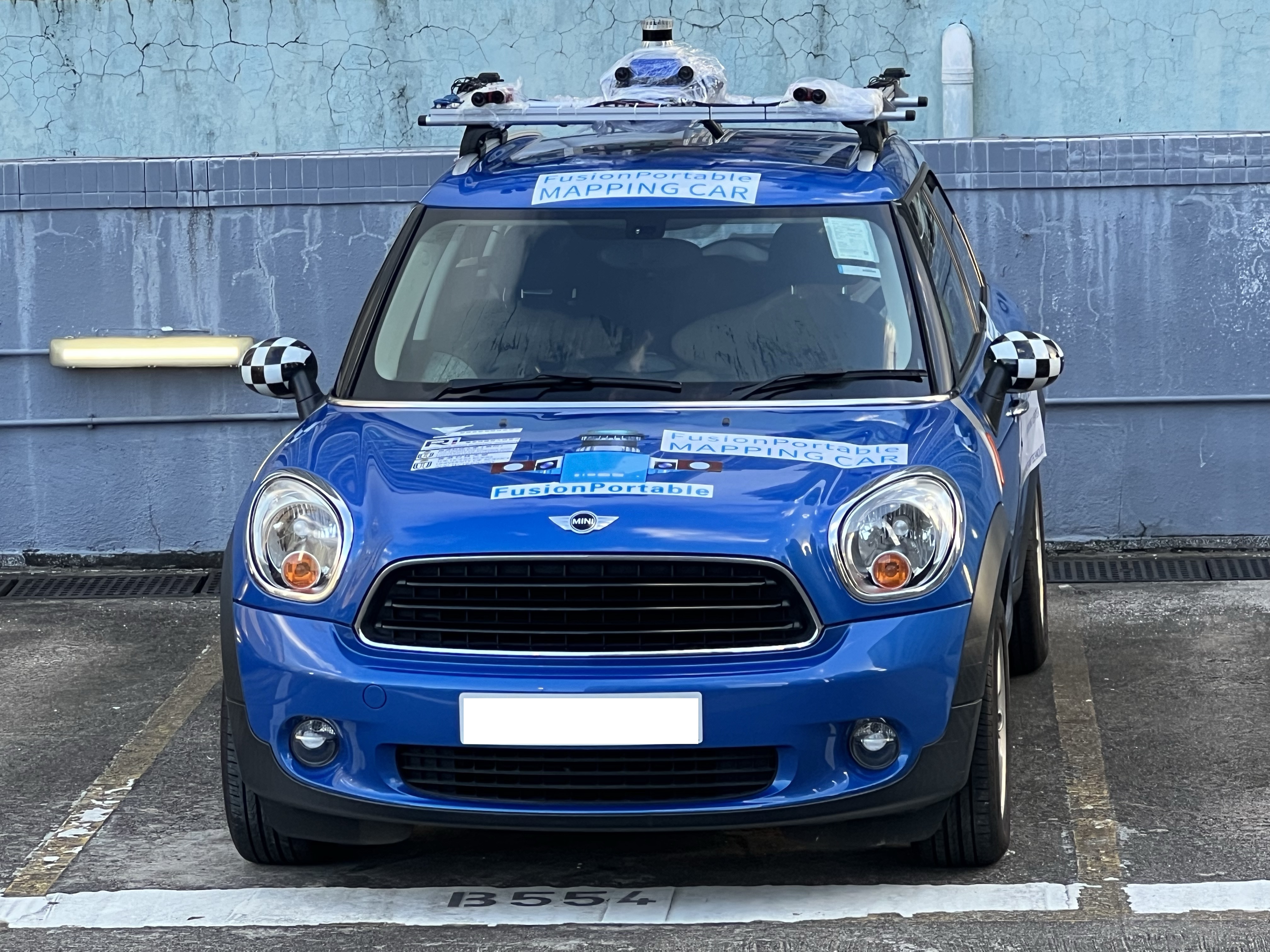}\hfill
  \includegraphics[width=0.18\linewidth]{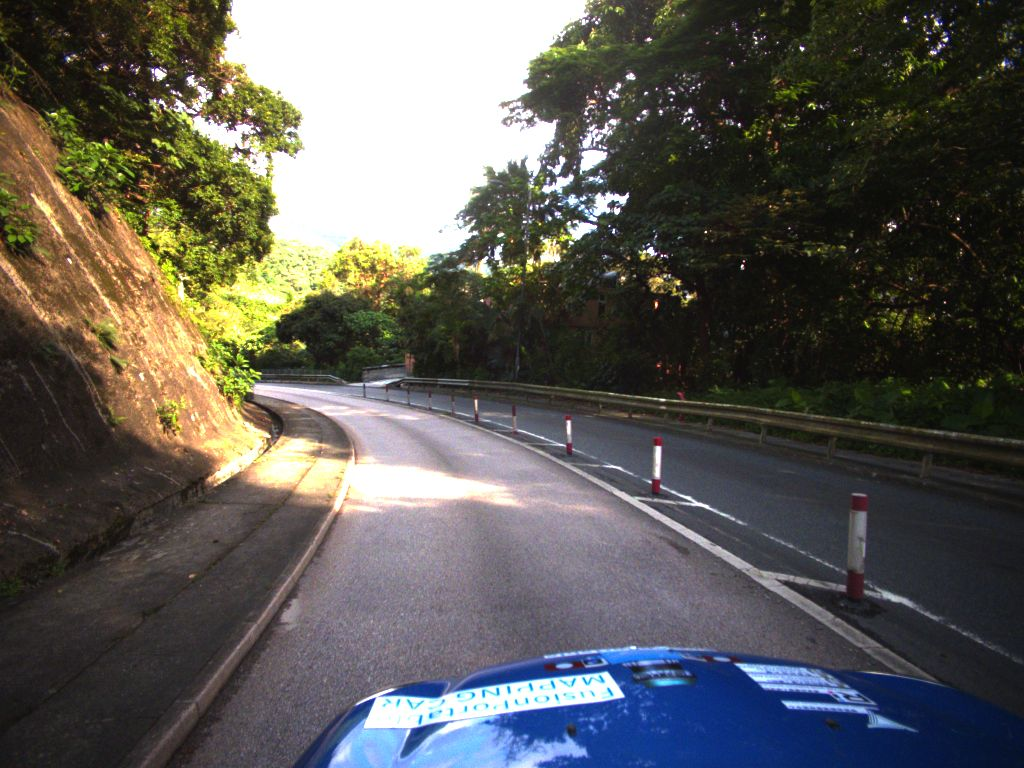}\hfill
  \includegraphics[width=0.18\linewidth]{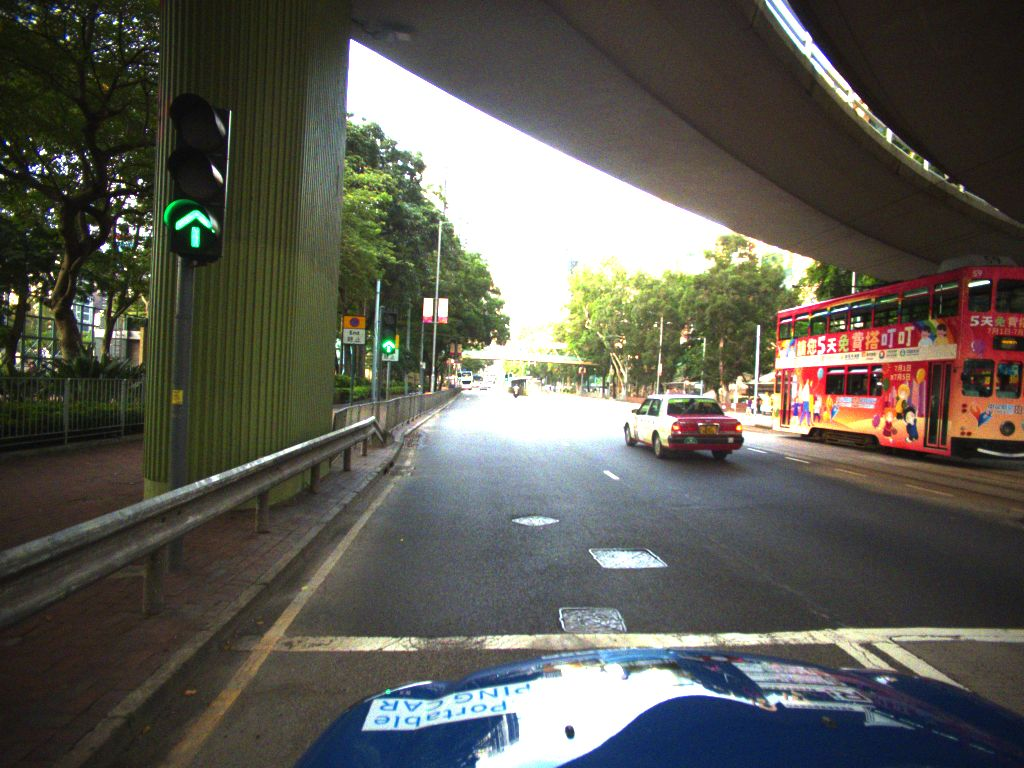}\hfill
  \includegraphics[width=0.18\linewidth]{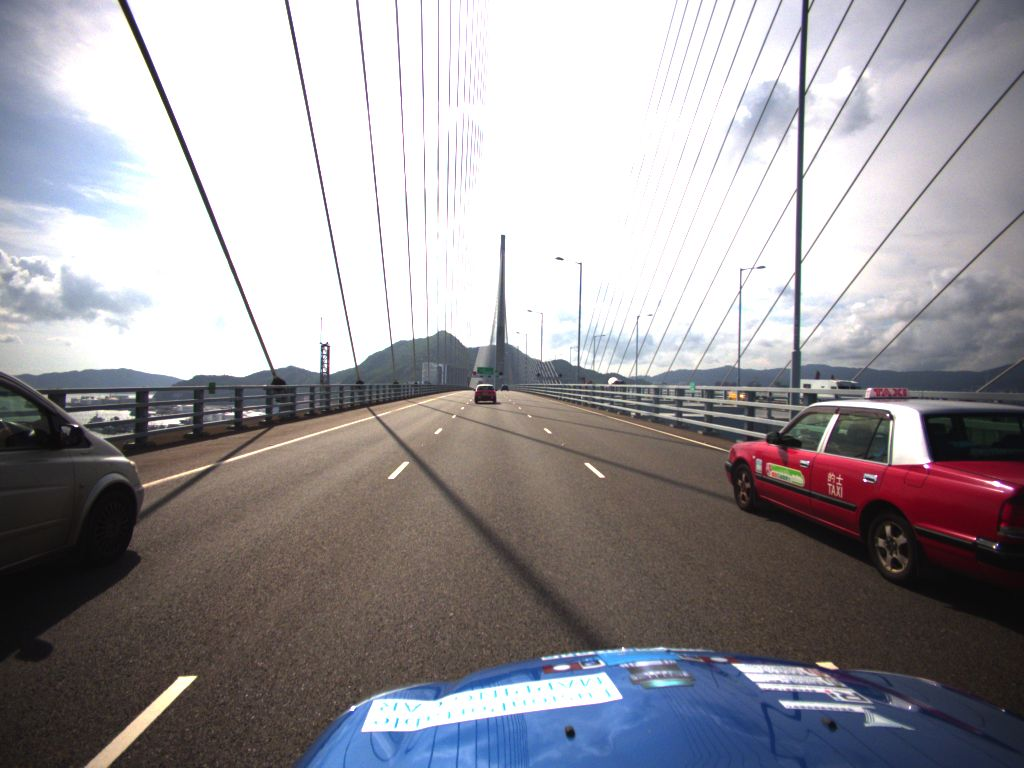}\hfill
  \includegraphics[width=0.18\linewidth]{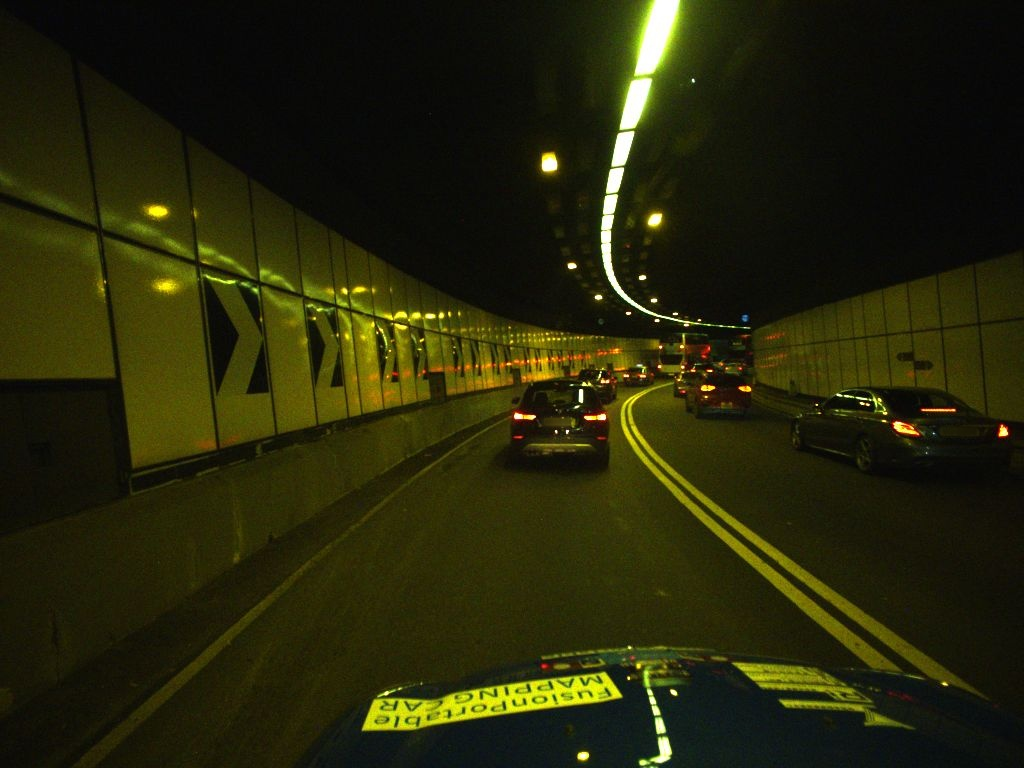}\\
  \centering{(d) The high-speed vehicle and scene images covering the mountain road, urban road, highway, and tunnel.}\vspace{0.1cm}\\
  \includegraphics[width=0.19\linewidth]{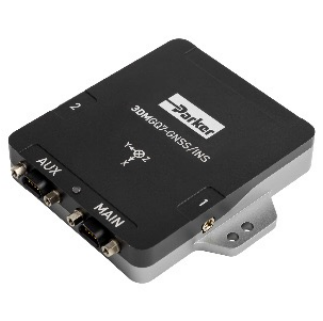}\hfill
  \includegraphics[width=0.19\linewidth]{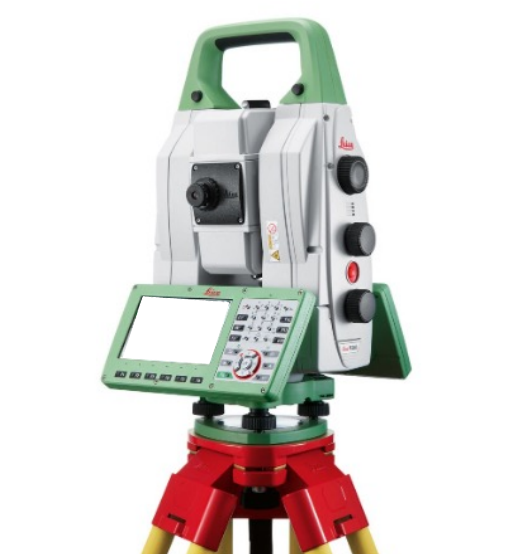}\hfill
  \includegraphics[width=0.19\linewidth]{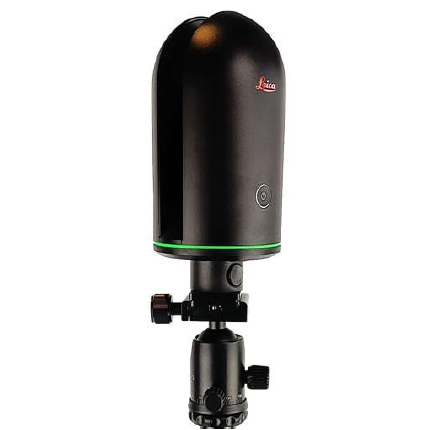}\hfill
  \includegraphics[width=0.19\linewidth]{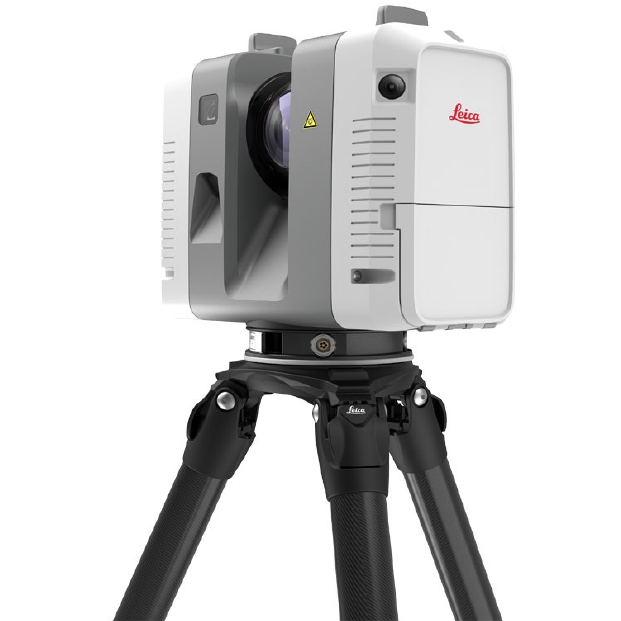} \\
  \centering{(e) Devices for generating GT trajectories and maps: $3$DM-GQ$7$, Leica MS$60$, Leica BLK$360$, and Leica RTC$360$.}
  \caption{Platform-Specific Data Samples:
    \textbf{(a)} The handheld multi-sensor rig across various environments,
    \textbf{(b)} the legged robot,
    \textbf{(c)} the low-speed UGV,
    \textbf{(d)} the high-speed vehicle, and
    \textbf{(e)} the GT generation device.
    The depicted scenes highlight the FusionPortableV2 dataset's comprehensive coverage across a spectrum of platforms and environmental conditions.}
  \label{fig:sample_data}
  \vspace{-0.2cm}
\end{figure*}

\subsubsection{\textbf{Inertial Measurement Unit:}}
The STIM$300$ IMU, a tactical-grade sensor\endnote{\url{https://www.sensonor.com/media/1132/ts1524r9-datasheet-stim300.pdf}}, serves as the primary inertial sensor of our system, mounted beneath the LiDAR.
It has a bias instability of $0.3^{\circ}/h$ for the gyroscope and $0.04mg$ for the accelerometer. The sensor outputs angular velocity and acceleration measurements at $200Hz$.
Other components, including the LiDAR, event cameras, and the $3$DM-GQ$7$ Inertial Navigation System (INS), are also integrated with IMUs. Further details are provided in subsequent sections.

\subsection{Platform-Specific Sensor Setup}
\label{sec:platform_sensor}
Our goal is to create a diverse dataset by capturing sequences with multiple mobile platforms, thereby increasing the dataset's complexity and challenge compared to those relying on a single platform.
Each platform is equipped with a handheld multi-sensor suite and platform-specific sensors, as shown in Fig. \ref{fig:layout}.
Fig. \ref{fig:sample_data} displays the platforms and exemplifies typical scenes from which data were gathered.
Platform-specific sensor settings are introduced in the subsequent sections, while the description of their motion and scenario patterns are presented in Section~\ref{sec:platform_pattern}.

\subsubsection{\textbf{Legged Robot:}}
We have selected the Unitree A1 quadruped robot as our legged platform, as shown in Fig. \ref{fig:legged_layout}.
This robot is equipped with $12$ joint motor encoders and $4$ contact sensors per leg,  located at the hip, thigh, calf, and foot.
These sensors provide kinematic measurements at a rate of $50Hz$.
The MSS is affixed to the robot's dorsal side and communicates with the kinematic sensors via Ethernet.
In addition to the raw sensor measurements, we record metadata for each motor, which includes torque, velocity, position, and temperature, along with kinematic-inertial odometry data.

\subsubsection{\textbf{Unmanned Ground Vehicle:}}
The MSS is integrated into a four-wheeled Ackerman UGV (see Fig. \ref{fig:ugv_layout1}, \ref{fig:ugv_layout2}), originally designed for logistics transportation \citep{liu2021role}.
To optimize signal reception, the dual GNSS antennas of the INS are positioned at UGV's rear side.
Kinematic data for the UGV is acquired through two incremental rotary encoders, strategically positioned at the center of the rear wheel.
These encoders, featuring $1000$ pulses per revolution, produce measurement data at a rate of approximately $100Hz$, which is then recorded.
\subsubsection{\textbf{Vehicle:}}
As depicted in Fig. \ref{fig:vehicle_layout}, we follow the KITTI setup \citep{geiger2013vision} by extending the baseline of both the stereo cameras and the dual antenna, with the stereo frame camera having a baseline of $83cm$ and the event camera having a baseline of $73cm$.
This extended baseline enhances the accuracy of depth estimation for distant objects, as compared with that in the UGV.
The MSS is mounted on the vehicle's luggage rack using a custom-designed aluminum frame.

\subsection{Ground Truth Provision Setup}
\label{sec:ground_truth_setup}
High-precision, dense RGB point cloud maps and GT trajectories are essential for evaluating SLAM and perception algorithms. This section describes three types of GT devices featured in our dataset, selected to meet the varied needs of the sequences. Through the integration of data from these GT devices, our dataset provides comprehensive support for algorithm benchmarking, not only in localization and mapping but also across diverse applications and requirements.

\subsubsection{\textbf{Dense RGB Point Cloud Map:}}
\label{sec:gt_map}
For creating dense point cloud maps of outdoor scenarios, the Leica RTC360 laser scanner was selected, because of its high scanning rate of up to $2$ million points per second and accuracy under $5.3mm$ within a $40m$ radius.
Some indoor areas were scanned with the Leica BLK$360$, which operates at a rate of $0.68$ million points per second and achieves an accuracy of $4mm$ within a $10m$ range.
All scans are registered and merged by the Leica Cyclone software\endnote{\url{https://leica-geosystems.com/products/laser-scanners/software/leica-cyclone/leica-cyclone-register-360}}, resulting in a dense and precise RGB point cloud map (Fig. \ref{fig:rgb_point_cloud_quality}).
This map with the resolution as $8cm$, covering all data collection areas, can be used to evaluate the mapping results of algorithms\endnote{\url{https://github.com/JokerJohn/Cloud_Map_Evaluation}} ranging from model-based \citep{lin2021r3live} and learning-based methods \citep{pan2024pin}.

\subsubsection{\textbf{3-DoF GT Trajectory:}}
\label{sec:gt_3d_traj}
For indoor and small-scale outdoor environments, the Leica MS$60$ total station\endnote{\url{https://leica-geosystems.com/en-us/products/total-stations/multistation/leica-nova-ms60}} was utilized to measure the GT trajectory of the robot at the $3$-DoF position.
As shown in Fig. \ref{fig:sample_data} (a), the tracking prism is placed atop the LiDAR.
The GT trajectory was captured at a frequency between $5$-$8Hz$, achieving an accuracy of $1mm$.
However, due to occasional instability in the measurement rate, the GT trajectory is resampled at $20Hz$ using cubic spline interpolation for a more consistent evaluation.
For further details on this process, please refer to Section \ref{sec:processing_ms60}.

\subsubsection{\textbf{6-DoF GT Trajectory:}}
\label{sec:gt_6d_traj}
While the stationary Leica MS$60$ provides accurate measurements, it cannot track the prism when it is occluded or outside the visible range.
Consequently, we employ the INS to capture $6$-DoF GT trajectories in large-scale and outdoor environments with available GNSS satellites.
This sensor integrates data from its internal dual-antenna RTK-GNSS, which provides raw data at a frequency of $2Hz$, and an IMU, to deliver estimated poses with an output rate of up to $30Hz$.
Before commencing data collection, we ensure that the GNSS has initialized with a sufficient satellite lock and the RTK is in a fixed status, typically achieving a positioning accuracy of up to $1.4cm$. This initialization process usually takes $1$-$3min$ in outdoor. To maintain the reliability of our ground truth data, we adhere to strict criteria. Firstly, the INS filter must be in a stable navigation status. Secondly, both dual-antenna GNSS must be in RTK-fixed status, receiving signals from at least $20$ satellites. Lastly, the positional error covariance must have converged to an optimal state.

\begin{table*}[ht]
    \centering
    \caption{Description of intrinsic and extrinsic parameter calibration.}
    \renewcommand\arraystretch{0.90}
    \renewcommand\tabcolsep{4pt}
    \footnotesize
    \begin{tabular}{lllll}
        \toprule
        Type                        & Sensor                     & Calibrated Parameter                                                                                                                    & Approach                                                                                                                                                             \\
        \midrule[0.03cm]

        \multirow{3}{*}{Intrinsics} &
        IMU                         & Noisy Density, Random Walk & Allen variance analysis \bt{Toolbox}\endnote{\url{https://github.com/ori-drs/allan_variance_ros}}                                                                                                                                                                                                              \\
                                    & Wheel Encoder              & Wheel Radius, Axle Track                                                                                                                & Minimize alignment error
        between $\mathcal{T}_{gt}$ and $\mathcal{T}_{est}$ \citep{jeong2019complex}$^{\#}$                                                                                                                                                                                                                                                                                        \\
                                    & Camera                     & Focal Length, Center Point, Distortion                                                                                                  & Minimize reprojection error \citep{zhang2000flexible}\endnote{\url{https://www.mathworks.com/help/vision/camera-calibration.html}}                                   \\
        \midrule[0.01cm]

        \multirow{6}{*}{Extrinsics} &
        IMU-IMU                     & Rotation, Translation      & Optimization \citep{rehder2016extending}\endnote{\url{https://github.com/ethz-asl/kalibr/wiki/Multi-IMU-and-IMU-intrinsic-calibration}}                                                                                                                                                                        \\
                                    & IMU-Camera                 & Rotation, Translation, Cons. Time Offset                                                                                                & Optimization                                              \citep{furgale2013unified}\endnote{\url{https://github.com/ethz-asl/kalibr/wiki/camera-imu-calibration}}   \\
                                    & IMU-Prism                  & Translation                                                                                                                             & Hand-eye calibration \citep{furrer2018evaluation}                                                                                                                    \\
                                    & IMU-Legged Sensors         & Rotation, Translation                                                                                                                   & Obtained from the CAD Model                            \endnote{\url{https://github.com/unitreerobotics/unitree_ros/blob/master/robots/a1_description/urdf/a1.urdf}} \\
                                    & Camera-Camera              & Rotation, Translation                                                                                                                   & Minimize reprojection errors \citep{zhang2000flexible}                                                                                                               \\
                                    & Camera-LiDAR               & Rotation, Translation                                                                                                                   & Minimize point-to-line and point-to-plane errors \citep{jiao2023lce}\endnote{\url{https://github.com/HKUSTGZ-IADC/LCECalib}}                                         \\

        \bottomrule[0.03cm]
        \multicolumn{3}{l}{
            $\#$ detailed in Section \ref{sec:ext_calibration} of the paper
        }
                                    &                            &                                                                                                                                                                                                                                                                                                                \\
    \end{tabular}
    \label{tab:calibration_description}
\end{table*}

\section{Sensor Calibration}
\label{sec:calibration}
\runninghead{Wei \textit{et~al.}}

\begin{figure}[]
    \centering
    \includegraphics[width=0.35\textwidth]{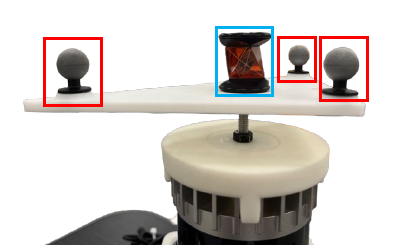}
    \caption{Sensor placement for the IMU-Prism calibration. Reflective balls for motion capture cameras (MCC) and the prism are marked in \textcolor{red}{red} and \textcolor{blue}{blue}, respectively. We use MCC's measurements to infer high-rate motion of the prism.}
    \label{fig:prism_calib}
\end{figure}

We meticulously calibrate the \textit{intrinsics} of each sensor, their \textit{extrinsics}, and the \textit{time offsets} between certain sensors beforehand.
The STIM$300$ IMU's coordinate system is designated as the \textit{body frame}, serving as the primary reference for most extrinsic calibrations.
For indirectly calibrated sensors, conversion is achieved through matrix multiplication: $\mathbf{T}^{A}_{C}=\mathbf{T}^{A}_{B}\mathbf{T}^{B}_{C}$. The positioning and orientation of sensors across devices and platforms are illustrated in Fig. \ref{fig:device_layout} and Fig. \ref{fig:layout}.
A summary of the calibration process is provided in Table \ref{tab:calibration_description}. We show detailed results and guidelines for good calibration on the dataset's website due to page constraints.

\subsection{Intrinsic Calibration}
\begin{figure}[]
    \centering
    \includegraphics[width=0.42\textwidth]{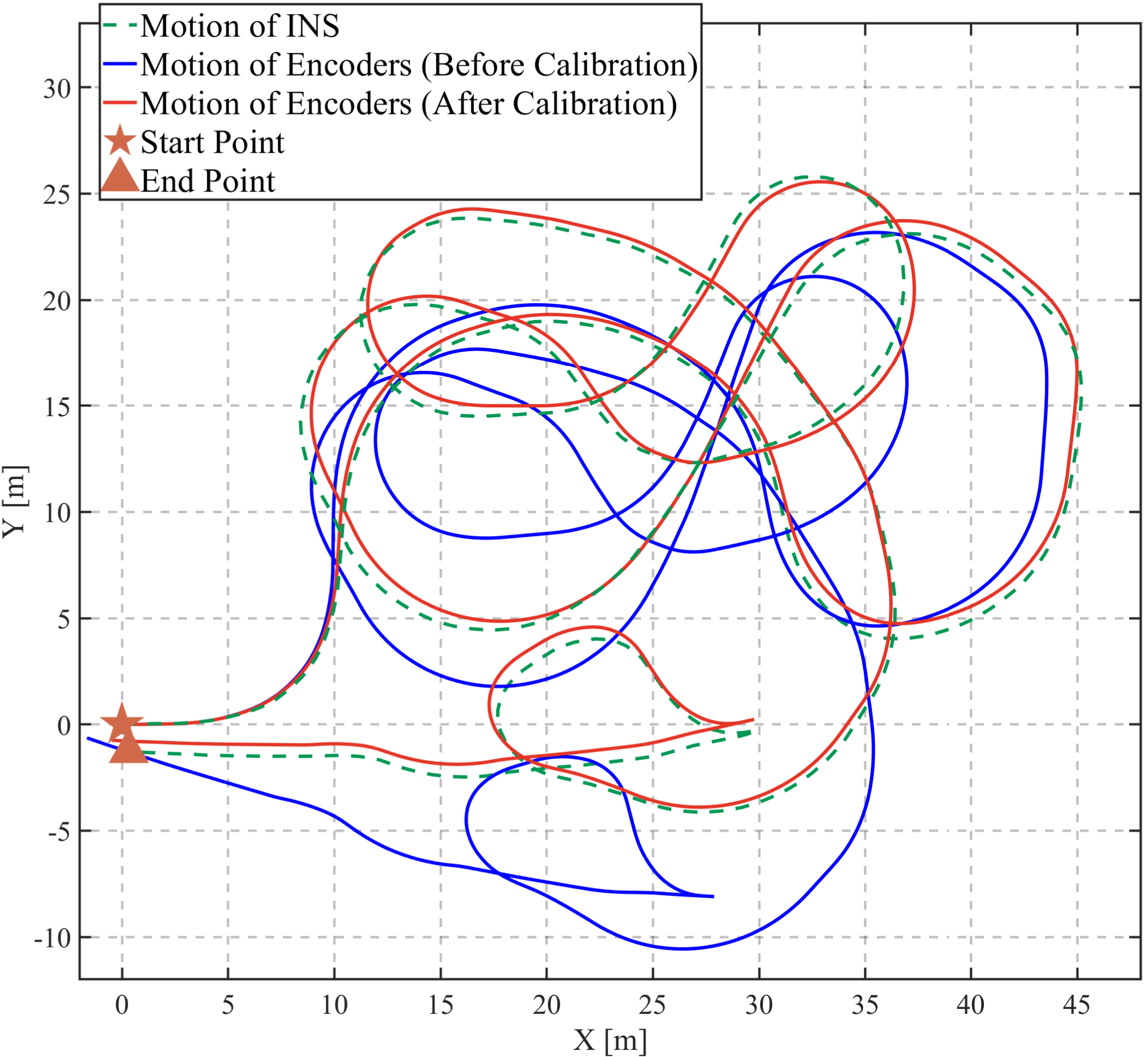}
    \caption{Comparison of trajectories: estimated motion by the INS ($3$DM-GQ$7$) (\textcolor{red}{red}), integration of encoders' measurements before calibration (\textcolor{green}{green}), and after calibration (\textcolor{blue}{blue}) using the sequence \texttt{Ugv\_parking00} for calibration. The ATE of the trajectory alignment is $0.98m$.}
    \label{fig:calibrate_odom}
\end{figure}

We calibrate IMUs and cameras using the off-the-shelf Kalibr toolbox \citep{furgale2013unified,rehder2016extending}.
For wheel encoder intrinsics, such as wheel radius and axle track, we implement the motion-based calibration algorithm outlined in \citep{jeong2019complex}.
This involves manually maneuver the UGV through significant transformations, as depicted in Fig. \ref{fig:calibrate_odom}. We calculate the UGV's planar motion for each interval $\tau \in[t_k,t_{k+1}]$ using encoder data to determine linear $v=(\omega_{l}r_{l}+\omega_{r}r_{r})/2$ and angular $\omega=(\omega_{l}r_{l}-\omega_{r}r_{r})/b$ velocities.
Concurrently, the INS captures more accurate motion estimates,
and intrinsics are then optimized by minimizing the trajectory alignment error between these two trajectories.

\subsection{Extrinsic Calibration}
\label{sec:ext_calibration}
Extrinsic calibrations, encompassing $6$-DoF transformations and time offsets for IMU-IMU, IMU-camera, IMU-prism, camera-camera, and camera-LiDAR pairs, are typically obtained with off-the-shelf toolboxes.
We specifically describe the calibration between the IMU and the prism that defines the reference frame of GT measurements relative to the total station (Leica MS$60$).
We design the indirect calibration method since the total station provides only $3$-DoF and low-rate trajectories.
We observed that the prism is visible to infrared cameras in the motion capture room.
We place and adjust three tracking markers around the prism to approximate its center, as shown in Fig. \ref{fig:prism_calib}.
We move the handheld device to perform the ``$8$''-shape trajectory.
Both the motion capture system and LiDAR-inertial odometry \citep{xu2021fast} can estimate trajectories of the prism and STIM$300$ IMU (\textit{body frame}), respectively.
With these trajectories, the IMU-Prism extrinsics are estimated using the hand-eye calibration algorithm \citep{furrer2018evaluation}.

\section{Dataset Description}
\label{sec:dataset_description}
\runninghead{Wei \textit{et~al.}}

This section begins by outlining the general motion patterns and potential applications for each platform. Following this, we consider the challenges posed by the dataset (as detailed in Section \ref{sec:dataset_challenging_factor}) and proceed to generate $\textbf{27}$ sequences designed for algorithm development and evaluation (refer to Section \ref{sec:dataset_seqence}).
A summary of the essential characteristics of these sequences is provided in Table~\ref{tab:dataset_summary}.
In our prior publication, FusionPortable \citep{jiao2022fusionportable}, we presented more handheld and legged robot sequences captured with a similar hardware configuration but without kinematic data. This dataset encompasses $\textbf{10}$ sequences featuring a variety of environments including garden, canteen, and escalator, captured via handheld devices, and $\textbf{6}$ sequences obtained from a legged robot within a motion capture room.

%%%%%%%%%%%%%%%%%%%%%%%%%%%%%%%%%%%%%%%%%%%%%%%%
\subsection{Analysis of Platforms Characteristics}
\label{sec:platform_pattern}
\begin{figure*}[t]
    \centering
    \subfigure[Motion Analysis of the Sequence: \texttt{Handheld\_room00}.]{
        \label{fig:handheld_motion}
        \centering
        \includegraphics[width=0.91\linewidth]{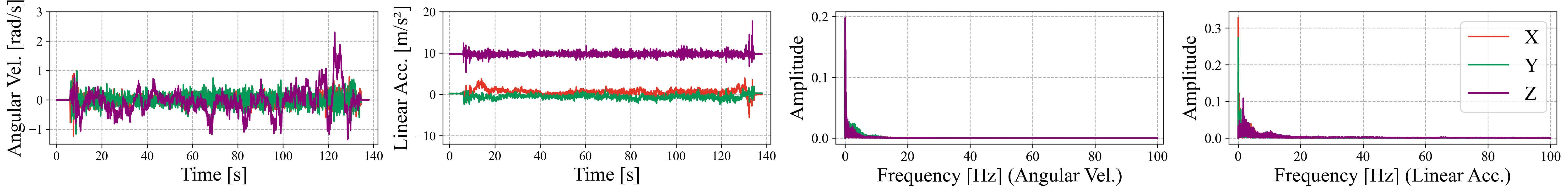}
    }
    \vspace{-0.3cm}
    \subfigure[Motion Analysis of the Sequence: \texttt{Legged\_grass00}.]{
        \label{fig:legged_motion}
        \centering
        \includegraphics[width=0.91\linewidth]{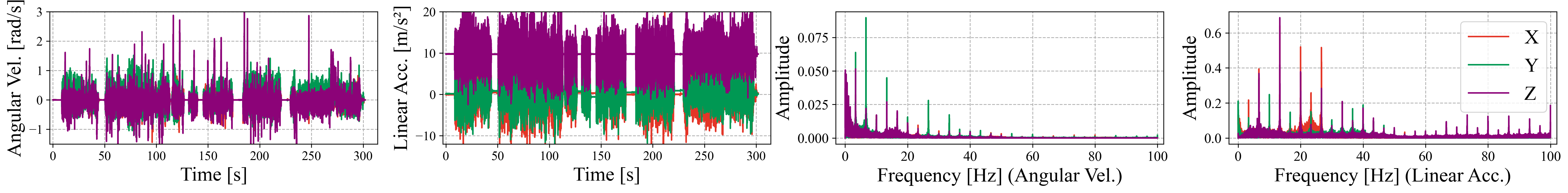}
    }
    \vspace{-0.3cm}
    \subfigure[Motion Analysis of the Sequence: \texttt{Ugv\_parking00}.]{
        \label{fig:ugv_motion}
        \centering
        \includegraphics[width=0.91\linewidth]{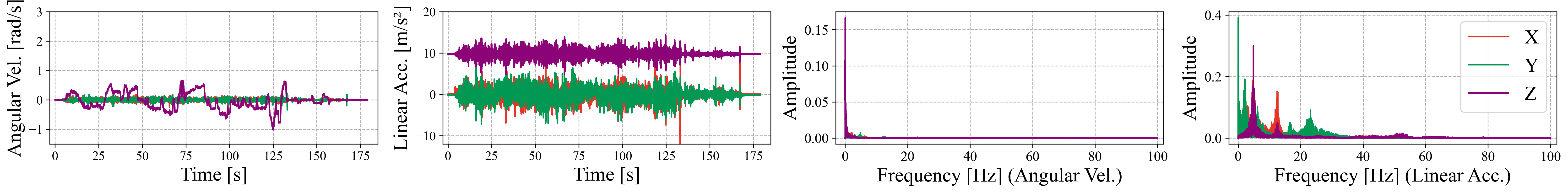}
    }
    \vspace{-0.3cm}
    \subfigure[Motion Analysis of the Sequence: \texttt{Vehicle\_highway00}.]{
        \label{fig:vehicle_motion}
        \centering
        \includegraphics[width=0.91\linewidth]{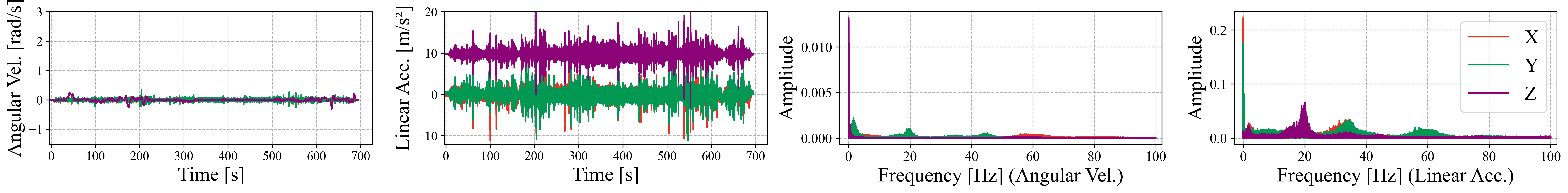}
    }
    \vspace{-0.3cm}
    \subfigure[Linear Velocity of Above Sequences: \texttt{Handheld\_room00}, \texttt{Legged\_grass00}, \texttt{Ugv\_parking00}, \texttt{Vehicle\_highway00}.]{
        \label{fig:velocity_motion}
        \centering
        \includegraphics[width=0.91\linewidth]{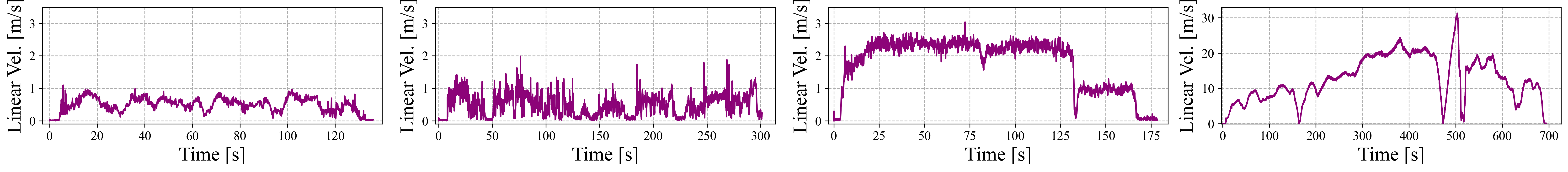}
    }
    \caption{Motion analysis with four mobile platforms in terms of linear acceleration $[m/s^{2}]$, angular velocity $[rad/s]$ and velocity $[m/s]$. We use measurements from STIM-$300$ to get linear acceleration and angular velocity and SLAM results to get rough velocity. (a)-(d): The left two columns of each row illustrate time-domain data, revealing immediate dynamic behaviors, while the right two columns display frequency-domain data, highlighting predominant motion features. Each line in the legend represents linear acceleration and angular velocity of each axis: $X$- (\textcolor{red}{Red}), $Y$- (\textcolor{green}{Green}), and $Z$- (\textcolor{purple}{Purple}). (e) The linear velocity of these sequences.}
    \label{fig:platforms_motion_pattern}
\end{figure*}

Each platform has its motion patterns (\textit{e.g.,} speed, angular velocity, dynamic frequency) and working ranges.
Fig. \ref{fig:platforms_motion_pattern} visualize typical motion patterns of different platforms on some example sequences.
Drawing from this observation, we meticulously design sequences to highlight the unique features of each platform.

%%%%%%%%%%%%%%%%%%%%%%%%%%%%%%%%%%%%%%%%%%%
\begin{table*}[]
  \centering
  \caption{Statistics and key challenges of each sequence are reported.
  Abbreviations: T: Total time. D: Total distance traveled. L: Large. M: Medium. S: Small.
  $||\overline{\bm{v}}||$: Mean linear velocity. $||\bm{v}||_{\text{max}}$: Max linear velocity ($3\sigma$). $3$-DoF (GNSS): Refer to Section \ref{sec:processing_3dmgq7}. 6-DoF (SLAM): Use FAST-LIO2 \citep{xu2021fast} to generate the reference trajectory.
  }
  \renewcommand\arraystretch{0.90}
  \renewcommand\tabcolsep{3pt}
  \footnotesize
  \begin{tabular}{llcccccccc}
    \toprule[0.03cm]
    Platform & Sequence                  & T$[s]$ & D$[m]$     &
    $||\overline{\bm{v}}||/||\bm{v}||_{\text{max}}[m/s]$
             & Scale                     & Motion & Challenges &
    GT Pose  & GT Map                                                                                                                            \\
    \midrule[0.03cm]

    \multirow{6}{*}{Handheld}
             & handheld\_grass$00$       & $140$  & $80$       & $0.55/1.76$   & S & 6-DoF Walk    & Textureless         & 3-DoF (Tracker) & Yes \\
             & handheld\_room$00$        & $140$  & $63$       & $0.41/1.40$   & S & 6-DoF Walk    & Dynamic             & 3-DoF (Tracker) & Yes \\
             & handheld\_room$01$        & $113$  & $46$       & $0.36/1.52$   & S & 6-DoF Walk    & Dynamic             & 3-DoF (Tracker) & Yes \\
             & handheld\_escalator$00$   & $247$  & $95$       & $0.45/1.45$   & S & 6-DoF Walk    & Non-inertial        & 3-DoF (Tracker) & Yes \\
             & handheld\_escalator$01$   & $254$  & $88$       & $0.47/1.54$   & S & 6-DoF Walk    & Non-inertial        & 3-DoF (Tracker) & Yes \\
             & handheld\_underground$00$ & $380$  & $403$      & $1.07/3.38$   & M & 6-DoF Walk    & Structureless       & 3-DoF (Tracker) & Yes \\

    \midrule[0.03cm]

    \multirow{6}{*}{
      \begin{tabular}[c]{@{}l@{}}
        Legged \\
        Robot
      \end{tabular}
    }
             & legged\_grass$00$         & $301$  & $112$      & $0.35/1.51$   & S & Jerky         & Deformable          & 3-DoF (Tracker) & Yes \\
             & legged\_grass$01$         & $355$  & $97$       & $0.32/1.58$   & S & Jerky         & Deformable          & 3-DoF (Tracker) & Yes \\
             & legged\_room$00$          & $173$  & $57$       & $0.28/1.30$   & S & Jerky         & Dynamic             & 3-DoF (Tracker) & Yes \\
             & legged\_transition$00$    & $233$  & $98$       & $0.41/1.60$   & S & Jerky         & Illumination        & 3-DoF (Tracker) & Yes \\
             & legged\_underground$00$   & $274$  & $167$      & $0.58/2.46$   & M & Jerky         & Structureless       & 3-DoF (Tracker) & Yes \\

    \midrule[0.03cm]

    \multirow{8}{*}{
      UGV
    }
             & ugv\_parking$00$          & $178$  & $319$      & $1.77/2.80$   & M & Smooth        & Structureless       & 6-DoF (INS)     & Yes \\
             & ugv\_parking$01$          & $292$  & $434$      & $1.48/4.07$   & M & Smooth        & Structureless       & 6-DoF (INS)     & Yes \\
             & ugv\_parking$02$          & $80$   & $242$      & $3.04/4.59$   & M & Jerky         & Structureless       & 6-DoF (INS)     & Yes \\
             & ugv\_parking$03$          & $79$   & $218$      & $2.75/4.92$   & M & Jerky         & Structureless       & 6-DoF (INS)     & Yes \\
             & ugv\_campus$00$           & $333$  & $898$      & $2.69/4.90$   & M & Smooth        & Scale               & 6-DoF (INS)     & Yes \\
             & ugv\_campus$01$           & $183$  & $343$      & $1.86/4.40$   & M & Jerky         & Fast Motion         & 6-DoF (INS)     & Yes \\
             & ugv\_transition$00$       & $491$  & $445$      & $1.04/3.25$   & S & Smooth        & GNSS-Denied         & 3-DoF (Tracker) & Yes \\
             & ugv\_transition$01$       & $375$  & $356$      & $0.92/3.51$   & S & Smooth        & GNSS-Denied         & 3-DoF (Tracker) & Yes \\
    \midrule[0.03cm]

    \multirow{8}{*}{Vehicle}
             & vehicle\_campus$00$       & $610$  & $2708$     & $4.43/9.31$   & M & Height-Change & Scale               & 6-DoF (INS)     & No  \\
             & vehicle\_campus$01$       & $420$  & $2086$     & $4.96/8.59$   & M & Height-Change & Scale               & 6-DoF (INS)     & No  \\
             & vehicle\_street$00$       & $578$  & $8042$     & $13.90/19.52$ & L & High-Speed    & Dynamic             & 3-DoF (GNSS)    & No  \\
             & vehicle\_tunnel$00$       & $668$  & $3500$     & $5.24/15.00$  & L & High-Speed    & LiDAR               & 3-DoF (GNSS)    & No  \\
             & vehicle\_downhill$00$     & $512$  & $3738$     & $7.29/15.59$  & L & Height-Change & Illumination        & 6-DoF (INS)     & No  \\
             & vehicle\_highway$00$      & $694$  & $9349$     & $13.46/30.87$ & L & High-Speed    & Structureless       & 6-DoF (INS)     & No  \\
             & vehicle\_highway$01$      & $377$  & $3641$     & $9.64/24.36$  & L & High-Speed    & Structureless       & 6-DoF (INS)     & No  \\
             & vehicle\_multilayer$00$   & $607$  & $1021$     & $1.68/4.53$   & M & Spiral        & Perceptual Aliasing & 6-DoF (SLAM)    & No  \\
    \bottomrule[0.03cm]
  \end{tabular}
  \label{tab:dataset_summary}
\end{table*}

%%%%%%%%%%%%%%%%%%%%%%%%%%%%%%%%%%%%%%%%%%%

\subsubsection{\textbf{Handheld:}}
Since the handheld MSS is commonly held by a user, it offers flexibility for data collection scenarios.
The handheld multi-sensor device provides adaptable data collection across diverse settings, akin to market counterparts like the Leica BLK$2$GO mobile scanning device, which excels in precision scanning and motion estimates.
Therefore, we collect data in scenarios including a \textit{laboratory} with furniture and dynamic elements, uneven \textit{grasslands}, an \textit{escalator} for vertical transitions, and an \textit{underground parking lot} resembling long tunnels.
The device performs motion influenced by the user's walking or running, sometimes leading to camera shake and rapid directional shifts.
Each sequence contains at least one loop.
The average movement speed is around $2m/s$.

\subsubsection{\textbf{Legged Robot:}}
The quadruped robot carries a sensor suite and commonly operates in \textit{indoors}, \textit{outdoors}, and \textit{underground} for missions such as rescue, inspection, and document transportation.
It exhibits complex motion patterns that involve a combination of walking, trotting, and running gaits.
Deformable and rugged terrain can also affect motion's stability.
Our experiments reveal that high-frequency jitters and sudden bumps are challenging to SOTA LiDAR-inertial odometry methods \citep{xu2021fast}.
Therefore, we believe that the integration sensor measurements from the joint motor and contact for a better motion estimation deserve further study and thus provide these data \citep{yang2023cerberus}.
The operational speed of the robot is approximately $1.5m/s$.

\subsubsection{\textbf{Unmanned Ground Vehicle:}}
The UGV is typically designed for last-mile delivery and navigates middle-scale areas like campuses and factories.
Constrained by Ackermann steering geometry, the UGV executes planar and smooth movements in response to the operator's inputs.
Data collection is conducted in various environments, including an \textit{outdoor parking lot} (open space), a \textit{campus}, and the challenging \textit{transition zones} between indoor and outdoor environments (where GNSS signals are unstable).
To mimic real-world complexities, commands for sudden stops and $45^{\circ}$ turns are occasionally issued.
The UGV can move at speeds of approximately $5m/s$.

\subsubsection{\textbf{Vehicle:}}
The vehicle collects data across diverse urban environments in Hong Kong, navigating through \textit{mountain roads} with elevation shifts, \textit{multi-story parking lots} with varying heights and orientations, dynamic \textit{downtown areas} with buildings, \textit{highways}, and GNSS-denied \textit{underground tunnels}, which are structureless and lack distinctive textures.
It operates at a range of speeds from $10km/h$ to $100km/h$, sometimes with abrupt speed and directional changes influenced by traffic and road conditions.

%%%%%%%%%%%%%%%%%%%%%%%%%%%%%%%%%%%%%%%%%%%%%%%
\subsection{Challenging Factors}
\label{sec:dataset_challenging_factor}
Prior to data collection, we acknowledge that practical factors contribute to sensor degradation and potential algorithmic failure.
Our data sequences, integrated with the platforms described, aim to comprehensively evaluate algorithm performance in terms of accuracy, efficiency, and robustness.
Additionally, we anticipate these sequences will draw the development of novel algorithms.

\subsubsection{\textbf{Illumination Conditions:}}
\label{sec:factor_illumination}
Different illumination conditions, such as bright sunlight, shadows, and low light, affect the quality of visual sensors and pose challenges for visual perception algorithms.
For example, in bright sunlight, cameras are sometimes overexposed, resulting in a loss of appearance information.
On the contrary, cameras are sometimes underexposed in low light conditions, leading to image noise and poor visibility.

\subsubsection{\textbf{Richness of Texture and Structure:}}
\label{sec:factor_texture_structure}
Structured environments (\textit{e.g.}, offices or buildings) can mainly be explained using geometric primitives, while semi-structured environments have both geometric and complex elements like trees and sundries. Scenarios like narrow corridors are structured but may challenge state estimators.
Additionally, texture-rich scenes facilitate visual algorithms to extract stable features (\textit{e.g.}, points and lines), while texture-less may negatively affect the performance.
Also, in texture-less environments, only a small amount of events is triggered.

\subsubsection{\textbf{Dynamic Objects:}}
\label{sec:factor_dynamic}
In dynamic environments, several elements (\textit{e.g.,} pedestrians or cars) are moving when the data are captured.
This is in contrast to static environments.
For instance, moving cars cause noisy reflections and occlusions to LiDAR data, while pedestrians cause motion blur to images.
Overall, dynamic objects induce negative effects from several aspects such as incorrect data association, occlusion, and ``ghost'' points remaining on the map.

\begin{figure}[t]
    \centering
    \includegraphics[width=0.45\textwidth]{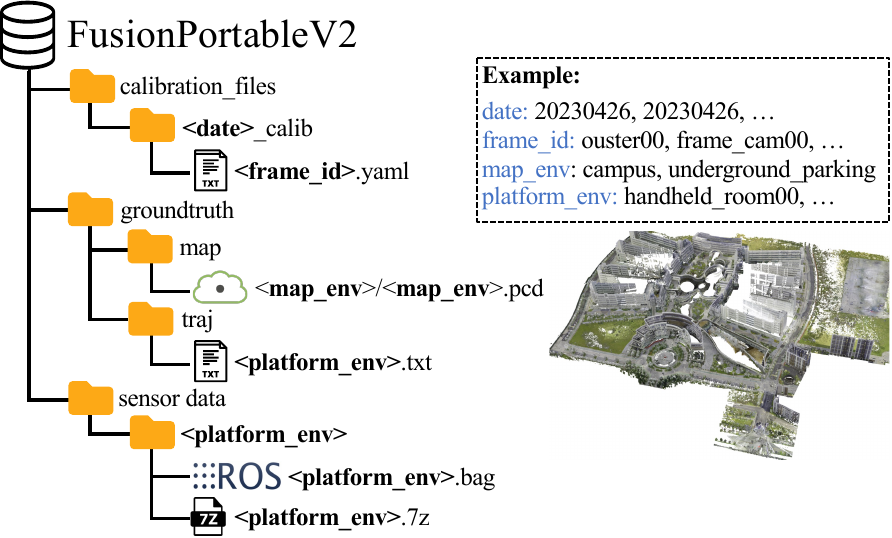}
    \caption{The dataset organization.}
    \label{fig:data_organization}
\end{figure}
\begin{figure}[t]
  \centering
  \subfigure[\texttt{Handheld\_escalator00}]
  {\label{fig:projection_escalator}\centering\includegraphics[width=.495\linewidth]{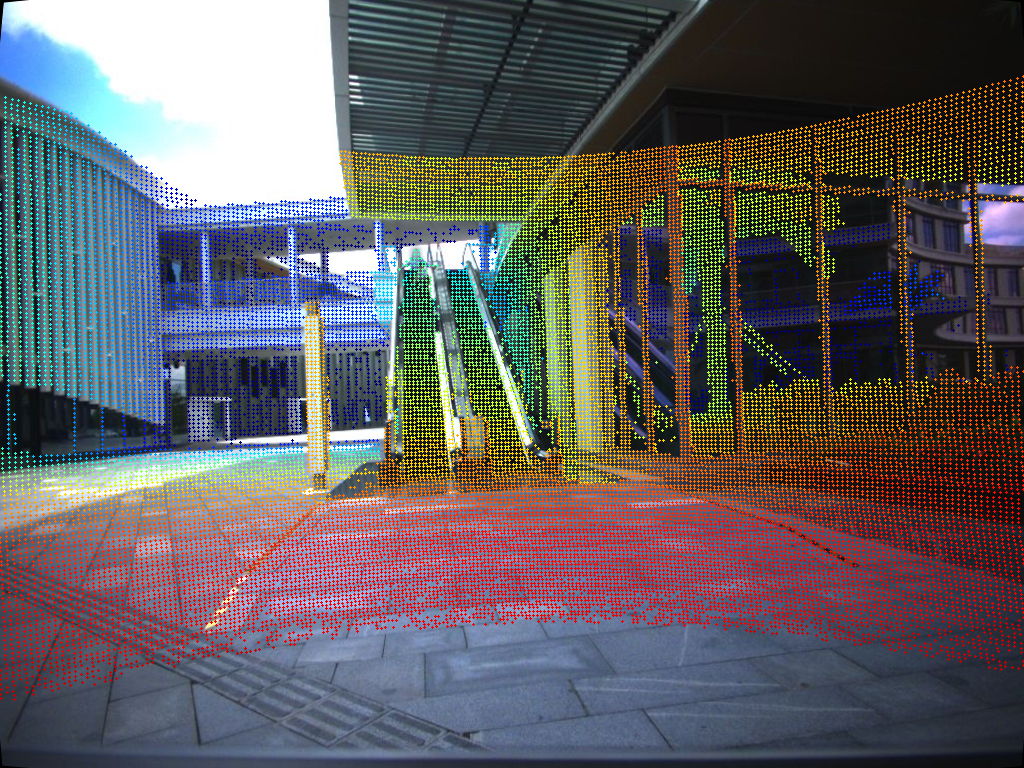}}
  \subfigure[\texttt{Vehicle\_campus00}]
  {\label{fig:projection_campus}\centering\includegraphics[width=.495\linewidth]{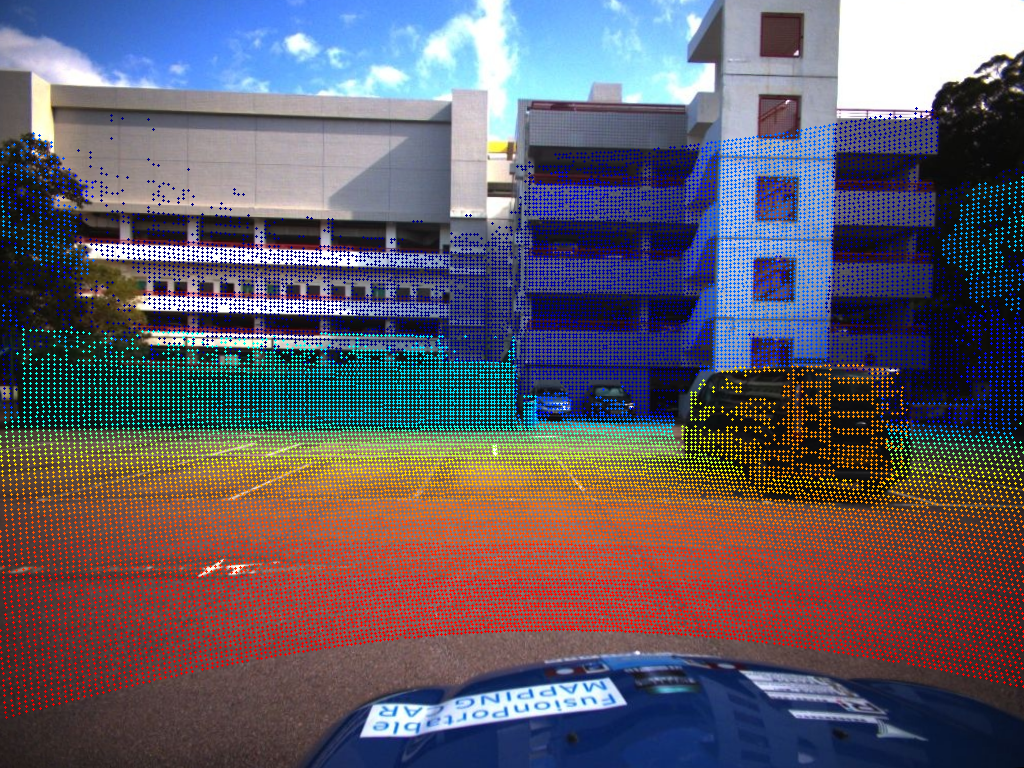}}
  \caption{The projected point cloud onto the left frame image using our SDK shows points' colors indicating relative distances. This involves a basic implementation, including a data loader, calibration loader, point cloud manipulation, and camera model.}
  \label{fig:development_tools}
\end{figure}

\subsubsection{\textbf{Intermittent GNSS:}}
\label{sec:factor_intermittent_gnss}
The intermittent GNSS signal issue typically arises in environments like places where dense and towering urban clusters are presented, overpasses, and indoor-outdoor transition areas.
A special example is the city center of Hong Kong.
In such scenarios, GNSS signals are often obstructed, leading to sporadic reception and significant uncertainty.

\subsubsection{\textbf{Scale Variability:}}
\label{sec:scale_variability}
Developing SLAM and perception algorithms for large-scale environments may encounter challenges such as an increased computational load and a heightened risk of perceptual aliasing.
The former necessitates stricter demands on algorithm latency and memory usage, whereas the latter requires more accurate long-term associations for place recognition \citep{yin2024general}, given the potential for environments to include geographically distant yet visually similar locations.

\subsubsection{\textbf{Viewpoint Change:}}
\label{sec:viewpoint_change}
We consider the viewpoint change from two perspectives: ``yaw-change'' and ``roll-pitch-change''.
The former often occurs in sequences with loops. Several sequences in our dataset feature at least one loop, posing challenges for place recognition and image matching algorithms.
For instance:
\texttt{ugv\_transition00} and \texttt{ugv\_transition01} contain loops in both indoor and outdoor environments; \texttt{vehicle\_multilayer00} captures multiple loops across different floors of a multi-story parking lot; and \texttt{vehicle\_campus00} and \texttt{vehicle\_campus01} collectively cover the entire HKUST campus, where loops are also present.
We also provide a similar sequence (\texttt{campus\_road\_day}) in FusionPortable \citep{jiao2022fusionportable}, collected $15$ months ago, during which several buildings and roads were updated.

The latter aspect is particularly relevant to the cross-view localization (CVL) problem \citep{shi2023boosting}, which involves viewpoint changes between down-facing satellite images and ground-level images. Open-source satellite images can be obtained from tools such as Google Earth.
Since several sequences include accurate GT geo-localization information and covers diverse scenarios (e.g., campus, urban roads, downhill mountain roads), it can serve as a challenging benchmark for CVL.

%%%%%%%%%%%%%%%%%%%%%%%%%%%%%%%%%%%%%%%%%%%%%%%
\subsection{Sequence Description}
\label{sec:dataset_seqence}
\begin{figure*}[]
    \centering
    \includegraphics[width=0.85\textwidth]{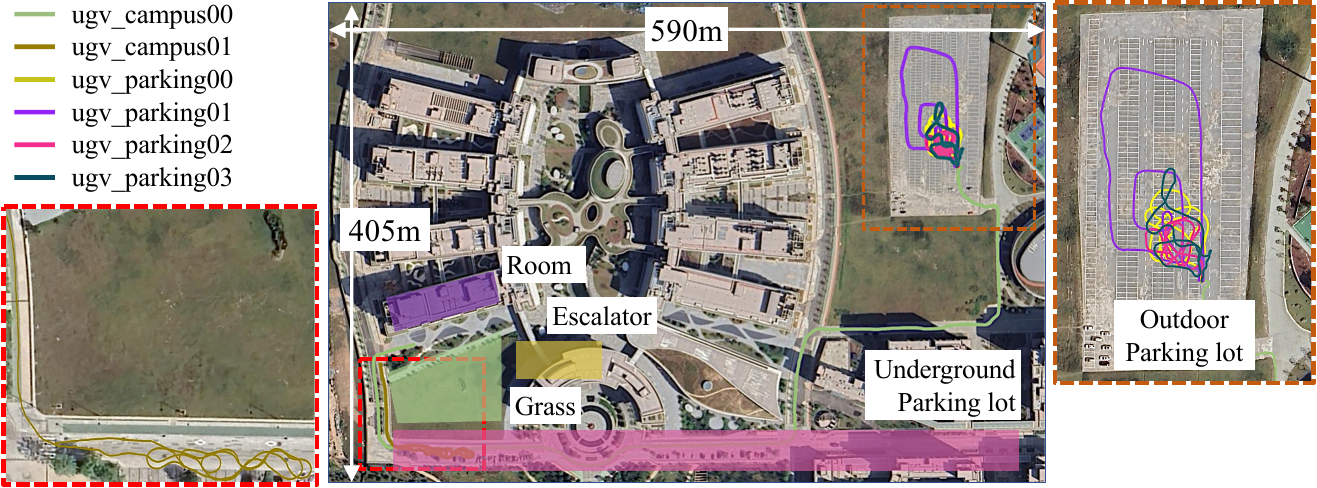}
    \caption{Trajectories of several sequences collected using the low-speed UGV in the campus, where environments with different structures and texture including room, escalator, grassland, parking lots are presented.}
    \label{fig:seq_traj_statellite_view_ugv}
    \vspace{-0.2cm}
\end{figure*}

\begin{figure*}[]
    \centering
    \includegraphics[width=0.85\textwidth]{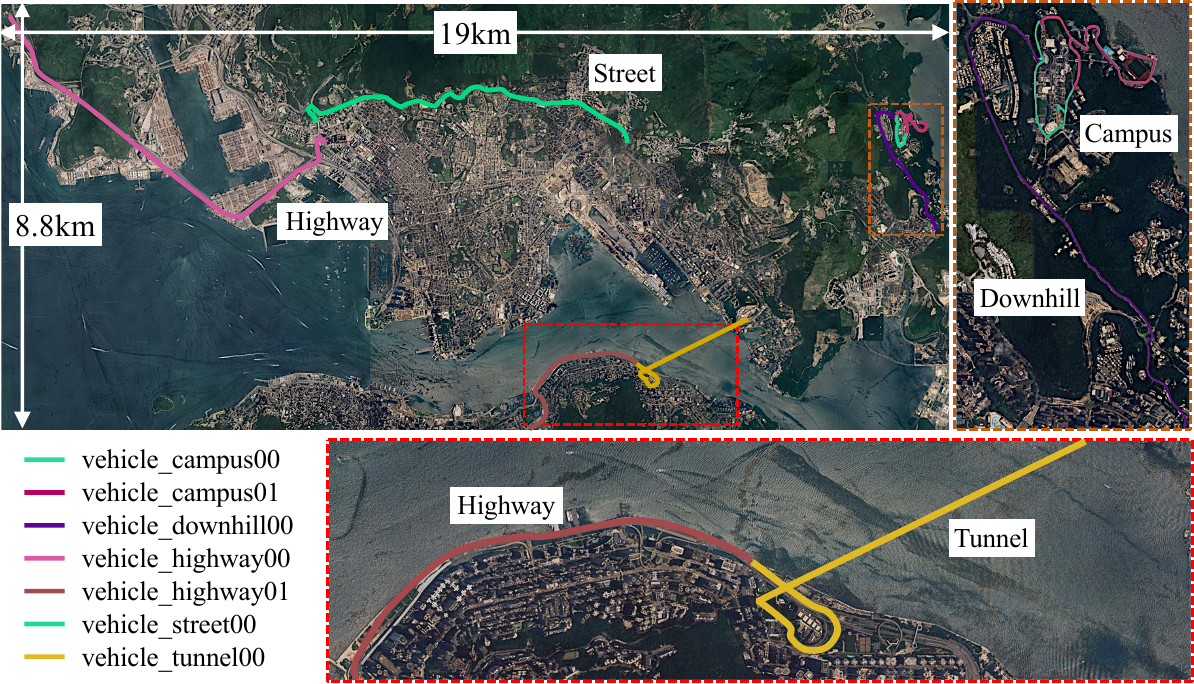}
    \caption{Trajectories of several sequences collected using the high-speed vehicle in Hong Kong.}
    \label{fig:seq_traj_statellite_view_vehicle}
    \vspace{-0.35cm}
\end{figure*}

Table \ref{tab:dataset_summary} summarizes the characteristics of our proposed sequences, detailing aspects such as temporal and spatial dimensions, motion patterns, locations, textural and structural richness, and whether GT poses and maps cover.
Fig. \ref{fig:seq_traj_statellite_view_ugv} and \ref{fig:seq_traj_statellite_view_vehicle} illustrate the coverage areas of the sequences from a satellite view perspective.

%%%%%%%%%%%%%%%%%%%%%%%%%%%%%%%%%%%%%%%%%%%%%%%%%%%%%%%%%%%%%%%%%%
\begin{figure*}[t]
    \centering
    \subfigure[Campus]{
        \label{fig:rgb_point_cloud_campus}
        \includegraphics[width=0.835\textwidth]{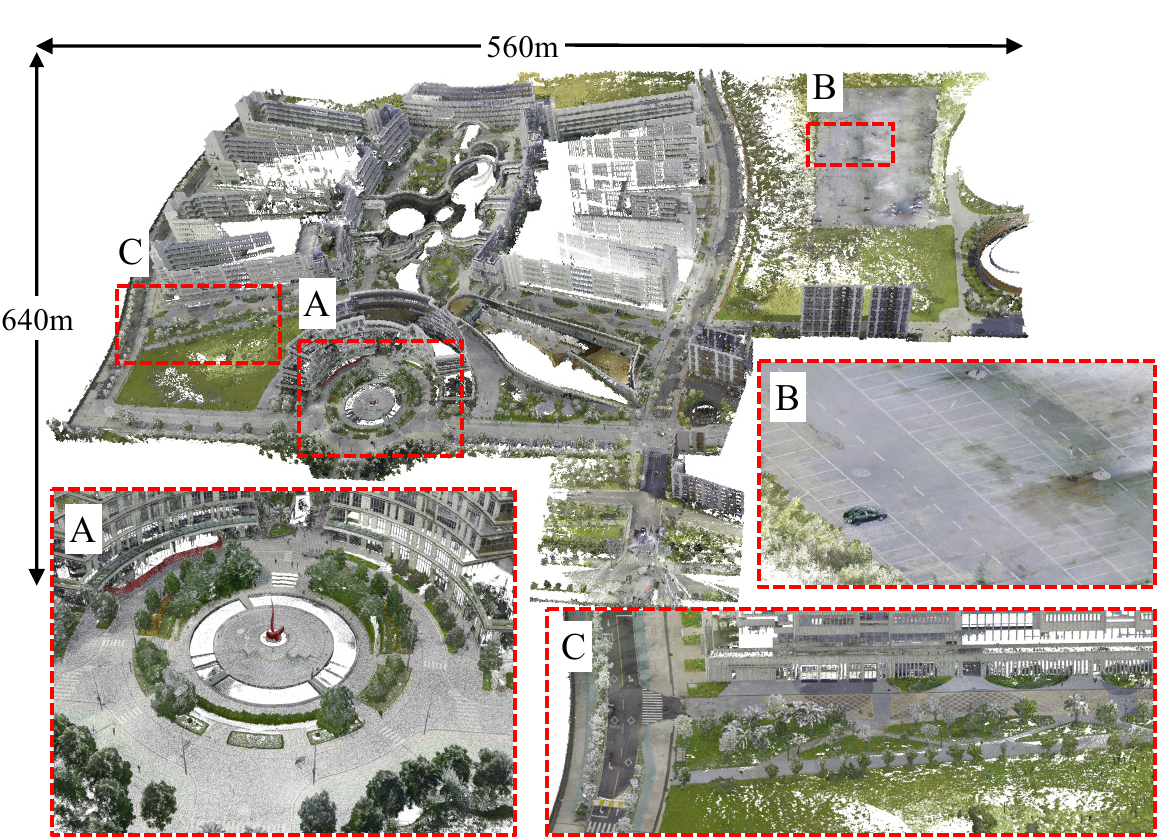}
    }
    \subfigure[Underground Parking Lot]{
        \label{fig:rgb_point_cloud_tunnel}
        \includegraphics[width=0.83\textwidth]{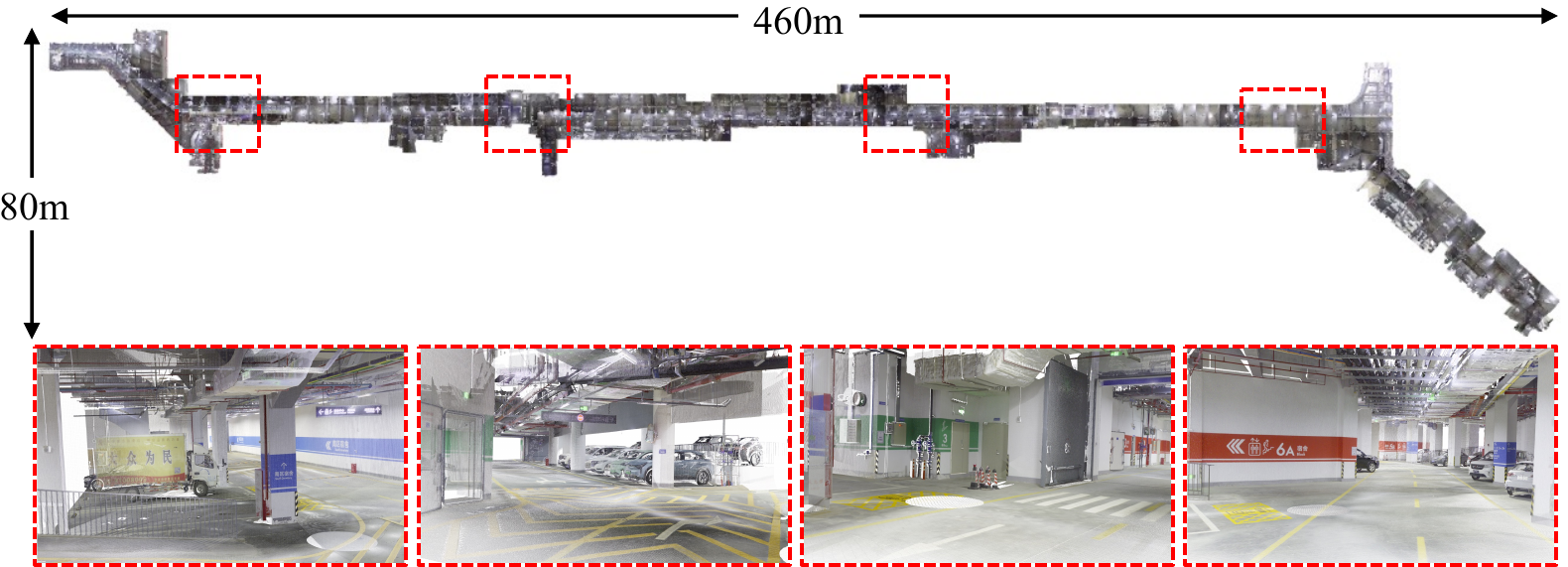}
    }
    \vspace{-0.3cm}
    \caption{GT RGB point cloud map of the (a) campus ($\approx 0.36km^2$) with $8cm$-resolution and (b) the underground parking lot ($\approx 0.037km^2$) with $4mm$-resolution. It almost encompasses the range of most sequences except for those related to vehicles. For detailed map information, please refer to our video presentation.}
    \label{fig:rgb_point_cloud_quality}
    \vspace{-0.3cm}
\end{figure*}
%%%%%%%%%%%%%%%%%%%%%%%%%%%%%%%%%%%%%%%%%%%%%%%%%%%%%%%%%%%%%%%%%%

%%%%%%%%%%%%%%%%%%%%%%%%%%%%%%%%%%
\subsection{Dataset Organization}
\label{sec:dataset_organization}

Fig. \ref{fig:data_organization} outlines our dataset's organization.
Sensor data were captured using the ROS bag tool\endnote{\url{https://wiki.ros.org/rosbag}}.
ROS bags are used due to their numerous advantages, such as mature tools for debugging, visualization of the tf tree, especially with multiple sensors present, a broadcasting mechanism for batching output messages of different types, and the ability to convert them into individual files.
To facilitate download, ROS bags were compressed with 7-Zip.
Each bag follows the naming convention $<$\textbf{platform}\_\textbf{environment}$>$. Sensor calibration parameters are saved in \textit{yaml} files naming as $<$\textbf{frame\_id}$>$ (\textit{e.g.,} \texttt{frame\_cam00.yaml} for the left RGB camera and \texttt{frame\_cam01.yaml} for the right.)
During the $6$-month dataset construction period, calibration was performed and documented multiple times, with parameters organized by calibration date (\textit{e.g.,} \texttt{20230426\_calib}).
Sequences must utilize the appropriate calibration files and such correspondences are provided in the development package.
GT poses at the TUM format are recorded in files matching the sequence names, detailing timestamp, orientation (as Hamilton quaternion), and translation vector per line.
GT map is provided as the \texttt{pcd} format, naming as $<$\textbf{location}\_\textbf{environment}$>$.
Please note that all the data provided have undergone additional post-processing steps, following the procedures detailed in Section \ref{sec:data_post_process}.

%%%%%%%%%%%%%%%%%%%%%%%%%%%%%%%%%%
\subsection{Development Tools}
\label{sec:tool}
We release a set of tools that enable users to tailor our dataset to their specific application needs.
Components are introduced as follows:

\subsubsection{\textbf{Software Development Kit (SDK):}}
\label{sec:development_package}
We present a Python-only SDK that is both extensible and user-friendly.
The kit includes foundational functions such as loading calibration parameters and visualizing them using a TF tree, parsing ROS messages into discrete files, data post-processing, and basic data manipulation.
Fig. \ref{fig:development_tools} shows the point cloud projection function provided by the package.

\subsubsection{\textbf{Evaluation:}}
We provide a set of scripts and tools for algorithm evaluation including localization and mapping.

\subsubsection{\textbf{Application:}}
We provide open-source repositories for users to try different applications with our dataset covering localization, mapping, monocular depth estimation, and anonymization of specific objects.
All can be found on the dataset website.

%%%%%%%%%%%%%%%%%%%%%%%%%%%%%%%%%%

\section{Data Post-Processing}
\label{sec:data_post_process}
\runninghead{Wei \textit{et~al.}}
The raw data captured by sensors and GT devices undergo post-processing before public release.
The specifics are outlined as follows.

\begin{figure}[]
    \centering
    \subfigure[Optimal Offset Enumeration]{
        \includegraphics[width=0.47\textwidth]{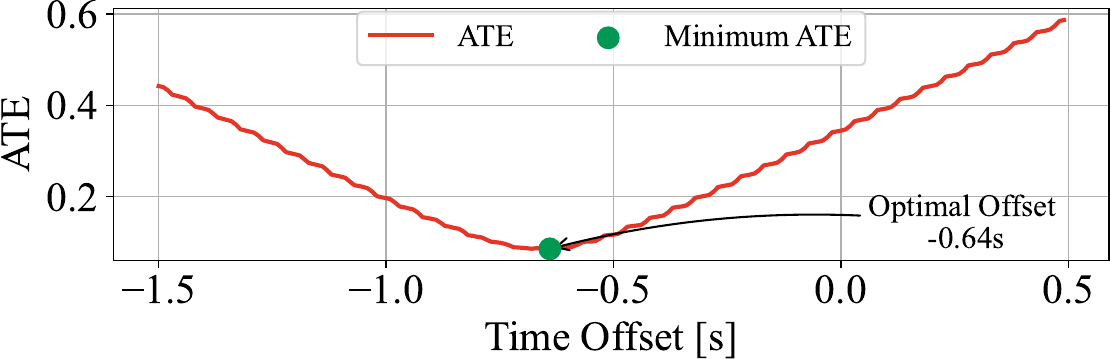}
    }
    \subfigure[Without Time Offset Correction]{
        \includegraphics[width=0.47\textwidth]{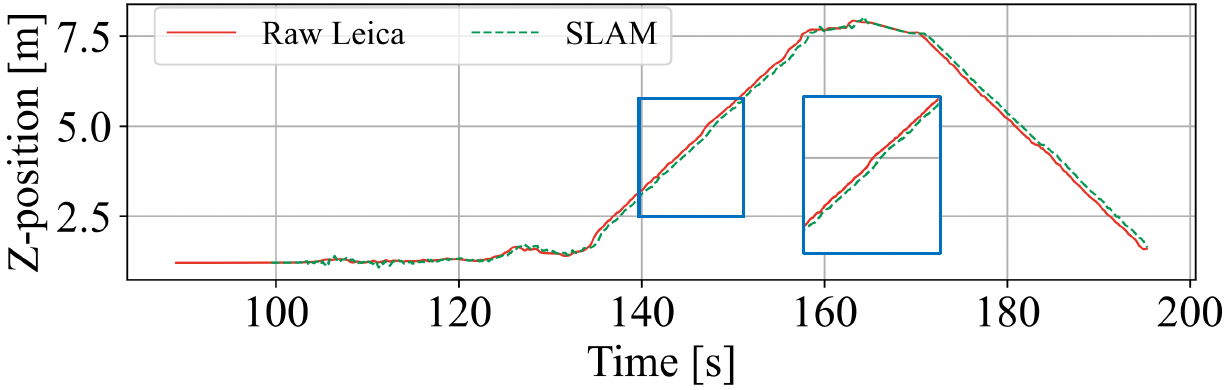}
    }
    \subfigure[With Time Offset Correction]{
        \includegraphics[width=0.47\textwidth]{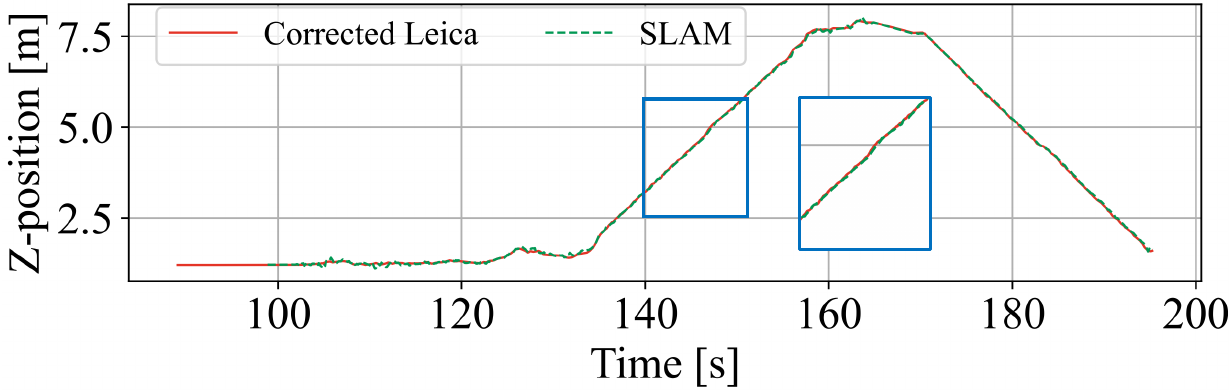}
    }
    \caption{Alignment process for \texttt{handheld\_escalator00} sequence. (a) deplicts the ATE versus time offset, pinpointing the optimal offset at $-0.64s$. (b) and (c) illustrate the $Z$-axis trajectories for the first $200$ seconds, before and after alignment, respectively, highlighting the applied time offset correction.}
    \label{fig:ms60_align}
\end{figure}
\vspace{-0.4cm}

\subsection{Privacy Management}
Data collection in public spaces such as the campus and urban roads was conducted with strict adherence to privacy regulations.
We employed the anonymization technique\endnote{\url{https://github.com/understand-ai/anonymizer}} introduced in \citep{burnett2023boreas} to obscure all human faces and license plates in images from our stereo frame cameras.
Building upon the original implementation, we enhanced the algorithm's efficiency using the ONNX Runtime Library\endnote{\url{https://onnxruntime.ai}}.
This upgraded version is now ROS-compatible, offering a valuable resource to the community, and is included in our development tools.

\subsection{GT Data Processing}
Due to diverse sources of GT trajectories and maps, it is necessary to standardize the GT data through processing and conversion and then verify them.
These steps should be executed sequentially.

\subsubsection{\textbf{3-DoF GT Poses of Total Station:}}
\label{sec:processing_ms60}
The preprocessing initiates with temporal alignment, crucial for synchronizing Leica MS$60$ total station measurements with sensor data, following the approach proposed in \citep{nguyen2022ntu}.
This synchronization finds the optimal time offset that minimizes the Absolute Trajectory Error (ATE) between the MS$60$'s recorded poses $\mathcal{T}_{ms60}$ and the SLAM-generated poses $\mathcal{T}_{alg}$.
This is done by enumerating offsets at intervals of $0.01s$ to generate various versions of time-shifted poses from the total station poses, as shown in Fig. \ref{fig:ms60_align}.
Upon adjusting their timestamps with the optimal $\delta$, all MS$60$ poses (recorded at $5$-$8Hz$) are resampled to a denser $20Hz$ sequence via cubic spline interpolation.
This method not only yields a smooth, continuous trajectory but also temporally synchronizes the data with SLAM algorithm outputs.
To maintain data accuracy, intervals longer than $1s$ indicating obstructions are omitted from the sequence.

\subsubsection{\textbf{6-DoF GT Poses of INS:}}
\label{sec:processing_3dmgq7}
Each $6$-DoF pose provided by the INS is accompanied by a variance value, indicating the measurement's uncertainty. This uncertainty increases when the GNSS signal is obstructed.
For a fair comparison, we manually removed data points with excessive uncertainty to serve as ground truth. We also provide the original data along with relevant tools for users to process the data according to their needs.
For the sequences \texttt{vehicle\_street00} and \texttt{vehicle\_tunnel00}, where intermittent GNSS signals disrupt the INS odometry filter convergence, we directly used the original GNSS data sampled at $2Hz$, providing 3-DoF GT poses instead.
For the sequence \texttt{vehicle\_multilayer00}, collected in a multi-layer parking garage, most of the trajectory is indoors without access to GNSS signals. Therefore, we used the results from the FAST-LIO2 algorithm as the baseline, given the rich structured information available. The start and end points of the sequence are located on the rooftop, where GNSS data can be used as loop closure references.

\subsubsection{\textbf{Accuacy of GT Maps:}}
In the construction of our ground truth (GT) maps, we employed a high-precision Leica scanner to capture detailed environments both indoors and outdoors.
Fig. \ref{fig:rgb_point_cloud_quality} displays the complete RGB point cloud map of the entire campus and a section of the underground parking garage. The fine details observable in the depicted areas highlight the high quality of both the point cloud and its color fidelity. According to the Leica Cyclone software report, the GT map exhibits an average error of less than $15mm$ across all pairwise scans, the average error is $3.4mm$ and with $90\%$ of these scans maintaining an error margin of $10mm$ or less.
The precision of our GT map exceeds that of point cloud maps constructed via current LiDAR SLAM technologies by nearly two orders of magnitude, making it suitable for algorithm evaluation.

\section{Experiment}
\label{sec:experiment}
\runninghead{Wei \textit{et~al.}}

We select $8$ representative sequences ($2$ sequences from each platform) from the dataset to conduct a series algorithm evaluation and verification.
Experiments include localization, mapping, and monocular depth estimation.

%%%%%%%%%%%%%%%%%%%%%%%%%%%%%%%%%%%%%%%%%%%%%%
%%%%%%%%%%%%%%%%%%%%%%%%%%%%%%%%%%%%%%%%%%%%%%
%%%%%%%%%%%%%%%%%%%%%%%%%%%%%%%%%%%%%%%%%%%%%%

\subsection{Evaluation of Localization}
\label{sec:experiment_localization}

\subsubsection{\textbf{Experiment Setting:}}
As one of the main applications, this dataset can be used to benchmark SOTA SLAM algorithms.
Here in, for evaluation of localization systems with different input modalities, we select four SOTA SLAM algorithms (including a learning-based method):
DROID-SLAM (left frame camera) \citep{teed2021droid},
VINS-Fusion (LC) (IMU + stereo frame cameras, with loop closure enabled) \citep{qin2018vins},
FAST-LIO2 (IMU+LiDAR) \citep{xu2021fast}, and
R3LIVE (IMU+LiDAR+left frame camera) \citep{lin2021r3live}.
The customized data loaders of each method are publicly released to foster research.

%%%%%%%%%%%%%%%%%%%%%%%%%%%%%%%%%%%%%%%%%%%%%%%%%%%%%%%%%%%%%%%%%%
\begin{figure*}[t]
  \centering
  \subfigure[\texttt{handheld\_room00}]{
    \includegraphics[width=0.16\textwidth]{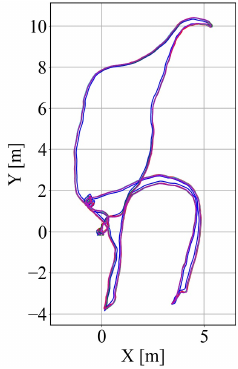}
  }
  \hspace{-0.25cm}
  \subfigure[\texttt{legged\_grass00}]{
    \includegraphics[width=0.18\textwidth]{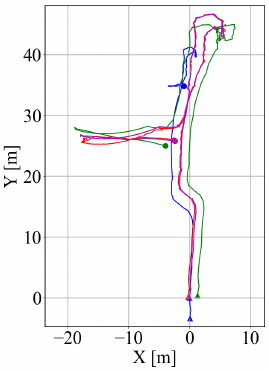}
  }
  \hspace{-0.3cm}
  \subfigure[\texttt{ugv\_parking00}]{
    \includegraphics[width=0.27\textwidth]{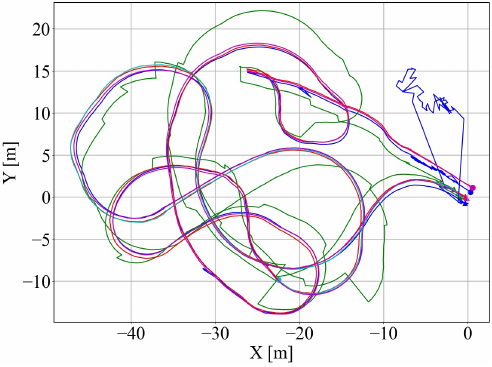}
  }
  \hspace{-0.35cm}
  \subfigure[\texttt{vehicle\_campus00}]{
    \includegraphics[width=0.37\textwidth]{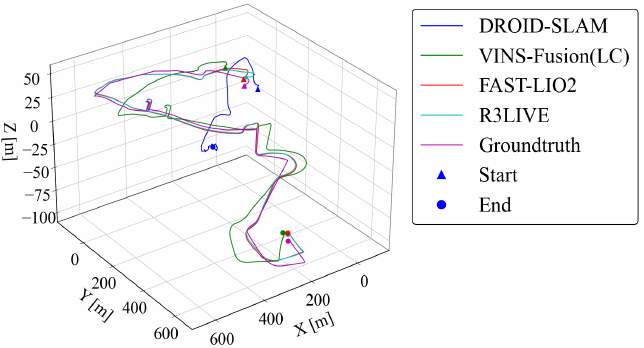}
  }
  \caption{This figure illustrates the comparative performance of leading SLAM algorithms operating across four distinct platforms and environmental contexts, from indoor spaces to a university campus. It is evident that LiDAR-based methods such as FAST-LIO2 and R3LIVE consistently outperform their vision-based counterparts across all scenarios, maintaining a higher trajectory accuracy. On the other hand, the performance of vision-based algorithms, particularly DROID-SLAM, deteriorates as the environment scale increases, with significant scale recovery issues observed in the expansive \texttt{vehicle\_campus00} sequence. This trend underscores the superior robustness of LiDAR-based SLAM in varied and large-scale environments.}
  \label{fig:trajectory_evaluation}
\end{figure*}
%%%%%%%%%%%%%%%%%%%%%%%%%%%%%%%%%%%%%%%%%%%%%%%%%%%%%%%%%%%%%%%%%%

Marching from traditional to deep learning-based SLAM methods, we evaluate DROID-SLAM, an end-to-end deep visual SLAM algorithm.
We employ DROID-SLAM on the monocular image stream with a pre-trained model\endnote{\url{https://github.com/princeton-vl/DROID-SLAM}} without fine-tuning to present a fair comparison to model-based methods. Meanwhile, testing the limits of its generalization ability is crucial for SLAM.
The pre-trained model is trained by supervision from optical flow and poses on the synthetic dataset TartanAir \citep{wang2020tartanair}, covering various conditions (\textit{e.g.}, appearance and viewpoint changes) and environments (\textit{e.g.}, from small-scale indoor to large-scale suburban).
All the experiments are conducted on NVIDIA GPU GeForce RTX $3090$ with a downsampled image resolution of $320\times 240$. The average runtime is $16$ FPS with a global bundle adjustment (BA) layer.
The average GPU memory consumption is below $11$ GB.

\subsubsection{\textbf{Evaluation:}}
We choose the typical evaluation metric: mean ATE to evaluate the accuracy of estimated trajectories against the GT using the EVO package.\endnote{\url{https://github.com/MichaelGrupp/evo}}
Table \ref{tab:exp_localization_accuracy} reports the quantitative localization results.

Our evaluation of SOTA SLAM systems, as summarized in Table \ref{tab:exp_localization_accuracy}, demonstrates that each system's performance varies across different environments, depending on its sensor configuration and algorithmic approach.
Due to the precise geometric information inherent in LiDAR raw data, methods incorporating LiDAR generally exhibit higher accuracy. However, as scene scale increases and becomes more complex (like the highway), segments lacking visual texture or structural features become challenging.
FAST-LIO2, which utilizes IMU and LiDAR data, showcased robust performance across a diverse array of environments. This highlights the inherent strength of LiDAR-based systems in tackling various and complex scenarios.
In contrast, R3LIVE, which integrates IMU, LiDAR, and visual data, consistently demonstrated superior accuracy in different settings, particularly outperforming FAST-LIO2 in scenarios where LiDAR degradation and jerky motion pattern are present (\textit{e.g.,} \texttt{ugv\_campus00}, \texttt{legged\_grass00}). However, in environments featuring intricate visual features such as water surfaces or reflective glass on the \texttt{ugv\_parking00}, the presence of visual mechanisms in R3LIVE may lead to a performance decrease.

For vision-based methods, VINS-Fusion outperforms DROID-SLAM on average, demonstrating robustness and generalization ability over learning-based methods. However, it is important to note that DROID-SLAM, using only monocular input, surpasses VINS-Fusion in three specific sequences: \texttt{legged\_room00}, and \texttt{ugv\_parking00}. These sequences are characterized by smaller-scale environments with constrained boundaries, such as closed rooms or parking areas.
This result highlights the promising potential of employing deep learning techniques in SLAM algorithms, particularly in scenarios where the environment is more confined and structured. The superior performance of DROID-SLAM in these specific cases suggests that learning-based methods can excel in certain conditions, despite their overall lower average performance compared to traditional approaches like VINS-Fusion.

%%%%%%%%%%%%%%%%%%%%%%%%%%%%%%%%%%%%%%%%%%%%%%
%%%%%%%%%%%%%%%%%%%%%%%%%%%%%%%%%%%%%%%%%%%%%%
%%%%%%%%%%%%%%%%%%%%%%%%%%%%%%%%%%%%%%%%%%%%%%
\begin{table}[t]
  \centering
  \caption{Localization accuracy: we calculate translation ATE $[m]$ for each sequence. The best result among all methods is shown in \textbf{bold}, and the best result among vision-based methods (VINS-Fusion and DROID-SLAM) is \underline{underlined}.}
  \renewcommand\arraystretch{1.10}
  \renewcommand\tabcolsep{4pt}
  \footnotesize
  \begin{tabular}{lcccc}
    \toprule[0.03cm]
    Sequence                  &
    \begin{tabular}[c]{@{}c@{}}R3LIVE\end{tabular} &
    \begin{tabular}[c]{@{}c@{}}FAST- \\
      LIO2\end{tabular} &
    \begin{tabular}[c]{@{}c@{}}VINS- \\
      Fusion (LC)\end{tabular} &
    \begin{tabular}[c]{@{}c@{}}DROID-\\SLAM\end{tabular}                                                                            \\
    \midrule[0.03cm]
    handheld\_room$00$        & $\bm{0.057}$ & $0.058$      & $\underline{0.063}$  & $0.118$             \\
    handheld\_escalator$00$   & $0.093$      & $\bm{0.085}$ & $\underline{0.258}$  & $4.427$             \\
    \midrule[0.03cm]

    legged\_grass$00$         & $\bm{0.069}$ & $0.327$      & $\underline{1.801}$  & $7.011$             \\
    legged\_room$00$          & $\bm{0.068}$ & $0.093$      & $0.149$              & $\underline{0.135}$ \\
    \midrule[0.03cm]

    ugv\_campus$00$           & $\bm{1.486}$ & $1.617$      & $\underline{1.866}$  & $43.869$            \\
    ugv\_parking$00$          & $0.424$      & $\bm{0.271}$ & $2.400$              & $\underline{2.019}$ \\
    \midrule[0.03cm]

    vehicle\_campus$00$       & $10.070$     & $\bm{8.584}$ & $\underline{66.428}$ & $\times$            \\
    vehicle\_highway$00$      & $\times$     & $686.940$    & $\times$             & $\times$            \\
    \bottomrule[0.03cm]
  \end{tabular}
  \label{tab:exp_localization_accuracy}
\end{table}

\subsection{Evaluation of Mapping}
\label{sec:experiment_mapping}

Localization and mapping represent the foundational tasks for robotic navigation, and evaluating trajectory accuracy alone does not suffice to encapsulate the efficacy of such processes comprehensively. Within the framework of SLAM algorithms predicated on Gaussian models, the map serves as a crucial output, and its accuracy assessment indirectly mirrors the precision of localization. For the broader spectrum of mapping tasks, whether conducted online or offline, sparse or dense, direct evaluation of map accuracy remains crucial. Hence, a module dedicated to assessing map accuracy has been developed to address this need, ensuring a holistic appraisal of navigational competencies.

\subsubsection{\textbf{Experiment Setting:}}
For evaluating point cloud maps estimated by SOTA SLAM algorithms, we first downsampled the estimated maps using a 0.1m grid. Initial alignment with the GT map was performed using CloudCompare software. We set the maximum threshold distance for corresponding points at $0.2m$. Thus, after evaluation, point pairs with a distance less than $0.2m$ were considered as the same point for distance calculation.

\subsubsection{\textbf{Evaluation:}}

\begin{figure*}[t]
  \centering
  \includegraphics[width=0.90\textwidth]{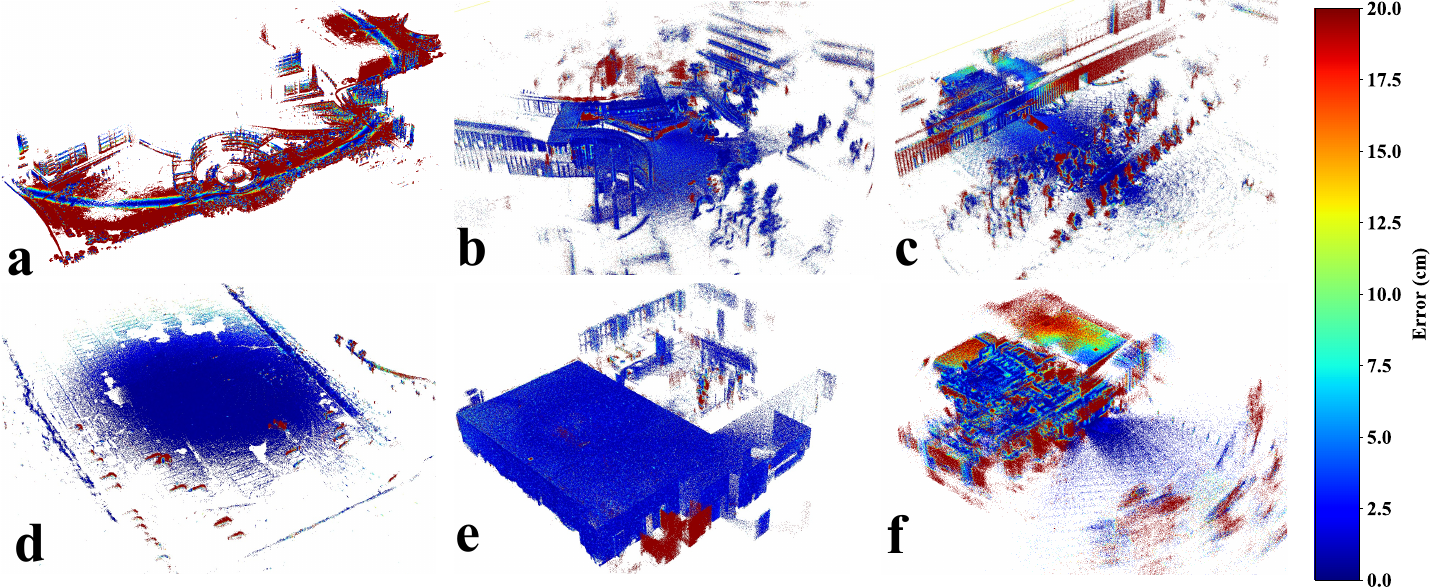}
  \caption{This figure presents the mapping performance of FAST-LIO2 in various environments: (a) \texttt{ugv\_campus00}, (b) \texttt{handheld\_escalator00}, (c) \texttt{legged\_grass00}, (d) \texttt{ugv\_parking00}, (e) \texttt{handheld\_room00}, and (f) \texttt{legged\_room00}.
    The color gradient, from red to blue, illustrates the range of errors in the map points generated by FAST-LIO2, with red indicating higher errors (up to \SI{20}{cm}) and blue denoting lower errors (down to 0), where deeper blue signifies higher mapping precision. Notably, (a) shows significant z-axis drift in an outdoor large-scale scenario, resulting in predominantly high-error red areas in the map evaluation. Conversely, (f) Illustrates the algorithm's application on a quadruped platform in an indoor office environment characterized by intense ground movement, glass-induced noise, and numerous dynamic obstacles, which are depicted by the red areas signifying higher error.}
  \label{fig:exp_map_evaluation}
\end{figure*}
\begin{figure}[t]
  \centering
  \includegraphics[width=0.48\textwidth]{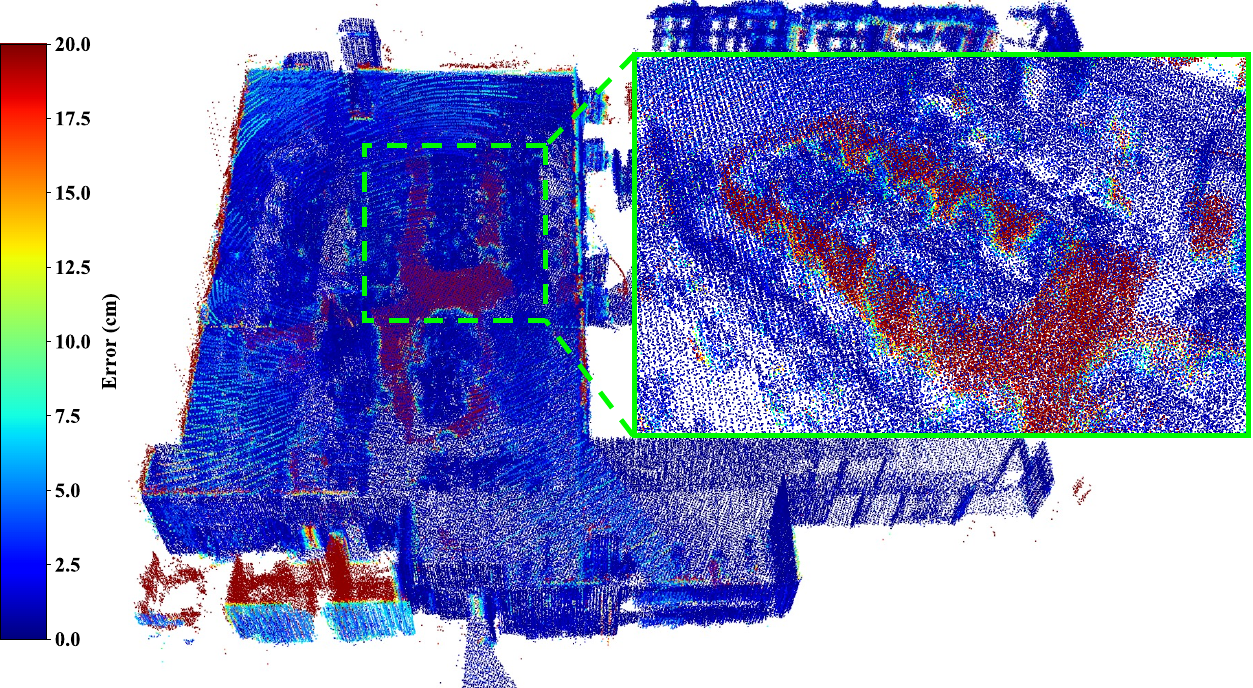}
  \caption{This figure demonstrates the map evaluation results within the \texttt{handheld\_room00} dataset using R3LIVE. The estimated point cloud map is compared against the ground truth map, with the color gradient from blue to red indicating accuracy discrepancies ranging from 0 to \SI{20}{cm}. The inset highlights significant errors in the seating area within the room.}
  \label{fig:rgb_point_cloud_error}
\end{figure}

We use the mapping evaluation metrics in PALoc \citep{hu2024paloc} to complement our localization evaluation. After initial alignment, the error metrics were then calculated using our map evaluation library as introduced in Section \ref{sec:gt_map}.
In the presence of high-precision RGB point cloud map ground truth, the accuracy of the maps reconstructed by the algorithm can be evaluated. We register the estimated point cloud map $\mathcal{M}$, reconstructed by the algorithm, to the ground-truth point cloud map $\mathcal{G}$, subsequently obtaining the corresponding set of associated points.
By computation of precision metrics on this set of associated points, we can also mitigate minor map variations due to temporal discrepancies in ground-truth point cloud collection and slight interference from dynamic obstacles.
We compare these two maps based on four metrics:
\textit{Reconstruction Error (RE)} in terms of the Root Mean Squared Error (RMSE), \textit{Completeness (COM)}, and \textit{Chamfer Distance (CD)}.
They are defined as below:
\begin{itemize}[leftmargin=0.7cm]
    \item \textit{Reconstruction Error} computes the average point-to-point distance between $\mathcal{M}$ and $\mathcal{G}$ \citep{pan2024pin}:
          \begin{equation}
              \textit{RE}
              =
              \sqrt{
                  \frac{1}{|\mathcal{M}|}
                  \sum_{\bm{p}\in\mathcal{M}}
                  {\underbrace{\min(\tau,\ \|\bm{p} - \bm{q}\|)}_{d(\bm{p}, \mathcal{G})}}^{2}
              }.
          \end{equation}
          where $\tau$ is the inlier distance and $\mathbf{q}\in\mathcal{G}$ is the nearest point to $\mathbf{q}$.
          We empirically set $\tau=0.2m$ in experiments.
    \item \textit{Completeness} describes how does $\mathcal{M}$ cover extent of $\mathcal{G}$. $\mathcal{G}'$ is the subset of $\mathcal{G}$. Each element of $\mathcal{G}'$ has a nearby point from $\mathcal{M}$ such as
          \begin{equation}
              \begin{split}
                  \textit{COM}
                  &=
                  \frac{|\mathcal{G}'|}{|\mathcal{G}|},\\
                  \mathcal{G}'
                  &=
                  \{\mathbf{q}\in\mathcal{G}\ |\ \ \
                  \exists \mathbf{p}\in\mathcal{M},
                  ||\mathbf{p}-\mathbf{q}||\leq\tau\}.
              \end{split}
          \end{equation}

    \item \textit{Chamfer Distance} computes the Chamfer-L1 Distance \citep{mescheder2019occupancy} as:
          \begin{equation}
              \textit{CD}
              =
              \frac{1}{2|\mathcal{M}|}
              \sum_{\bm{p}\in\mathcal{M}}
              d(\bm{p}, \mathcal{G})
              +
              \frac{1}{2|\mathcal{G}|}
              \sum_{\bm{q}\in\mathcal{G}}
              d(\bm{q}, \mathcal{M}).
          \end{equation}
\end{itemize}

Fig. \ref{fig:exp_map_evaluation} employs the map evaluation module to present the assessment outcomes of the FAST-LIO2 algorithm across six indoor and outdoor data sequences, with varying colors representing the accuracy levels across different map regions. The map accuracy estimated by FAST-LIO2 notably decreases in outdoor large-scale scenes (Fig. \ref{fig:exp_map_evaluation} (a)) or areas with dense vegetation (Fig. \ref{fig:exp_map_evaluation} (c) and Fig. \ref{fig:exp_map_evaluation} (f)), attributable to significant measurement noise from trees or overall z-axis drift in outdoor LiDAR odometry and mapping applications. Conversely, indoor settings, barring the effects introduced by dynamic obstacles (Fig. \ref{fig:exp_map_evaluation} (f)), predominantly exhibit high map quality (Fig. \ref{fig:exp_map_evaluation} (e)).

Table \ref{tab:exp_mapping_accuracy} further delineates the map evaluation results for FAST-LIO2 and R3LIVE across $6$ indoor and outdoor sequences, with R3LIVE retaining only points with RGB colors, hence showing inferior performance on the COM metric compared to FAST-LIO2. However, FAST-LIO2 outperforms R3LIVE in most scenes in terms of RE and CD metrics, particularly in handheld and quadruped robot sequences. In expansive campus environments, both algorithms exhibit comparable CD metrics, as seen in scenarios like \texttt{handheld\_escalator} and \texttt{ugv\_campus}. Fig. \ref{fig:rgb_point_cloud_error} illustrates the mapping results of R3LIVE on the \texttt{handheld\_room00} sequence, employing a color scheme consistent with that of Fig. \ref{fig:exp_map_evaluation}. The presence of glass within the room introduces noise, resulting in some blurred regions within the map. This depiction underscores the impact of environmental features on mapping clarity.

\begin{table}[t]
  \centering
  \caption{Mapping accuracy: we calculate four metrics to evaluate \textit{FAST-LIO2 (\textbf{FL2})} and \textit{R3LIVE (\textbf{R3L})}}
  \renewcommand\arraystretch{1.1}
  \renewcommand\tabcolsep{3pt}
  \footnotesize
  \begin{tabular}{lcccccc}
    \toprule[0.03cm]
    \multirow{2}{*}{Sequence} & \multicolumn{2}{c}{RE $[m,\downarrow]$} & \multicolumn{2}{c}{COM $[\%,\uparrow]$} & \multicolumn{2}{c}{CD $[m,\downarrow]$}                                         \\
    \cmidrule(lr){2-3} \cmidrule(lr){4-5} \cmidrule(lr){6-7}
                              & FL2                                     & R3L                                     & FL2                                     & R3L     & FL2          & R3L          \\

    \midrule[0.03cm]

    handheld\_room$00$        & $\bm{0.144}$                            & $0.269$                                 & $\bm{0.949}$                            & $0.802$ & $\bm{0.109}$ & $0.131$      \\
    handheld\_escalator$00$   & $\bm{0.273}$                            & $0.544$                                 & $\bm{0.846}$                            & $0.340$ & $0.128$      & $\bm{0.126}$ \\

    \midrule[0.03cm]

    legged\_grass$00$         & $\bm{0.092}$                            & $0.199$                                 & $\bm{0.818}$                            & $0.406$ & $\bm{0.158}$ & $0.161$      \\
    legged\_room$00$          & $0.442$                                 & $\bm{0.196}$                            & $\bm{0.445}$                            & $0.316$ & $\bm{0.132}$ & $0.163$      \\
    \midrule[0.03cm]

    ugv\_campus$00$           & $\bm{0.765}$                            & $0.767$                                 & $\bm{0.232}$                            & $0.217$ & $0.112$      & $\bm{0.107}$ \\
    ugv\_parking$00$          & $\bm{0.105}$                            & $0.122$                                 & $\bm{0.956}$                            & $0.567$ & $\bm{0.110}$ & $0.166$      \\

    \bottomrule[0.03cm]
  \end{tabular}
  \label{tab:exp_mapping_accuracy}
\end{table}

\begin{figure*}
    \centering
    \includegraphics[width=0.90\textwidth]{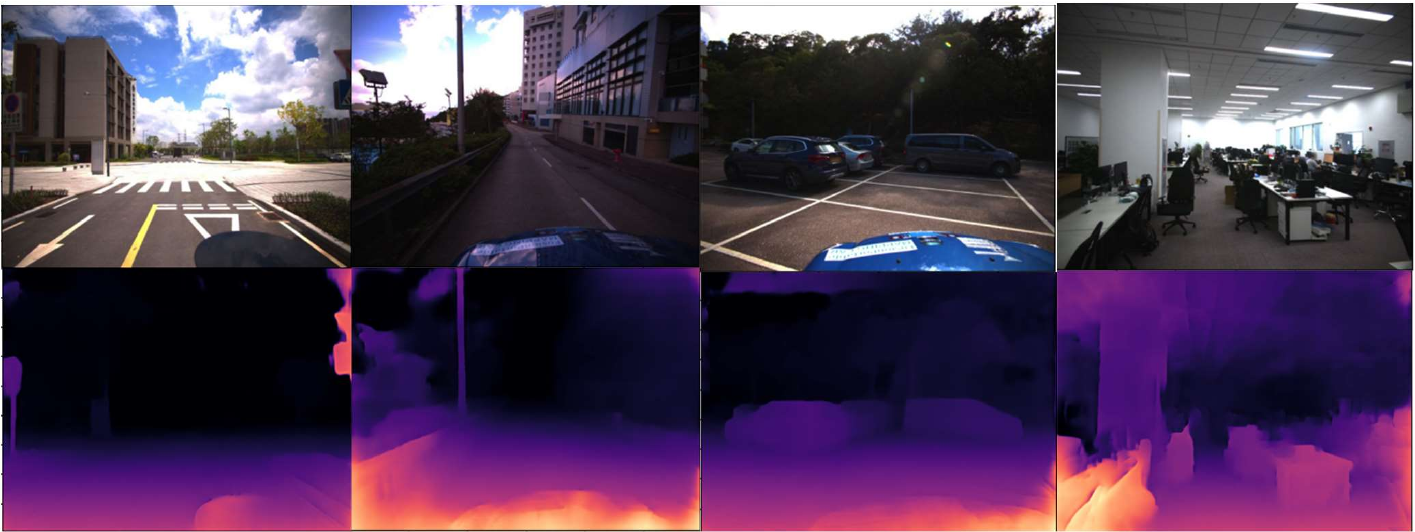}
    \caption{Results of the unsupervised depth prediction method \cite{Liu2023FSNet} which can generalize over different environments.}
    \label{fig:exper_monocular_perception}
\end{figure*}
\begin{table*}[t]
  \caption{Performance of FSNet MonoDepth \citep{Liu2023FSNet} on FusionPortableV2.
    Results are categorized by the training domain and testing domain.
    The left four metrics: ARD, SRD, RMSE-linear, and RMSE-log are error metrics (the lower the better).
    The right three metric: $\delta<\delta_{thr}$ are accuracy metrics (the higher the better).}
  \renewcommand\arraystretch{1.1}
  \renewcommand\tabcolsep{4.5pt}
  \centering
  \footnotesize
  \begin{tabular}{ccccccccc}
    \toprule[0.03cm]
    \multirow{1}{*}{Model}                            &
    \multirow{1}{*}{Test Sequence}                    &
    \multirow{1}{*}{ARD}                              &
    \multirow{1}{*}{SRD}                              &
    \multirow{1}{*}{RMSE-linear $[m,\downarrow],$}    &
    \multirow{1}{*}{RMSE-log}                         &
    \multirow{1}{*}{$\delta<1.25\ [\%,\uparrow]$}     &
    \multirow{1}{*}{$\delta<1.25^{2}\ [\%,\uparrow]$} &
    \multirow{1}{*}{$\delta<1.25^{3}\ [\%,\uparrow]$}                                                                                                                     \\
    \toprule[0.03cm]
    \multirow{2}{*}{FSNet-Handheld}
                                                      & Handheld & $0.592$      & $7.885$      & $5.750$      & $0.552$      & $0.440$      & $0.697$      & $0.825$      \\
                                                      & Vehicle  & $0.235$      & $1.724$      & $5.210$      & $0.287$      & $0.670$      & $0.887$      & $0.961$      \\
    \midrule[0.01cm]
    \multirow{2}{*}{FSNet-Vehicle}
                                                      & Handheld & $1.031$      & $15.786$     & $6.970$      & $0.742$      & $0.300$      & $0.531$      & $0.679$      \\
                                                      & Vehicle  & $\bm{0.125}$ & $\bm{1.522}$ & $\bm{4.561}$ & $\bm{0.199}$ & $\bm{0.882}$ & $\bm{0.955}$ & $\bm{0.978}$ \\
    \bottomrule[0.03cm]
  \end{tabular}
  \label{tab:depth_prediction}
\end{table*}

%%%%%%%%%%%%%%%%%%%%%%%%%%%%%%%%%%%%%%%%
%%%%%%%%%%%%%%%%%%%%%%%%%%%%%%%%%%%%%%%%
%%%%%%%%%%%%%%%%%%%%%%%%%%%%%%%%%%%%%%%% 
\subsection{Evaluation of Depth Estimation}
The diversity of sensors, mobile platforms, and scenarios make our dataset appealing for algorithm verification not limited to localization and mapping.
In this section, we demonstrate that our dataset can serve for the evaluation of advanced perception algorithms.
Due to the easily accessible GT, we set the benchmark for measuring the generalization ability of unsupervised monocular depth prediction. The benchmark measures how unsupervised monocular depth prediction networks could perform on scenes collected from different data collection platforms.

\subsubsection{\textbf{Data Preparation:}}
Each frame image is accompanied by a GT depth image of identical size for evaluation.
Depth images are produced by projecting point clouds,  generated by FAST-LIO2 \citep{xu2021fast} through IMU interpolation, onto these frames:
\begin{equation}
    D_{gt}( \mathbf{x})
    =
    Z,
    \ \ \
    \mathbf{x}
    =
    \lfloor \pi(\mathbf{p}^{c}) \rceil,
    \ \ \
    \mathbf{p}^{c}
    =
    \mathbf{R}^{c}_{l}\mathbf{p}^{l} + \mathbf{t}^{c}_{l}.
    \label{equ:image_projection}
\end{equation}
where $Z$ is the $z$-axis value of $\mathbf{p}^{c}$,
$\pi(\cdot)$ is the camera projection function,
$\lfloor \cdot \rceil$ is the rounding operation,
and
$(\mathbf{R}^{c}_{l}, \mathbf{t}^{c}_{l})$ represents the extrinsics from the left frame camera to the LiDAR.

%%%%%%%%%%%%%%%%%%%%%%%%%%%%%%%%%%%%%%%%%%%%%%
%%%%%%%%%%%%%%%%%%%%%%%%%%%%%%%%%%%%%%%%%%%%%%
%%%%%%%%%%%%%%%%%%%%%%%%%%%%%%%%%%%%%%%%%%%%%%
\label{sec:experiment_perception}
\subsubsection{\textbf{Experiment Setting:}}
Monocular depth estimation tests are essential for evaluating a system's ability to perceive relative object distances from a single camera view, a key aspect of understanding spatial relationships.
Based on the \href{https://github.com/Owen-Liuyuxuan/FSNet}{FSNet} that is a self-supervised depth estimation model \citep{Liu2023FSNet}, we fine-tune this model with our dataset using GT poses.
We organize the train-validation data into two groups.
The first group allocates $70\%$ of handheld indoor sequence data (\textit{i.e.}, \texttt{handheld\_room00} and \texttt{handheld\_escalator00}) for training, and combines $30\%$ of this data with all vehicle-related outdoor sequences for validation.
The second group uses $70\%$ of vehicle sequence data (\textit{i.e.}, \texttt{vehicle\_campus00} and \texttt{vehicle\_parking00}) for training, and blends $30\%$ of this data with all handheld sequences for validation.
We train the FSNet with these groups of data respectively and obtain two models:
\textbf{FSNet-Handheld} and
\textbf{FSNet-Vehicle}.

\subsubsection{\textbf{Evaluation:}}
We assess models' performance of unsupervised monocular depth prediction models use the proposed scale-invariant metrics in \citep{NIPS2014_eigens}: \textit{Absolute Relative Difference} (ARD), \textit{Squared Relative Difference} (SRD), \textit{Root Mean Squared Error} (RMSE)-linear, RMSE-log, and \textit{Threshold}.
They are defined as follows:
\begin{equation}
    \begin{aligned}
        \text{ARD}
         & =
        \frac{1}{|\mathcal{X}|}
        \sum_{\mathbf{x}\in\mathcal{X}}
        |d_{est} - d_{gt}|/d_{gt},           \\
        \text{SRD}
         & =
        \frac{1}{|\mathcal{X}|}
        \sum_{\mathbf{x}\in\mathcal{X}}
        \|d_{est} - d_{gt}\|^{2}/d_{gt},     \\
        \text{RMSE-linear}
         & =
        \sqrt{
        \frac{1}{|\mathcal{X}|}
        \sum_{\mathbf{x}\in\mathcal{X}}
        \|d_{est} - d_{gt}\|^{2}},           \\
        \text{RMSE-log}
         & =
        \sqrt{
        \frac{1}{|\mathcal{X}|}
        \sum_{\mathbf{x}\in\mathcal{X}}
        \|\log d_{est} - \log d_{gt}\|^{2}}, \\
        \text{Threshold}
         & =
        \% \ \text{of}\ d_{est}              \\
         & \ \ \ \ \ \ \text{s.t.}\
        \max(\frac{d_{est}}{d_{gt}}, \frac{d_{gt}}{d_{est}})=\delta \leq \delta_{thr}.
    \end{aligned}
\end{equation}
where $d_{est}$ is the estimated depth value at the pixel $\mathbf{x}$ with the corresponding GT depth $d_{gt}$ and
where $\delta_{thr}\in\{1.25, 1.25^{2}, 1.25^{3}\}$.
Quantitative and some qualitative results are presented in Fig. \ref{fig:exper_monocular_perception} and Table \ref{tab:depth_prediction}, respectively.
These results not only validate our dataset for depth estimation but also highlight the limitations of current unsupervised depth prediction techniques.
These limitations are revealed as the FSNet-Handheld model struggles with generalization to handheld sequences, despite training on data with similar appearance.
The challenge for monocular depth estimation is further amplified by the significant scale variation in indoor sequences.
Advancements in depth formulation and learning strategies are expected to markedly improve the performance in future benchmarks.
For a more detailed and comprehensive analysis, we encourage viewing the dataset videos available on our website.

%%%%%%%%%%%%%%%%%%%%%%%%%%%%%%%%%%%%%%%%%%
%%%%%%%%%%%%%%%%%%%%%%%%%%%%%%%%%%%%%%%%%%
%%%%%%%%%%%%%%%%%%%%%%%%%%%%%%%%%%%%%%%%%%
\subsection{Known Issues and Limitations}
\label{sec:known_issue}
Creating a comprehensive dataset spanning multiple platforms, sensors, and scenes is labor-intensive.
Despite our efforts to resolve many issues, we acknowledge the presence of several imperfections within the dataset.
We detail these common challenges in the subsequent sections and present our technical solutions.
We hope this discussion will provide valuable insights and lessons for future researchers.

\subsubsection{\textbf{Calibration:}}
Achieving the life-long sensor calibration poses significant challenges \citep{maddern20171}.
We try our best to provide the best estimate of calibration parameters.
Calibration was performed each time when the data collection platform changed, employing SOTA methods for parameter adjustments, which were also manually verified and fine-tuned.
For extrinsic parameters difficult to estimate, such as the relative transformation between specific components, we refer to the CAD model.
Efforts were made to reinforce the mechanical structure and minimize external disturbances during the data collection process.
Nevertheless, it is acknowledged that high accuracy for specific traversals cannot be assured.
Users are encouraged to use our calibration estimates as initial values and explore innovative approaches for long-term extrinsic calibration, such as \citep{wu2021simultaneous,ulrich2023uncertainty,luo2024zero}.
To aid in these endeavors, we provide raw calibration data and reports, allowing users to develop their methodologies and consider our estimates as a foundational benchmark.

\subsubsection{\textbf{Synchronization:}}
Section \ref{sec:synchronization} presents our hardware synchronization solution that guarantees the IMU, frame cameras, and LiDAR are triggered by the same clock source.
However, the timestamp of the ROS message of each sensor data has minor differences since the time of data transmission and decode varies.
For vehicle-related sequences, the average relative time latency (ARTL) among stereo frame images is smaller than $20ms$. This is mainly caused by the long connection between the camera and the signal trigger.
For other sequences, the ARTL is smaller than $5ms$.
Due to the special design of the event cameras, the ARTL between the left event camera and the LiDAR is unstable and sometimes smaller than $15ms$.

\subsubsection{\textbf{Partial Loss of Information:}}
In the construction of the dataset, real-world challenges have led to partial loss of sensor information in certain sequences, reflecting practical issues encountered during robotic deployment. Specifically, in \texttt{ugv\_transition00} and \texttt{ugv\_transition01}, the wheel encoder driver was not activated correctly, resulting in the absence of wheel encoder data. Additionally, the \texttt{legged\_grass00} sequence experienced a brief interruption in data transmission, amounting to several seconds of lost data, due to a loose RJ45 network port connection. These instances underscore the importance of robustness in algorithm development to handle incomplete or missing sensor data in realistic operational conditions.

\subsubsection{\textbf{Camera Exposure Setting:}}
To ensure image consistency during the whole sequence, we fixed the camera exposure time with a specific value before collecting each sequence, mitigating color varies from illumination changes.
This scheme is also important to stereo matching since consistent brightness is commonly desirable.
However, this scheme can darken images in significantly different lighting conditions, such as entering a tunnel. The darker appearance can be a challenge for most visual perception algorithms.

\subsubsection{\textbf{Limited Diversity and Volume:}}
\label{sec:scale_model_issue}
Drawing inspiration from foundational models in computer vision and natural language processing, developing a robot-agnostic general model for robotics could be an exciting avenue for future research \citep{padalkar2023open,khazatsky2024droid,shah2023gnm}.
However, acquiring a large, diverse real-world dataset poses significant challenges.
We address key problems in data collection for field robots, including system integration and data postprocessing.
But we acknowledge the limitations in our dataset's scale and diversity:
(1) \textbf{Platforms}: Our dataset does not include drones, underwater robots, or multi-robot systems.
(2) \textbf{Scenarios}: Critical environments such as factories, forests, and underground mines, which are also important for SLAM but not presented.
(3) \textbf{Time Periods}: The dataset was collected over a short time frame, lacking day-night cycles, seasonal variations, and structural changes, which are crucial for place recognition and long-term SLAM \citep{carlevaris2016university,barnes2020oxford}.
(4) \textbf{Volume}: The dataset is relatively small, with a limited number of sequences and short durations.
FusionPortableV2 is just the beginning. As more researchers and institutions contribute to this field, we anticipate the development of larger, more diverse datasets that will support the creation of a truly generalizable robotic model.

\section{Conclusion and Future Work}
\label{sec:conclusion}
\runninghead{Wei \textit{et~al.}}

This paper presents the FusionPortableV2 dataset, a comprehensive multi-sensor collection designed to advance research in SLAM and mobile robot navigation. The dataset is built around a compact, multi-sensor device that integrates IMUs, stereo cameras (both frame-based and event-based), LiDAR, and INS, all carefully calibrated and synchronized. This primary device is deployed on various platforms, including a legged robot, a low-speed UGV, and a high-speed vehicle, each equipped with additional platform-specific sensors such as wheel encoders and legged sensors.
The FusionPortableV2 dataset features a diverse range of environments, spanning indoor spaces, grasslands, campuses, parking lots, tunnels, downhill roads, and highways. This environmental diversity challenges existing SLAM and navigation technologies with realistic scenarios involving dynamic objects and variable lighting conditions. To ensure the dataset's utility for the research community, we have meticulously designed 27 sequences, totaling 2.5 hours of data, and provided ground truth data for the objective evaluation of SOTA methods in localization, mapping, and monocular depth estimation.

As we explore future directions, we aim to enhance this dataset's applicability beyond SLAM by developing novel navigation methods based on the proposed dataset. We will continue to improve the quality of the data and the integration of the system to facilitate easier use by non-expert users. Alongside this dataset, we also release our implementation details and tools to encourage further research advancements.

%%%%%%%%%%%%%%%%%%%%%%%%%%
\begin{acks}
    This research greatly benefited from the guidance and expertise of many contributors. We extend our profound gratitude to colleagues: Ruoyu Geng, Lu Gan, Bowen Yang, Tianshuai Hu, Mingkai Jia, Mingkai Tang, Yuanhang Li, Shuyang Zhang, Bonan Liu, Jinhao He, Ren Xin, Yingbing Chen, etc. from HKUST and HKUSTGZ for their suggestions in improving our dataset's quality. Special acknowledgment goes to the BIM Lab at HKUST, particularly Prof. Jack Chin Pang Cheng and his students, for their crucial BLK360 scanning expertise and insights that were essential in shaping our dataset.
    The authors also gratefully acknowledge Dr. Thien-Minh Nguyen (NTU) and Ms. Qingwen Zhang (KTH) for their insightful feedback and technical support; Prof. Dimitrios Kanoulas (UCL), Prof. Martin Magnusson (Örebro University), Prof. Hong Zhang (Southern University of Science and Technology), and Prof. Peng Yin (CityU Hong Kong) for their invaluable dataset writing advice; Mr. Seth G. Isaacson (University of Michigan) for his suggestions on dataset structure.
    ; and OpenAI's GPT for enhancing the paper's linguistic quality.
    The author(s) received no financial support for the research, authorship, and/or publication of this article.
\end{acks}
%%%%%%%%%%%%%%%%%%%%%%%%%%%
% \begin{funding}

% \end{funding}
%%%%%%%%%%%%%%%%%%%%%%%%%%
\begin{dci}
    The authors declared no potential conflicts of interest with respect to the research, authorship, and/or publication of this article.
\end{dci}
%%%%%%%%%%%%%%%%%%%%%%%%%%

\theendnotes

%\clearpage
\normalem
\bibliographystyle{SageH}
\bibliography{ref.bib}

% \clearpage
% \input{supp/supp}

\end{document}